\pdfoutput=1
\documentclass{article}
\PassOptionsToPackage{hyphens}{url}

\usepackage{iclr2027_conference,times}

\usepackage[utf8]{inputenc}
\usepackage[T1]{fontenc}
\usepackage{hyperref}
\usepackage{url}
\usepackage{booktabs}
\usepackage{amsfonts}
\usepackage{amsmath}
\usepackage{amssymb}
\usepackage{nicefrac}
\usepackage{microtype}
\usepackage{xcolor}
\usepackage{graphicx}
\usepackage{subcaption}
\usepackage{multirow}
\usepackage{listings}
\usepackage{float}
\usepackage{placeins}  %
\usepackage[skins,breakable]{tcolorbox}
\usepackage{needspace}

\tcbset{
  leaderboardstyle/.style={
    enhanced,
    colback=blue!4!white,
    colframe=blue!45!black,
    coltitle=white,
    colbacktitle=blue!45!black,
    fonttitle=\bfseries,
    titlerule=0pt,
    arc=2pt,
    boxrule=0.7pt,
    left=10pt, right=10pt, top=6pt, bottom=6pt,
    width=\linewidth,
  }
}

\graphicspath{{figures/}}

\newcommand{\EthreePairs}{36}
\newcommand{\EthreeRho}{$+0.82$}

\newcommand{\EthreeOwnRho}{$+0.82$}
\newcommand{\EthreeValueRho}{$+0.54$}
\newcommand{\PooledSixRho}{$+0.95$}

\lstdefinestyle{rulebook}{%
  basicstyle=\ttfamily\fontsize{6pt}{7pt}\selectfont,
  breaklines=true, breakindent=0pt, breakatwhitespace=true,
  columns=fullflexible, keepspaces=true, showstringspaces=false,
  inputencoding=utf8, extendedchars=true,
  literate={—}{{--}}1 {…}{{...}}1 {«}{{<<}}1 {»}{{>>}}1, upquote=true,
  aboveskip=0pt, belowskip=0pt, xleftmargin=4pt,
  framesep=2pt, framerule=0.2pt,
}

\title{\textsc{GENSTRAT}: Toward a Science of Strategic Reasoning in Large Language Models}

\author{%
  Vartan Shadarevian\thanks{Corresponding author. \texttt{vartan@princeton.edu}} \\
  Princeton University \\
  \And
  Kia Ghods \\
  Princeton University \\
  \And
  Alex Kenich \\
  Google \\
  \And
  Anany Kotawala \\
  Princeton University \\
}
\iclrfinalcopy

\begin{document}

\maketitle
\lhead{Preprint}

\begin{abstract}
Fixed game suites for evaluating strategic reasoning in large language models (LLMs) saturate and are
hard to keep free of contamination. We introduce \textsc{GENSTRAT}, a system for evaluating the strategic
capabilities of LLMs driven by procedural generation of two-player zero-sum imperfect-information betting
games, allowing an evaluator to draw novel, fresh games on demand. Each game is characterized on a scale
of five complexity axes (state space, temporal contingency, opponent dependence, private information,
precision) and measured by Monte Carlo simulation. Using this generator, we construct a 50-game benchmark
and hold a tournament of twelve frontier and open-weight models, gathering data from over $40{,}000$ LLM
vs.\ LLM matches. We further construct a composite complexity axis, and find that, on the eight hardest
games, the gap between the best and worst model is $1.7\times$ its value across the whole set of games ($1.6\times$ with held-out weights or with hard games chosen by the outcome-free temporal
axis alone, and $1.35\times$ with payoffs scaled by their random-play standard deviation).
Ablations show that a higher thinking budget, disclosing the opponent's strategy, and higher reasoning
matched to a game's specific demands (for private information and planning depth) were associated with higher margins for the LLM player.
On 27 solved games, frontier models play at the value of the equilibrium while weaker models lose to it, and the
leaderboard predicts strength on fresh games and on six new games, even after controlling for a
general-capability leaderboard.
\end{abstract}

\section{Introduction}
\label{sec:intro}

This paper introduces \textsc{GENSTRAT}, a framework for evaluating LLMs' strategic capabilities built upon a
procedural generator of generalized betting games (GBGs), a family of two-player zero-sum imperfect-information
card games. The generator can continually generate fresh imperfect-information games with novel features and
structure. We propose five complexity axes---state space, temporal contingency, opponent dependence, private
information, and precision---to characterize the space of games and study LLM performance across the games
the generator can create. We then run a 12-model tournament on a benchmark of 50 generated games,
consisting of over $40{,}000$ matches. We find that general strategic intelligence is a phenomenon in the space
of GBGs: newer and larger models generally pull ahead in the leaderboard, especially compared to open-source models. The differences between
models grow with the complexity of the games, most of all on games where the best action depends on how the rest
of the deal unfolds, suggesting that as the current games saturate, we can continue to generate larger, more
complex games. Furthermore, even within the GBG space, models exhibit significant heterogeneity in how their
abilities vary over the space of games, suggesting that different models and training regimes may lead to
different strategic capabilities.

Frontier LLMs are placed in economic-agent roles in controlled experiments, ranging from running small commerce
operations \citep{anthropic2025projectvend1} to marketplace simulations \citep{anthropic2026projectdeal}. Models
have also exhibited algorithmic-collusion behavior in LLM-based pricing studies
\citep{fish2024algorithmiccollusion}. How a model will perform once deployed is hard to anticipate, and
performance on canonical games---especially ones for which a model has training data to fall back on---may not
cleanly transfer to the environment a deployer cares about.

Existing strategic-reasoning benchmarks evaluate LLMs on fixed games or suites. These include poker settings
\citep{guo2024suspicion,huang2024pokergpt}, AvalonBench \citep{light2023avalon}, Diplomacy \citep{cicero2022},
and broader suites such as GTBench and GameBench \citep{duan2024gtbench,costarelli2024gamebench}. These benchmarks
share three limits. First, fixed suites may saturate as the frontier improves. Second, canonical games often exist in the training
data of LLMs, which makes training-data contamination hard to rule out. Finally, benchmarks with fixed or
canonical games do not show how model performance shifts as the environment shifts, limiting the deployer's
ability to understand how a model will generalize to novel environments, especially ones with adversarial
elements.

\textsc{GENSTRAT} addresses these limits with a procedurally generated distribution of GBGs. Procedural
generation has been shown to be productive for single-agent reinforcement learning (RL)
\citep{cobbe2020procgen,chevalier2023minigrid}. Multi-agent settings add an \emph{amplification effect}, whereby
small increases in environment complexity can produce large increases in the complexity of the strategic problem
an agent must solve. Because a generator can draw freely from the same distribution, a fresh draw is very
likely not one a model will have seen before or memorized. Freshness rules out memorizing a game's solution. It does
not rule out, and is not meant to rule out, the transfer of general strategic knowledge, which is what we aim to measure. We then consider five complexity axes that describe
what each game demands, computed by Monte Carlo simulation with randomly generated strategies, and provide a
capability-profile decomposition for each LLM.

In the tournament we conduct, larger, more recent, reasoning-capable models emerge as the more competent players
of GBGs. Significant heterogeneity emerges, however. \texttt{gemma-4-31b}, for example, gains on games with high
private information ($+0.19$ chips/game per standard deviation of the axis, Section~\ref{sec:profiles}). We conduct ablations, including against fixed
LLM-authored programs, both disclosed and hidden, and at different levels of thinking budget. Predictably,
disclosing a program's strategy helps, by about $0.7$ chips per game, and so does a higher thinking budget, in
every model family tested (significant in one of four, Section~\ref{sec:programs}). Moreover, we use an LLM
judge to examine the reasoning traces of LLMs, finding that when LLMs use more of the types of reasoning
associated with a game's particular complexities, they tend to earn higher margins on that game, for private information and planning depth. Finally, we
report local jaggedness, a noise-corrected measure of how far an LLM's advantage departs from a smooth function of
the games' properties. Exact equilibria for 27 games measure how far each model is from perfect play. The leaderboard also predicts how
models do on games they have never seen, from generated card games without betting to a board game, a market and a
negotiation built with the same engine, better than a general-capability leaderboard does. In single-deal play, the frontier leads most where the best action depends on how the deal unfolds,
and reasoning about the opponent does not yet pay off (Section~\ref{sec:beyond}).

\section{Related work}
\label{sec:related}

Strategic games have a long history in AI research, from AlphaZero \citep{silver2018alphazero}, Libratus
and Pluribus \citep{brown2018libratus,brown2019pluribus}, DeepNash \citep{perolat2022deepnash} and CICERO
\citep{cicero2022}, which are specialized systems for single games, to benchmarks of general-purpose LLMs
on fixed games such as GTBench \citep{duan2024gtbench}, GameBench \citep{costarelli2024gamebench} and
AvalonBench \citep{light2023avalon}. \citet{akata2025repeated} study LLMs in repeated games,
\citet{lore2024strategic} disentangle game structure from framing, \citet{collins2026evalgames} probe
LLMs' ability to evaluate novel games, and \citet{lin2026toolpoker} study professional-poker-style tasks
with tool use. \citet{li2026modular} use LLM agents to discover general game-playing algorithms and test them on more than 400 games. The theory-of-mind literature is close \citep{strachan2024tom,ullman2023tom}, and the
poker-ToM coding scheme of \citet{lin2026readableminds} reads strategic reasoning from traces, as our
judge rubric does. Most closely related, gg-bench \citep{verma2025ggbench} generates games by LLM
authoring and scores win rate against a self-play RL agent. In contrast, with \textsc{GENSTRAT}, game rules come from a parameterized, highly controllable generator. Games are also imperfect information games, and thus test a model's ability to perform under uncertainty and asymmetric information. Complexity is an input we set directly, and raising it requires no rebuilding.
We decompose performance along measured axes, and a large-scale tournament maps model performance across
them.
Our concern with generalization off the evaluation distribution follows
\citet{vafa2024humangeneralization,vafa2024worldmodel,vafa2025foundationmodel}. Procedural generation is
standard in single-agent RL \citep{cobbe2020procgen,chevalier2023minigrid,shaker2016pcg}.

\section{Generalized betting games and GENSTRAT}
\label{sec:gbg}

We call the games we study \emph{generalized betting games} (GBGs). A GBG is a two-player zero-sum extensive-form game
of imperfect information. Kuhn poker \citep{kuhn1950} and Leduc poker \citep{southey2005leduc} are both GBGs,
and the family adds non-betting actions and rounds, other game-tree and information structures, and
conditional branches. A game consists of a deck, private hands and other piles, a graph of phases (betting, simultaneous-move,
auction and observation), and conditions on its branches. The generator randomizes structure, starting
from the phase graph. Inside the graph it draws ranks, suits and hand sizes, betting order, and the auction
and simultaneous rounds. It also draws observation triggers, position criteria, showdown metrics, side bets and
branch predicates. A series of structural checks and repairs helps ensure games are coherent and playable
(Appendix~\ref{app:design}). The integer seed and its stored generation configuration are sufficient to fully regenerate a game. To track the model frontier, complexity can
be raised by relaxing the generator caps. A draw is rejected if random play takes more than ten moves per player on average, if one seat wins more than 65\% of the decided deals under random play, if more than 40\% of its branch conditions hold almost always or almost never, if more than 20\% of its phases fire in fewer than 5\% of episodes, or if a branch can never fire. Of $12{,}351$ sampled seeds, $2{,}000$ were accepted.

\paragraph{A concrete game.} Seed 4643 (full rulebook in Appendix~\ref{app:rulebook_sample}) uses a deck of eight cards (10, J, Q, K in two suits), with three private cards per player, 10 chips per player and an ante of 3. Hidden positions are assigned from the dealt hands, and a player who holds one peeks at one of the opponent's cards. Betting has up to two rounds, the first with any stake up to the player's stack and the second with fixed 1-chip bets. Between them each player may peek at an opponent card at a 50\% risk of paying a chip. Both players then draw a card, and the chip leader gets a hidden position with a further peek. After betting comes a simultaneous round (Press, Hold, or Bluff) if either player holds a pair, and a card swap is offered if Alice holds a King. At showdown a hand scores the sum of its ranks, plus 3 for a pair of Js and minus 3 for each King. A player needs a plan for subsequent phases, and must engage in reasoning about opponent strategy and private information on both sides. Among the 50 games it sits at the 82nd percentile on state space and the 88th on temporal contingency (Section~\ref{sec:axes}).

\subsection{Five complexity axes}
\label{sec:axes}

We describe each game along five axes, all measured through simulated play without using LLM
results. The simulations use a population of simple policies: uniform random play and a subset of the 200 sampled policies, selected for performance. The axes are as follows
(full definitions are in Appendix~\ref{app:axis_formals}).

\begin{itemize}\itemsep1pt
\item \textbf{State space.} This measures the combinatorial size of the game, as seen by a player.
We estimate it by counting distinct information states in 3{,}000 deals of random play, adding the
counts for the two seats and taking $\log_{10}$. A state includes the decision, the player's own cards and
chips, its last three decisions, the events it has observed and the positions.

\item \textbf{Temporal contingency.} In a game with several phases, a good decision now may depend
on what the players will do later. We measure this dependence by simulating each available action
with different policies for subsequent play, and with different draws of the unseen cards. Some of
the payoff variation comes from actions being better on average, and some from favorable cards
or play regardless of the action. Some variation comes from their interaction: how much the comparison
between actions changes with the conditions of play. Temporal contingency is measured as this interaction's share of total variance.

\item \textbf{Opponent dependence.} Opponent dependence measures how much the relative payoffs of actions depend on
who the opponent is. We average over cards and our player's later choices, then measure variation
in these payoff differences across opponents.

\item \textbf{Information.} For this axis, we measure how much a player using limited-depth search, which
chooses each action by simulating the rest of the deal (Appendix~\ref{app:axis_formals}),
gains from planning around its own cards. We compare two such players. One searches over deals
consistent with the cards it has seen. The other does not condition its simulated deals on its own hand. We measure the difference in
their average chip leads per deal against the same opponents. The cards in the actual game
are unchanged.

\item \textbf{Precision.} The precision axis measures the cost of small departures from the best
action. We compare the best action with its neighbors, using adjacent amounts of the wager grid
for bets and bids (Appendix~\ref{app:axis_formals}), and all other actions for categorical choices. The loss is its payoff minus the average
over those neighbors. To limit selection bias, the best action is chosen on one half of the
simulations and its payoff evaluated on the other.
\end{itemize}

The temporal, opponent and precision scores are averages over decisions encountered during play,
where each visit to a state counts once. For all four axes other than state space, we average over the two seats.
Scores measured in chips are divided by the standard deviation of chip lead in population play so that
different stakes are comparable. We standardize each axis across the 50 games for analysis
(Section~\ref{sec:profiles}). Pairwise correlations are at most $|\rho|=0.50$, and the largest variance inflation
factor is $2.0$. Appendix~\ref{app:axis_formals} reports the estimation checks and the checks from repeated runs.

\section{Benchmark construction}
\label{sec:construction}

\begin{figure}[t]
\centering
\includegraphics[width=0.62\textwidth]{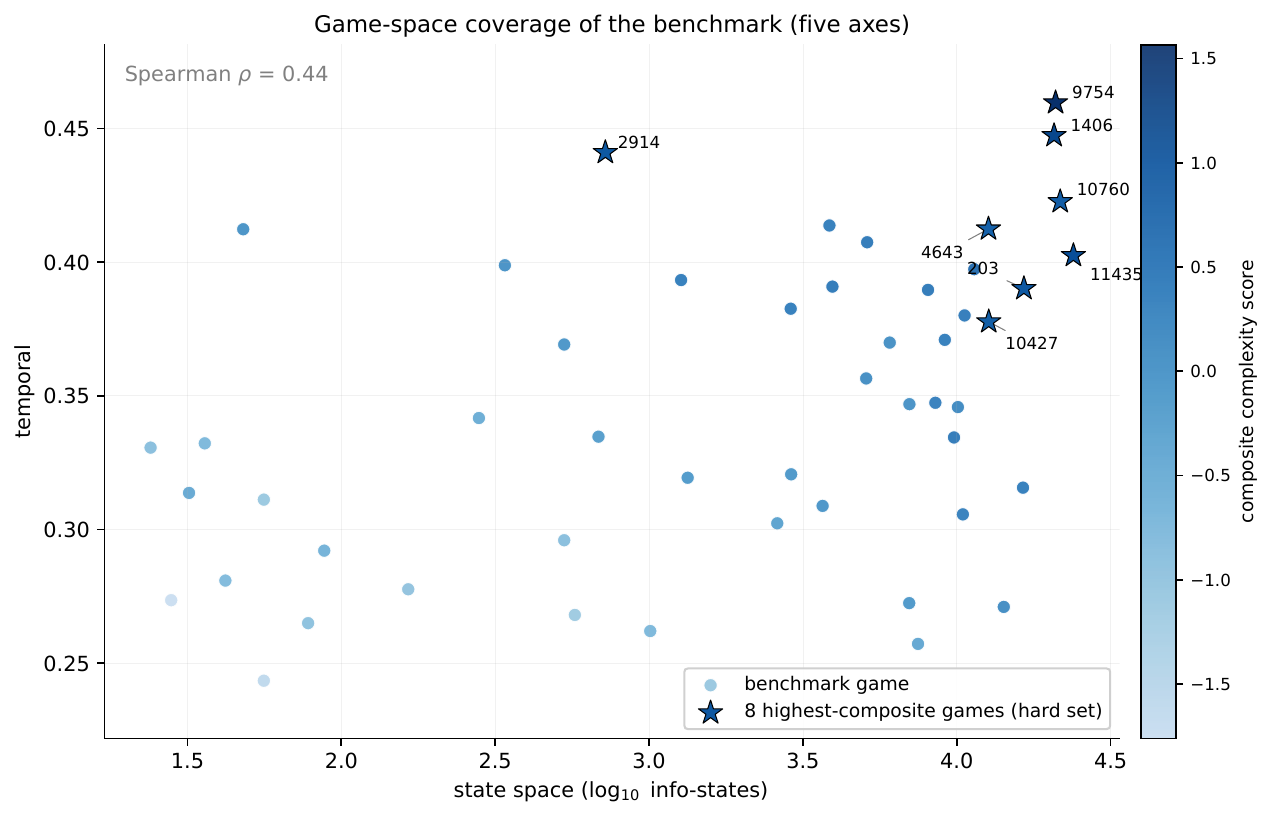}
\caption{\textbf{The 50-game benchmark in two of the five axes} (state space vs.\ temporal), colored by
composite complexity (Section~\ref{sec:leaderboard}). Stars mark the eight hardest games by composite complexity score
(Table~\ref{tab:hard8}).}
\label{fig:game_space}
\end{figure}

We selected 50 games from the accepted pool of 2,000 using farthest-point sampling on a sampling space composed of six preliminary complexity axes (Appendix~\ref{app:axis_formals}). Each axis was min-max normalized. In selecting games, each next game is picked to maximize its minimum distance to the games already selected. Then, we compute the five axes used in the paper on the set of 50 games. Figure~\ref{fig:game_space} shows the 50 games in two of the five axes.

\section{Tournament design}
\label{sec:setup}

\paragraph{Models.} Our tournament features twelve models. Nine models were played in an April--May 2026 tournament block:
\texttt{gpt-5.4}, \texttt{gemini-3.1-pro}, \texttt{claude-sonnet-4.6}, \texttt{gemini-2.5-pro}, \texttt{gemma-4-31b},
\texttt{deepseek-v3.1}, \texttt{gemini-3.1-flash-lite}, \texttt{qwen-3.5} and \texttt{llama-3.3-70b}. For those models,
every model played every other model on a subset of games. Three more models were added in a September 2026 tournament
block: \texttt{gpt-6-astra}, \texttt{claude-fable-5.1} and \texttt{gemini-3.8-flash}. Each September model played three
to five other models spread across all 50 games. Because the September models played fewer models each, they have wider
confidence intervals, and we mark them $^\dagger$ throughout. Appendix~\ref{app:models} gives each model's provider
identifier, and every request setting is recorded with the data so that runs can be reproduced.

\paragraph{Runs and single-deal matches.} For each (model pair, game) matchup, a \emph{run} (a paired deal) consists of two deals of the game, where each model plays once as Alice and once as Bob. The two deals share a play seed, so that each model in a seat faces the same conditions. Each matchup is played for up to 20 runs. Every match is a single deal, and models have no memory of earlier deals or identifying information about their opponent. Thus, each match tests planning under uncertainty with a fixed, unobservable opponent. Within a deal, a model sees only the opponent's (visible) actions so far, so opponent dependence is tested through a model's reasoning about likely opponent behavior, and the little evidence available within the deal.

\paragraph{Prompting and corpus.} First, a model is provided with the rulebook as a system prompt. At each turn, a model is also given the information it would have been able to observe so far, and a legal action menu. It must then respond with an action, given in JSON format. A lenient parser attempts to fix minor formatting issues, and should it fail, falls back to a random legal action (in the worst case, this was $0.33\%$ of moves, Appendix \ref{app:fallback}). After removing time-outs and other failed or defective matches, and keeping only complete pairs (i.e., full runs), the full tournament has $40{,}412$ matches. Counts of paired deals per model pair are in Appendix \ref{app:h2h_table}.

\section{Overall results}
\label{sec:leaderboard}

\paragraph{Estimator.} For match $s$ between models $i$ and $j$, we regress the signed chip margin, the chips $i$ gains minus the chips $j$ gains (winning 3 chips is a margin of 6), on model
indicators: $y_s = \alpha_i - \alpha_j + \varepsilon_s$ with $\sum_m \alpha_m = 0$. This is effectively Bradley--Terry
\citep{bradley1952rank} with chip margins in place of win indicators, since models were instructed to maximize expected
chips. $\hat\alpha_m$ is an estimate of a model's strength in chips per game. Standard errors come from a bootstrap over runs ($B{=}2{,}000$).

\begin{table}[t]
\centering\small
\caption{\textbf{Overall leaderboard} on April block and September block together ($20{,}206$ paired deals,
50 games), $\hat\alpha$ in chips/game, clustered bootstrap 95\% CIs ($B{=}2{,}000$).
$^\dagger$ September models.}
\label{tab:leaderboard}
\setlength{\tabcolsep}{8pt}
\begin{tabular}{lrr}
\toprule
\textbf{Model} & $\hat\alpha$ & \textbf{95\% CI} \\
\midrule
\texttt{gpt-6-astra}$^\dagger$ & $+0.72$ & $[+0.52, +0.94]$ \\
\texttt{gpt-5.4} & $+0.64$ & $[+0.55, +0.73]$ \\
\texttt{gemini-3.8-flash}$^\dagger$ & $+0.63$ & $[+0.42, +0.84]$ \\
\texttt{gemini-3.1-pro} & $+0.63$ & $[+0.56, +0.70]$ \\
\texttt{claude-sonnet-4.6} & $+0.45$ & $[+0.34, +0.56]$ \\
\texttt{claude-fable-5.1}$^\dagger$ & $+0.45$ & $[+0.24, +0.64]$ \\
\texttt{gemini-2.5-pro} & $+0.16$ & $[+0.08, +0.25]$ \\
\texttt{gemma-4-31b} & $+0.05$ & $[-0.03, +0.13]$ \\
\texttt{deepseek-v3.1} & $-0.14$ & $[-0.22, -0.06]$ \\
\texttt{gemini-3.1-flash-lite} & $-0.47$ & $[-0.56, -0.39]$ \\
\texttt{qwen-3.5} & $-0.55$ & $[-0.64, -0.47]$ \\
\texttt{llama-3.3-70b} & $-2.56$ & $[-2.67, -2.45]$ \\
\bottomrule
\end{tabular}

\end{table}

\paragraph{Leaderboard.} About three chips per game separate the top and bottom of Table~\ref{tab:leaderboard}.
The highest ranked models are \texttt{gpt-6-astra}, \texttt{gpt-5.4}, \texttt{gemini-3.8-flash} and
\texttt{gemini-3.1-pro}. \texttt{claude-sonnet-4.6} and \texttt{claude-fable-5.1} follow, tied at $+0.45$.
\texttt{gemini-2.5-pro}, \texttt{gemma-4-31b} and \texttt{deepseek-v3.1} form a middle tier, and
\texttt{gemini-3.1-flash-lite}, \texttt{qwen-3.5} and \texttt{llama-3.3-70b} the tail. Refitting only the April round-robin gives the nine April models in this same order, Kendall
$\tau{=}1.00$ (Appendix~\ref{app:no_llama}). The September models, whose intervals are widest, are the least
settled: \texttt{gpt-6-astra} may or may not lead the next three.

\paragraph{Robustness.} Leave-one-out refits with one game removed give mean Kendall $\tau{=}0.956$ (min $0.879$), with
tiers intact every time, so models change places only within a tier. Dropping the bottom outlier
(Appendix~\ref{app:no_llama}) gives the same picture. We also refit the leaderboard on random subsets of 12
games, and find a median Spearman correlation of $0.94$ with the full ranking (Appendix~\ref{app:leaderboards}). Of the 47 model pairs that met, 41 have a mean
head-to-head margin in the direction of the leaderboard, and none of the six exceptions is significant
(Appendix~\ref{app:h2h_table}). A win-indicator
Bradley--Terry fit orders the models somewhat differently ($\rho{=}0.77$, Appendix~\ref{app:bt}).

\paragraph{Strength on each game.} We also fit the estimator separately on each game, and write $\hat\alpha_{m,g}$ for the
strength of model $m$ on game $g$, in chips per game. On each $g$ the twelve $\hat\alpha_{m,g}$ sum
to zero and, as in the overall fit, are net of the opponents' strength on $g$. No cell of the $12\times50$ grid
is empty. Splitting its variance two ways (model main effect, model$\times$game interaction) and correcting for
estimation noise in each cell, the interaction is about a third of the main effect (ratio $0.34$, bootstrap 95\% CI
$[0.20, 0.54]$), and comparable to it once the outlier
\texttt{llama-3.3-70b} is excluded ($0.65$, $[0.14, 1.31]$, Appendix~\ref{app:variance}).
Half of the uncorrected ratio of $0.62$ is noise.

\paragraph{The hardest games.} We collapse the five axes into a composite complexity score, a learned discrimination score fit to the model outcomes: the
first principal component of the models' slopes of $\hat\alpha_{m,g}$ on each axis separately (weights:
temporal $+0.35$, state space $+0.29$, information $+0.09$, precision $-0.19$, opponent dependence $-0.07$,
Appendix~\ref{app:leaderboards}). Refitting on the eight highest-scoring games
(Table~\ref{tab:hard8}) widens the spread from $3.3$ chips/game on all fifty to
$5.6$, $1.7$ times. The
eight games highest on the temporal axis alone, selected without any outcome, give $5.2$
(Table~\ref{tab:hard8_temporal}), and by temporal tertile the leaderboard spans $1.8$, $3.0$ and $5.0$
chips/game. The top six still overlap within the hard set, and we make no claim about their order. With each
game's margins divided by the standard deviation of its payoff under random play, which puts games with different stakes
on one scale, the ratio is 1.35 $[1.23, 1.50]$ (Appendix~\ref{app:payoff}).

\section{Capability profiles}
\label{sec:profiles}

\begin{figure}[t]
\centering
\includegraphics[width=0.86\textwidth]{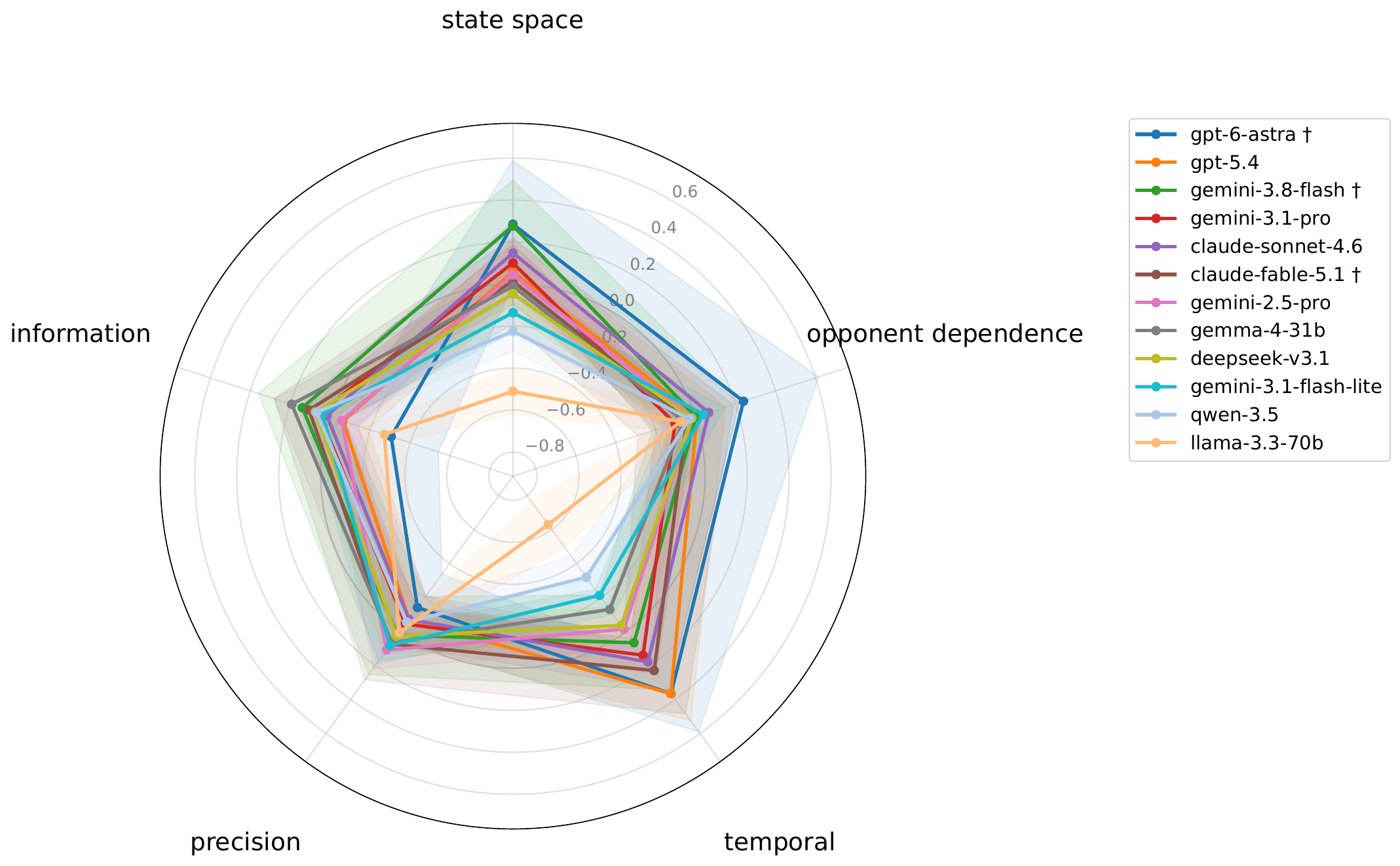}
\caption{\textbf{Capability profile.} For each model, we regress its game strength $\hat\alpha_{m,g}$ on all five
$z$-scored axes jointly, with an intercept, over the 50 games. Each spoke shows the coefficients for one axis,
in chips/game per SD of the axis, with one polygon per model. A model lies outside the zero ring on an axis
when its advantage over the average increases as a game increases in that axis value. Eleven of the 60 cells,
six on temporal, are significant after applying Benjamini--Hochberg correction at $q<0.05$ to clustered
bootstrap $p$-values ($B{=}2{,}000$), and the shaded bands are 95\% CIs. Figure~\ref{fig:profile_intervals}
marks the significant cells, and the coefficients are in Appendix~\ref{app:regression_full}.}
\label{fig:radar}
\end{figure}

Two models with similar average chip margins can derive their advantages from different parts of game space. This section asks, for each model, how its advantage changes along each of the five axes. For each model $m$ across the 50 games, we fit the regression
$\hat\alpha_{m,g} = \beta_{m,0} + \sum_{a=1}^{5}\beta_{m,a} z_a(g) + \varepsilon_{m,g}$. Here, $z_a(g) = (x_a(g) - \bar x_a)/s_a$, where $x_a(g)$ is game $g$'s raw value on axis $a$ and $\bar x_a$ and $s_a$ are its mean and standard deviation respectively, across the 50 games. We note that $\sum_g z_a(g) = 0$ for every axis, so the intercept is $m$'s average advantage, $\hat\beta_{m,0} = \frac{1}{50}\sum_g \hat\alpha_{m,g}$. The coefficient $\hat\beta_{m,a}$ denotes the predicted change in $\hat\alpha_{m,g}$ when $z_a(g)$ increases by one. Intervals use $B{=}2{,}000$ resamples (clustered at the run level), and we apply a Benjamini--Hochberg (BH) correction across the 60 model$\times$axis entries (Figures~\ref{fig:radar} and~\ref{fig:profile_intervals}). Eleven of the 60 slope coefficients are significant, of which six are on the temporal axis, suggesting that the temporal contingency axis is important in understanding the differential capabilities of the different models.

\paragraph{Factor analysis.} We
standardize each model's $\hat\alpha_{m,g}$ across games and take the first factor of the singular value decomposition (Appendix~\ref{app:factor_rank}). The higher a game's score on this factor, the more the
stronger models gain and the weaker models lose. One factor accounts for 38\% of the standardized variance. The score correlates with temporal contingency at $\rho{=}+0.73$. Its correlation with
state space is $+0.65$. After we remove the factor, no residual correlation between two models is significant after Benjamini--Hochberg correction at $q<0.05$ across the 66 model pairs.

\begin{figure}[t]
\centering
\includegraphics[width=0.6\textwidth]{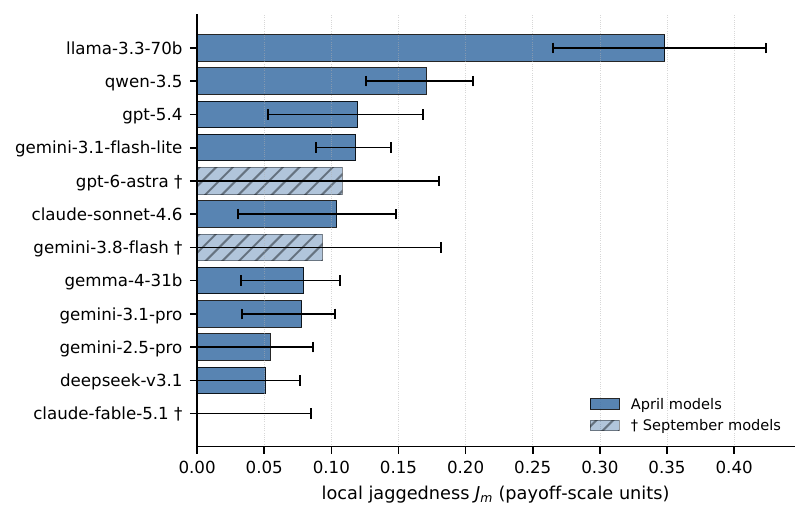}
\caption{\textbf{Local jaggedness.} $J_m$ is the spread of a model's advantage around a smooth function of
the five axes, net of sampling noise. 95\% CIs from a bootstrap over games ($B{=}500$). Hatched bars mark the
September models, whose strength on most games rests on a single matchup.}
\label{fig:jaggedness_main}
\end{figure}

\paragraph{Local jaggedness.}
\label{sec:jaggedness}
A model's strength can vary smoothly with the axes, or it can jump between games that are alike. These jumps can make performance on a new game hard to anticipate. For each model we fit a smooth surface of its strength on
each game over the five axes, where each game is predicted from the other 49. The local jaggedness $J_m$ is the
standard deviation of the residuals after we subtract the sampling variance of the estimates for each game
(Appendix~\ref{app:jaggedness_formal}), in units of each game's payoff scale under random play. \texttt{llama-3.3-70b} ($0.35$) is by far the most jagged.
\texttt{qwen-3.5} follows ($0.17$), and the April frontier models lie between $0.08$ and $0.12$
(Figure~\ref{fig:jaggedness_main}). For \texttt{gpt-6-astra} and \texttt{gpt-5.4} the surface explains about a third of
the variation across games ($R^2 = 0.32$ and $0.35$), mostly their gain on temporally contingent games. The September models met one opponent per game on most games, and their
values are less precise.

\section{Ablations and trace analysis}
\label{sec:programs}\label{sec:ablations}

We run two ablations and one observational analysis of reasoning traces (Appendix \ref{app:ablations}). They test whether improvements in strategic performance come from
knowing the opponent, from more reasoning, or from reasoning that matches the specific challenges of a game.

\paragraph{Disclosed and hidden strategy programs.} We study how tournament ratings fare against opponents
that are not LLMs, and how a model responds to an opponent's strategy it is made aware of. Three author
models (\texttt{claude-sonnet-4.6}, \texttt{gemini-3.1-pro} and \texttt{gemma-4-31b}) wrote complete strategy
programs for both seats of 20 games. Each program is run by the engine and rendered into a procedural reading
that specifies its moves. Five models played the programs on identical deals with the programmed opponent strategy appended to the rulebook (the \emph{disclosed}
arm), and three of them played the programmed opponent whose strategy was hidden (the \emph{hidden} arm). We fit
$\text{margin} = \lambda_{\text{arm}}\hat\alpha_{\text{player}} - \gamma_{\text{group},\text{game},\text{arm}}$,
where $\lambda_{\text{arm}}$ is the slope of margin on tournament rating and $\gamma$ is a fixed effect for each author group, game and arm, excluding each model's own author's programs (Table~\ref{tab:programs_compact} and
Appendix~\ref{app:programs}).

\begin{table}[t]
\centering\small
\caption{\textbf{Programmed opponent ablation.} 520 complete pairs on 20 games, with bootstrap 95\% CIs over deals.
Full table in Appendix~\ref{app:ablations}.}
\label{tab:programs_compact}
\setlength{\tabcolsep}{4.5pt}
\begin{tabular}{lcc}
\toprule
 & \textbf{disclosed} & \textbf{hidden} \\
\midrule
transfer slope $\lambda$ of margin on tournament rating & $1.10$ $[0.76, 1.41]$ & $0.92$ $[0.24, 1.58]$ \\
disclosure effect, disclosed $-$ hidden on the same deals (chips/game) & \multicolumn{2}{c}{$+0.72$ $[0.19, 1.30]$} \\
\midrule
\multicolumn{3}{l}{\emph{Efficiency on the five solved games ($0$ = worst response, $1$ = best response to the program)}} \\
\texttt{gpt-5.4} & $1.02$ $[0.82, 1.28]$ & $0.81$ $[0.70, 0.90]$ \\
\texttt{claude-sonnet-4.6} & $1.01$ $[0.80, 1.25]$ & $0.71$ $[0.53, 0.84]$ \\
\texttt{llama-3.3-70b} & $0.53$ $[0.37, 0.72]$ & $0.53$ $[0.24, 0.79]$ \\
\bottomrule
\end{tabular}

\end{table}

The leaderboard transfers to the case where opponent strategies are disclosed, but we do not find evidence that disclosure helps strong or weak models differentially. Disclosing the program helps, by $+0.72$ chips/game over hidden play on the same deals (95\% CI $[0.19, 1.30]$), with $+0.87$ $[+0.27, +1.53]$ for \texttt{gpt-5.4}, $+0.90$ $[-0.23, +2.08]$ for \texttt{claude-sonnet-4.6} and $+0.45$ $[-0.55, +1.53]$ for \texttt{llama-3.3-70b}, the weakest model. Only the benefit for \texttt{gpt-5.4} is individually significant. Against disclosed programs, a model's margin rises with its tournament rating at $\lambda_{\text{disclosed}}{=}1.10$ $[0.76, 1.41]$, which is consistent with one chip of margin for each chip of rating. Against hidden programs the slope is $\lambda_{\text{hidden}}{=}0.92$ $[0.24, 1.58]$, and the difference of $+0.18$ $[-0.40, +0.75]$ is consistent with disclosure helping strong and weak models equally. We do not test whether a model could work out its opponent over repeated play. Moreover, on five games we solve exactly for the best response to each program, and score efficiency from the worst response ($0$) to the best ($1$). The three frontier models score $0.98$ to $1.02$ (values above 1 are within sampling error of the best response), close to fully exploiting the programmed strategy. Without disclosure, frontier model efficiency drops to $0.81$ $[0.70, 0.90]$ and $0.71$ $[0.53, 0.84]$ (understandably, models cannot fully exploit a strategy they have not been explicitly told about). \texttt{llama-3.3-70b} scores $0.53$ in both arms.

\paragraph{Thinking budget.} We played low and high effort variants of four families (\texttt{gpt-5.4},
\texttt{claude-sonnet-4.6}, \texttt{gemini-3.1-pro}, \texttt{gemini-2.5-pro}) against the same opponents on
identical deals. Within each of the 260 sibling pairs (the two variants of a family on the same deal and opponent), a difference in chip margin can be attributed to the
thinking budget. In every family the high effort variant won more chips, from $+0.06$ chips/game for
\texttt{gemini-2.5-pro} to $+0.53$ for \texttt{gemini-3.1-pro}, but only the \texttt{gemini-3.1-pro} gain is significant.

\paragraph{Reasoning traces.} We use an LLM judge (\texttt{gemini-3-flash-preview}) to score $9{,}778$ move traces on three rubrics (opponent modeling, private information, planning depth). How much of each
type of reasoning a trace contains barely tracks its matched axis. Only opponent modeling does, and weakly. Reasoning that matches
a game's complexities is nonetheless associated with higher margins on that game, for private information and planning depth though not for opponent modeling. Reasoning about private information
is associated with $+1.15$ chips per rubric point more on games in the top tertile of the information axis, and planning depth $+0.93$ more in the top tertile of the temporal axis. These findings hold when the same traces are scored by four judges from three model families (Appendix~\ref{app:judges}).

\section{Perfect play, new games, and model separation}
\label{sec:beyond}

\paragraph{Against perfect play.} We solved 27 of the 50 games and had ten models play both seats of the
same deals against the equilibrium ($1{,}080$ matches, Appendix~\ref{app:solved}). The five frontier models break even
($-0.004$ chips/deal, 95\% CI $[-0.24, +0.28]$), so almost every move is a best response to it, and the
three weakest lose $0.88$ $[0.46, 1.32]$. Scored decision by
decision ($66{,}306$ decisions), the frontier's share of zero-regret moves falls from 98\% on the easiest third of the
games to 93\% on the hardest (we could score only two of its 16 games), and the tail's from 89\% to 69\%. Most frontier models lose ground on the hardest games,
but \texttt{gpt-6-astra} plays at equilibrium quality almost everywhere we can measure: 99.1\% of its decisions have
zero regret, and at least 98.6\% in every hardness third. Models typically do not randomize as equilibrium play does, and the strongest almost always gives the
same answer. Still, a best response that holds their other moves at equilibrium gains only about 0.06 chips/deal
against the top tier and 0.13 against the tail (Appendix~\ref{app:solved}). The hardest games, where models differ most, are too large to solve exactly on one machine, so
there only play between models can rank them.

\begin{figure}[t]
\centering
\includegraphics[width=\linewidth]{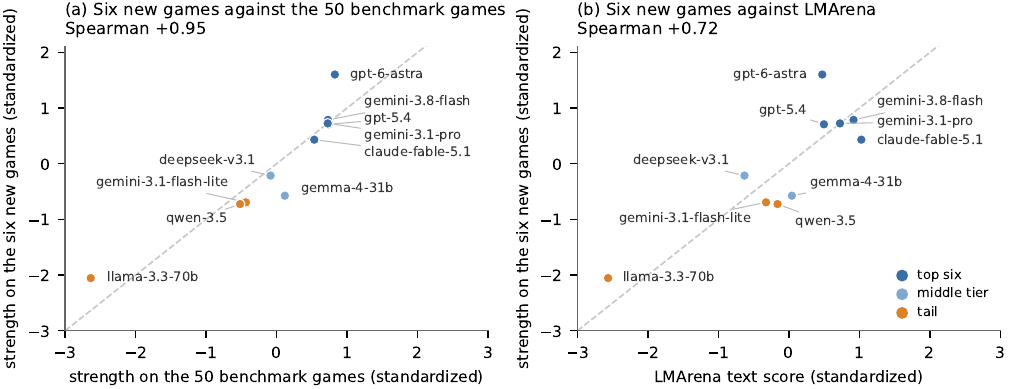}
\caption{\textbf{Strength on the six new games, pooled.} (a) Each model's mean $\hat\alpha_{m,g}/\sigma_U(g)$ over the
six new games against its strength on the 50 benchmark games. (b) The same strength against the model's LMArena text
score. Each axis is standardized over the ten models (mean 0, SD 1), so a model on the dashed line has the same
standing on both. Strength on the benchmark predicts strength on the new games more closely than LMArena does.}
\label{fig:families_scatter}
\end{figure}

\paragraph{New games.} Strengths predicted from the 50 games before any match agree with play on ten fresh games at
Spearman $+0.96$, and strength on ten generated games with no betting agrees at $+0.83$ (Appendix~\ref{app:oos}). The
engine is not limited to betting games: with its other module libraries we built six new games, three card games, a
board game, a market and a negotiation. Each orders ten models like the leaderboard ($\rho$ from $+0.66$ to $+0.88$, \PooledSixRho{} pooled, Figure~\ref{fig:families_scatter}a, Appendix~\ref{app:families}).
The agreement holds after controlling for a general-capability leaderboard: with LMArena held fixed, the
partial rank correlation between the benchmark and the six new games is $+0.90$ $[+0.34, +1.00]$
(Figure~\ref{fig:families_scatter}b, Appendix~\ref{app:oos}). Controlling for the five axes,
a game's distance from Kuhn or Leduc poker does not predict how strongly it reproduces the leaderboard.

\paragraph{Model separation.} The gap between the top models and the weakest grows by $1.89$ chips/game from the
low to the high third of temporal contingency and by $1.73$ with state space, and shrinks with opponent dependence
(Appendix~\ref{app:where}). Weaker models lose mostly on basic hand strength, and frontier models separate from each
other on hard games high in temporal contingency. Models in the same capability tier agree on identical prompts more often than models
from the same lab. Together with the trace and program results, this suggests that in single-deal play the frontier's
lead lies in games whose best action depends on how the deal unfolds, while reasoning about the
opponent does not yet pay off. Single deals without memory leave little room to adapt to an opponent, so opponent
modeling is a target for richer settings.

\section{Limitations and conclusion}
\label{sec:limitations}

We used \textsc{GENSTRAT} to build a benchmark of zero-sum betting games for two players with imperfect information. The six games of
Appendix~\ref{app:families} show that its ranking carries over to other two-player games, including general-sum ones,
but cooperative, multiplayer and agentic settings and play over several deals are left to future work. Within this
scope, the axis results are robust at fifty games, although a larger benchmark would tighten
them. The state space axis is moderately correlated with three of the other four. The
three September models met three to five opponents each, giving them the widest intervals and, on most
games, profiles resting on one matchup. Each module of the generator and its rule text are checked once, composed games are checked automatically,
and people adjudicate only the discrepancies those checks flag (Appendix~\ref{app:validation}). Applied after the
tournament, these checks found one consequential engine bug, already fixed, and wording issues: some rulebooks
leave out details the engine resolves itself, such as whether a prize card is random, and 12 misstate when a
hidden cost is deducted. Reruns on the corrected text and engine moved no model by more than one rank (Appendix~\ref{app:validation}), and since every defect applies to both seats, dropping the affected games leaves the tiers unchanged (Appendix~\ref{app:dataquality}). Matches affected by two earlier bugs, since
fixed, were rerun where possible (Appendix~\ref{app:dataquality}).

We introduced \textsc{GENSTRAT}, which makes evaluation renewable, and allows games of arbitrary complexity to be drawn on demand. On this distribution we found three main results. First, frontier models had a significant advantage in the tournament over older or open-source models. Second, similarly strong models differed in what kinds of games they excelled in, with certain kinds of model gaining more on games with more temporal contingency. Finally, models differed in local jaggedness, the part of a model's advantage that departs from its smooth trend over the axes, with \texttt{llama-3.3-70b} by far the most jagged. In ablations, disclosing an opponent's strategy raised margins by about $0.7$ chips per game, a higher thinking budget raised them in all four families tested, significantly in one, and reasoning matched to the demands of a game was associated with higher margins on that game, for private information and planning depth. On the solved games, frontier models play at the value of the equilibrium. \textsc{GENSTRAT} is a general-purpose, modular engine, and new and harder games can be drawn from a seed when the current ones saturate. The betting games are the engine's first family, and the same engine already produces card games, a board game, a market and a contract negotiation. Adding families puts more interactions between agents with partly aligned interests inside one generated distribution, and the betting leaderboard already predicts strength on the six new games at a rank correlation of 0.95, above a general-capability leaderboard.

\subsection*{AI use statement}
Twelve LLMs played the games in this study. We used LLM agents to develop features of the game generator. We used coding assistants to write statistical analysis code, and to draft manuscript text. Human auditors and LLM agents checked rulebooks for consistency, correctness, and clarity (Appendix~\ref{app:dataquality}). Three LLMs wrote opponent strategy programs used in the program ablation (Section~\ref{sec:programs}, Appendix~\ref{app:programs}). LLM judges scored reasoning traces using a handwritten rubric (Section~\ref{sec:ablations}, Appendix~\ref{app:rubric}).

The authors developed research questions and designed the generator, tournament, and experimental axes. We reviewed all AI-assisted code and regenerated reported numbers. We reviewed and edited all text drafted with AI assistance. Authors take full responsibility for the final paper and its accompanying artifacts.

\subsection*{Ethics statement}
We evaluate commercial and open-source LLMs on generated card games. The study involves no human subjects or personal data. No real money was at stake.

\subsection*{Reproducibility statement}
Each game is determined by its integer seed and its stored generation configuration (Section~\ref{sec:gbg}). Sections~\ref{sec:gbg} and~\ref{sec:construction} describe the evaluation framework in detail. Appendix~\ref{app:axis_formals} defines the five complexity axes. Tournament design details are provided in Section~\ref{sec:setup} and Appendix~\ref{app:dataquality}. Estimators for statistical analyses are made available in Sections~\ref{sec:leaderboard} and~\ref{sec:profiles}. Further details are provided in Appendices~\ref{app:leaderboards}, \ref{app:regression_full}, and~\ref{app:jaggedness_formal}. Each model's provider identifier and request settings are recorded with the data.

\subsection*{Acknowledgements}
We thank Peter Henderson, Zeyu Shen, Vikram Kakaria, Marc Lanctot, and Ian Gemp for valuable discussions, assistance, and
feedback that helped shape this work.

\phantomsection\label{endmain}
{\small
\bibliographystyle{iclr2027_conference}
\bibliography{genstrat}

\begin{thebibliography}{34}
\providecommand{\natexlab}[1]{#1}
\providecommand{\url}[1]{\texttt{#1}}
\expandafter\ifx\csname urlstyle\endcsname\relax
  \providecommand{\doi}[1]{doi: #1}\else
  \providecommand{\doi}{doi: \begingroup \urlstyle{rm}\Url}\fi

\bibitem[Akata et~al.(2025)Akata, Schulz, Coda-Forno, Oh, Bethge, and
  Schulz]{akata2025repeated}
Elif Akata, Lion Schulz, Julian Coda-Forno, Seong~Joon Oh, Matthias Bethge, and
  Eric Schulz.
\newblock {Playing repeated games with large language models}.
\newblock \emph{Nature Human Behaviour}, 9\penalty0 (7):\penalty0 1380--1390,
  2025.

\bibitem[{Anthropic}(2025)]{anthropic2025projectvend1}
{Anthropic}.
\newblock {Project Vend: Can Claude run a small shop? (And why does that
  matter?)}.
\newblock Anthropic Research, 2025.
\newblock URL \url{https://www.anthropic.com/research/project-vend-1}.

\bibitem[{Anthropic}(2026)]{anthropic2026projectdeal}
{Anthropic}.
\newblock {Project {D}eal: our {Claude}-run marketplace experiment}.
\newblock Anthropic, 2026.
\newblock URL \url{https://www.anthropic.com/features/project-deal}.

\bibitem[Benjamini \& Hochberg(1995)Benjamini and
  Hochberg]{benjamini1995controlling}
Yoav Benjamini and Yosef Hochberg.
\newblock {Controlling the False Discovery Rate: A Practical and Powerful
  Approach to Multiple Testing}.
\newblock \emph{Journal of the Royal Statistical Society. Series B
  (Methodological)}, 57\penalty0 (1):\penalty0 289--300, 1995.

\bibitem[Bradley \& Terry(1952)Bradley and Terry]{bradley1952rank}
Ralph~Allan Bradley and Milton~E. Terry.
\newblock {Rank Analysis of Incomplete Block Designs: I. The Method of Paired
  Comparisons}.
\newblock \emph{Biometrika}, 39\penalty0 (3/4):\penalty0 324--345, 1952.

\bibitem[Brown \& Sandholm(2018)Brown and Sandholm]{brown2018libratus}
Noam Brown and Tuomas Sandholm.
\newblock {Superhuman {AI} for heads-up no-limit poker: {Libratus} beats top
  professionals}.
\newblock \emph{Science}, 359\penalty0 (6374):\penalty0 418--424, 2018.

\bibitem[Brown \& Sandholm(2019)Brown and Sandholm]{brown2019pluribus}
Noam Brown and Tuomas Sandholm.
\newblock {Superhuman {AI} for multiplayer poker}.
\newblock \emph{Science}, 365\penalty0 (6456):\penalty0 885--890, 2019.

\bibitem[Buja \& Eyuboglu(1992)Buja and Eyuboglu]{buja1992remarks}
Andreas Buja and Nermin Eyuboglu.
\newblock {Remarks on Parallel Analysis}.
\newblock \emph{Multivariate Behavioral Research}, 27\penalty0 (4):\penalty0
  509--540, 1992.

\bibitem[Chevalier-Boisvert et~al.(2023)Chevalier-Boisvert, Dai, Towers,
  Perez-Vicente, Willems, Lahlou, Pal, Castro, and
  Terry]{chevalier2023minigrid}
Maxime Chevalier-Boisvert, Bolun Dai, Mark Towers, Rodrigo Perez-Vicente, Lucas
  Willems, Salem Lahlou, Suman Pal, Pablo~Samuel Castro, and J~K Terry.
\newblock {Minigrid \& Miniworld: Modular \& Customizable Reinforcement
  Learning Environments for Goal-Oriented Tasks}.
\newblock In \emph{Advances in Neural Information Processing Systems (NeurIPS)
  Datasets and Benchmarks Track}, volume~36, pp.\  73383--73394, 2023.

\bibitem[Cobbe et~al.(2020)Cobbe, Hesse, Hilton, and
  Schulman]{cobbe2020procgen}
Karl Cobbe, Christopher Hesse, Jacob Hilton, and John Schulman.
\newblock {Leveraging Procedural Generation to Benchmark Reinforcement
  Learning}.
\newblock In \emph{International Conference on Machine Learning (ICML)}, volume
  119 of \emph{Proceedings of Machine Learning Research}, pp.\  2048--2056.
  PMLR, 2020.

\bibitem[Collins et~al.(2026)Collins, Zhang, Todd, Ying, da~Costa, Liu, Sharma,
  Weller, Kuperwajs, Wong, Tenenbaum, and Griffiths]{collins2026evalgames}
Katherine~M. Collins, Cedegao~E. Zhang, Graham Todd, Lance Ying, Mauricio~Barba
  da~Costa, Ryan Liu, Prafull Sharma, Adrian Weller, Ionatan Kuperwajs, Lionel
  Wong, Joshua~B. Tenenbaum, and Thomas~L. Griffiths.
\newblock {Evaluating Language Models' Evaluations of Games}.
\newblock In \emph{International Conference on Learning Representations
  (ICLR)}, 2026.

\bibitem[Costarelli et~al.(2024)Costarelli, Allen, Hauksson, Sodunke,
  Hariharan, Cheng, Li, Clymer, and Yadav]{costarelli2024gamebench}
Anthony Costarelli, Mat Allen, Roman Hauksson, Grace Sodunke, Suhas Hariharan,
  Carlson Cheng, Wenjie Li, Joshua Clymer, and Arjun Yadav.
\newblock {GameBench: Evaluating Strategic Reasoning Abilities of LLM Agents},
  2024.
\newblock arXiv:2406.06613v2.

\bibitem[Duan et~al.(2024)Duan, Zhang, Diffenderfer, Kailkhura, Sun,
  Stengel-Eskin, Bansal, Chen, and Xu]{duan2024gtbench}
Jinhao Duan, Renming Zhang, James Diffenderfer, Bhavya Kailkhura, Lichao Sun,
  Elias Stengel-Eskin, Mohit Bansal, Tianlong Chen, and Kaidi Xu.
\newblock {GTBench: Uncovering the Strategic Reasoning Limitations of LLMs via
  Game-Theoretic Evaluations}.
\newblock In \emph{Advances in Neural Information Processing Systems
  (NeurIPS)}, volume~37, pp.\  28219--28253, 2024.

\bibitem[{FAIR} et~al.(2022){FAIR}, Bakhtin, Brown, Dinan, Farina, Flaherty,
  Fried, Goff, Gray, Hu, Jacob, Komeili, Konath, Kwon, Lerer, Lewis, Miller,
  Mitts, Renduchintala, Roller, Rowe, Shi, Spisak, Wei, Wu, Zhang, and
  Zijlstra]{cicero2022}
{FAIR}, Anton Bakhtin, Noam Brown, Emily Dinan, Gabriele Farina, Colin
  Flaherty, Daniel Fried, Andrew Goff, Jonathan Gray, Hengyuan Hu, Athul~Paul
  Jacob, Mojtaba Komeili, Karthik Konath, Minae Kwon, Adam Lerer, Mike Lewis,
  Alexander~H. Miller, Sasha Mitts, Adithya Renduchintala, Stephen Roller, Dirk
  Rowe, Weiyan Shi, Joe Spisak, Alexander Wei, David Wu, Hugh Zhang, and Markus
  Zijlstra.
\newblock {Human-level play in the game of {Diplomacy} by combining language
  models with strategic reasoning}.
\newblock \emph{Science}, 378\penalty0 (6624):\penalty0 1067--1074, 2022.

\bibitem[Fish et~al.(2024)Fish, Gonczarowski, and
  Shorrer]{fish2024algorithmiccollusion}
Sara Fish, Yannai~A Gonczarowski, and Ran~I Shorrer.
\newblock {Algorithmic Collusion by Large Language Models}, 2024.
\newblock arXiv:2404.00806v5, revised 2026.

\bibitem[Guo et~al.(2024)Guo, Yang, Yoo, Lin, Iwasawa, and
  Matsuo]{guo2024suspicion}
Jiaxian Guo, Bo~Yang, Paul Yoo, Bill~Yuchen Lin, Yusuke Iwasawa, and Yutaka
  Matsuo.
\newblock {Suspicion Agent: Playing Imperfect Information Games with Theory of
  Mind Aware {GPT}-4}.
\newblock In \emph{First Conference on Language Modeling (COLM)}, 2024.

\bibitem[Horn(1965)]{horn1965rationale}
John~L. Horn.
\newblock {A Rationale and Test for the Number of Factors in Factor Analysis}.
\newblock \emph{Psychometrika}, 30\penalty0 (2):\penalty0 179--185, 1965.

\bibitem[Huang et~al.(2024)Huang, Cao, Wen, Zhou, and Zhang]{huang2024pokergpt}
Chenghao Huang, Yanbo Cao, Yinlong Wen, Tao Zhou, and Yanru Zhang.
\newblock {PokerGPT: An End-to-End Lightweight Solver for Multi-Player Texas
  Hold'em via Large Language Model}, 2024.
\newblock arXiv:2401.06781.

\bibitem[Kuhn(1950)]{kuhn1950}
Harold~W. Kuhn.
\newblock {A simplified two-person poker}.
\newblock In H.~W. Kuhn and A.~W. Tucker (eds.), \emph{Contributions to the
  Theory of Games, Vol.~I}, number~24 in Annals of Mathematics Studies, pp.\
  97--103. Princeton University Press, 1950.

\bibitem[Li et~al.(2026)Li, Schultz, Lanctot, and Hennes]{li2026modular}
Zun Li, John Schultz, Marc Lanctot, and Daniel Hennes.
\newblock {Modular Discovery of General Game-Playing Algorithms with Large
  Language Models}, 2026.
\newblock arXiv:2609.33115.

\bibitem[Light et~al.(2023)Light, Cai, Shen, and Hu]{light2023avalon}
Jonathan Light, Min Cai, Sheng Shen, and Ziniu Hu.
\newblock {AvalonBench: Evaluating LLMs Playing the Game of Avalon}, 2023.
\newblock arXiv:2310.05036v3.

\bibitem[Lin \& Hou(2026)Lin and Hou]{lin2026readableminds}
Hsieh-Ting Lin and Tsung-Yu Hou.
\newblock {Readable Minds: Emergent Theory-of-Mind-Like Behavior in LLM Poker
  Agents}, 2026.
\newblock arXiv:2604.04157.

\bibitem[Lin et~al.(2026)Lin, Dai, Liu, Tang, Yan, Dai, Zeng, Zhang, Wang, Gao,
  Luo, Zhang, He, and Wang]{lin2026toolpoker}
Minhua Lin, Enyan Dai, Hui Liu, Xianfeng Tang, Yuliang Yan, Zhenwei Dai,
  Jingying Zeng, Zhiwei Zhang, Fali Wang, Hongcheng Gao, Chen Luo, Xiang Zhang,
  Qi~He, and Suhang Wang.
\newblock {How Far Are {LLM}s from Professional Poker Players? Revisiting
  Game-Theoretic Reasoning with Agentic Tool Use}.
\newblock In \emph{International Conference on Learning Representations
  (ICLR)}, 2026.

\bibitem[Lor{\`e} \& Heydari(2024)Lor{\`e} and Heydari]{lore2024strategic}
Nunzio Lor{\`e} and Babak Heydari.
\newblock {Strategic behavior of large language models and the role of game
  structure versus contextual framing}.
\newblock \emph{Scientific Reports}, 14\penalty0 (1):\penalty0 18490, 2024.

\bibitem[Perolat et~al.(2022)Perolat, Vylder, Hennes, Tarassov, Strub, de~Boer,
  Muller, Connor, Burch, Anthony, McAleer, Elie, Cen, Wang, Gruslys, Malysheva,
  Khan, Ozair, Timbers, Pohlen, Eccles, Rowland, Lanctot, Lespiau, Piot,
  Omidshafiei, Lockhart, Sifre, Beauguerlange, Munos, Silver, Singh, Hassabis,
  and Tuyls]{perolat2022deepnash}
Julien Perolat, Bart~De Vylder, Daniel Hennes, Eugene Tarassov, Florian Strub,
  Vincent de~Boer, Paul Muller, Jerome~T. Connor, Neil Burch, Thomas Anthony,
  Stephen McAleer, Romuald Elie, Sarah~H. Cen, Zhe Wang, Audrunas Gruslys,
  Aleksandra Malysheva, Mina Khan, Sherjil Ozair, Finbarr Timbers, Toby Pohlen,
  Tom Eccles, Mark Rowland, Marc Lanctot, Jean-Baptiste Lespiau, Bilal Piot,
  Shayegan Omidshafiei, Edward Lockhart, Laurent Sifre, Nathalie Beauguerlange,
  Remi Munos, David Silver, Satinder Singh, Demis Hassabis, and Karl Tuyls.
\newblock {Mastering the game of Stratego with model-free multiagent
  reinforcement learning}.
\newblock \emph{Science}, 378\penalty0 (6623):\penalty0 990--996, 2022.

\bibitem[Shaker et~al.(2016)Shaker, Togelius, and Nelson]{shaker2016pcg}
Noor Shaker, Julian Togelius, and Mark~J Nelson.
\newblock \emph{{Procedural Content Generation in Games}}.
\newblock Springer, 2016.

\bibitem[Silver et~al.(2018)Silver, Hubert, Schrittwieser, Antonoglou, Lai,
  Guez, Lanctot, Sifre, Kumaran, Graepel, Lillicrap, Simonyan, and
  Hassabis]{silver2018alphazero}
David Silver, Thomas Hubert, Julian Schrittwieser, Ioannis Antonoglou, Matthew
  Lai, Arthur Guez, Marc Lanctot, Laurent Sifre, Dharshan Kumaran, Thore
  Graepel, Timothy Lillicrap, Karen Simonyan, and Demis Hassabis.
\newblock {A general reinforcement learning algorithm that masters chess,
  shogi, and {Go} through self-play}.
\newblock \emph{Science}, 362\penalty0 (6419):\penalty0 1140--1144, 2018.

\bibitem[Southey et~al.(2005)Southey, Bowling, Larson, Piccione, Burch,
  Billings, and Rayner]{southey2005leduc}
Finnegan Southey, Michael Bowling, Bryce Larson, Carmelo Piccione, Neil Burch,
  Darse Billings, and Chris Rayner.
\newblock {Bayes' Bluff: Opponent Modelling in Poker}.
\newblock In \emph{Proceedings of the Twenty-First Conference on Uncertainty in
  Artificial Intelligence (UAI)}, pp.\  550--558, 2005.

\bibitem[Strachan et~al.(2024)Strachan, Albergo, Borghini, Pansardi, Scaliti,
  Gupta, Saxena, Rufo, Panzeri, Manzi, Graziano, and Becchio]{strachan2024tom}
James W.~A. Strachan, Dalila Albergo, Giulia Borghini, Oriana Pansardi, Eugenio
  Scaliti, Saurabh Gupta, Krati Saxena, Alessandro Rufo, Stefano Panzeri, Guido
  Manzi, Michael S.~A. Graziano, and Cristina Becchio.
\newblock {Testing theory of mind in large language models and humans}.
\newblock \emph{Nature Human Behaviour}, 8\penalty0 (7):\penalty0 1285--1295,
  2024.

\bibitem[Ullman(2023)]{ullman2023tom}
Tomer Ullman.
\newblock {Large Language Models Fail on Trivial Alterations to Theory-of-Mind
  Tasks}, 2023.
\newblock arXiv:2302.08399v5.

\bibitem[Vafa et~al.(2024{\natexlab{a}})Vafa, Chen, Rambachan, Kleinberg, and
  Mullainathan]{vafa2024worldmodel}
Keyon Vafa, Justin~Y. Chen, Ashesh Rambachan, Jon Kleinberg, and Sendhil
  Mullainathan.
\newblock {Evaluating the World Model Implicit in a Generative Model}.
\newblock In \emph{Advances in Neural Information Processing Systems
  (NeurIPS)}, volume~37, pp.\  26941--26975, 2024{\natexlab{a}}.

\bibitem[Vafa et~al.(2024{\natexlab{b}})Vafa, Rambachan, and
  Mullainathan]{vafa2024humangeneralization}
Keyon Vafa, Ashesh Rambachan, and Sendhil Mullainathan.
\newblock {Do Large Language Models Perform the Way People Expect? {M}easuring
  the Human Generalization Function}.
\newblock In \emph{International Conference on Machine Learning (ICML)}, volume
  235 of \emph{Proceedings of Machine Learning Research}, pp.\  48919--48937.
  PMLR, 2024{\natexlab{b}}.

\bibitem[Vafa et~al.(2025)Vafa, Chang, Rambachan, and
  Mullainathan]{vafa2025foundationmodel}
Keyon Vafa, Peter~G. Chang, Ashesh Rambachan, and Sendhil Mullainathan.
\newblock {What Has a Foundation Model Found? Using Inductive Bias to Probe for
  World Models}.
\newblock In \emph{International Conference on Machine Learning (ICML)}, 2025.

\bibitem[Verma et~al.(2025)Verma, Huang, Chen, Klein, and
  Tomlin]{verma2025ggbench}
Vivek Verma, David Huang, William Chen, Dan Klein, and Nicholas Tomlin.
\newblock {Measuring General Intelligence with Generated Games}, 2025.
\newblock arXiv:2505.07215.

\end{thebibliography}
}

\clearpage
\raggedbottom
\appendix
\section{Modular game design}
\label{app:design}

Construction of GBGs proceeds via a GBG builder function. The core layer of this builder includes the underlying \texttt{GameState} (players, variables, piles), the \texttt{Rulebook} (containing the phase graph that describes the progression of play), and \texttt{Phase} (ordered lists of actions). A GameState evolves via game operations---events that happen in-game. These include \texttt{Deal}, \texttt{ChipTransfer}, \texttt{Shuffle}, \texttt{PeekAtHand}, \texttt{SwapWithOpponent}, \texttt{StealCardMove}, \texttt{Wager}, and \texttt{ScoreAdjustment}. Each module contains an NL description of its own logic, allowing for automated generation of rulebooks by stitching together individual NL components. Then, in the phase graph, transitions between phases can be conditioned on variables and game outcomes. As a result, games can flow non-linearly via branches, bounded loops, or conditional sub-phases. A complexity parameter $c \in [0,1]$ determines how often complex structural features are drawn. As we increase $c$, the distribution changes gradually from simple games (like Kuhn poker) to more complex games with compound phase graphs. Given a game's seed $s$ and its stored generation configuration, the builder function can build the game $G = \texttt{reconstruct}(s)$ deterministically, making games reproducible.

Natural language rulebooks (Appendix~\ref{app:rulebook_sample}) are generated from the phase graph and provided to the LLM as a system prompt. The natural language version communicates all rules and structure of the game.

Sampled features can combine to produce a game that plays poorly. We address this with deterministic
construction rules and acceptance checks based on $2{,}000$ episodes of random play. We check that
conditions trigger sufficiently often, that phases are generally reachable, and that the two seats are
reasonably balanced under random play. A game is rejected if random play averages more than ten moves per player,
if more than $40\%$ of its conditions hold in fewer than $15\%$ or more than $85\%$ of their evaluations, if more than
$20\%$ of its phases are reached in fewer than $5\%$ of episodes, if either seat wins more than $65\%$ of the episodes
that have a winner, or if a static check of the rulebook finds a branch that can never fire. Of $12{,}351$ seeds tried, $2{,}000$ were accepted.

\section{Sample game rulebook}
\label{app:rulebook_sample}

Compared to the 50 benchmark games, seed 4643 scores high on state space (82nd percentile) and temporal contingency (88th), near the middle on precision (48th) and information (46th) and opponent dependence (44th). The rulebook below is the verbatim system prompt that the LLM agents receive at the start of play. It is generated deterministically from the phase graph of the game.

\begin{tcolorbox}[breakable, colback=gray!4, colframe=gray!50, boxrule=0.5pt,
                  arc=2pt, left=4pt, right=4pt, top=4pt, bottom=4pt]
{\scriptsize
\begin{lstlisting}[style=rulebook]
## Overview
**Players:** Alice, Bob
**Player order:** Player 1 = Alice (acts first in betting), Player 2 = Bob.
**Deck:** 8 cards (10, J, Q, K of clubs and hearts)
**Starting chips:** 10 each (visible to all). Chips move between players' stacks and the pot; the total of all chips (stacks plus pot) is conserved across the game.
**Pot:** starts at 0
**General rules:**
- **Information:** The number of cards remaining in the Deck is public information. Hand sizes are private. Players do not know how many cards opponents hold unless revealed by an action. If the Deck is empty when a draw action occurs, no cards are drawn and the action has no effect. Public draws emit a public skipped-draw notice; private draws emit that notice only to the drawing player — but per the inferability principle, the Deck-size change from any successful draw is publicly visible.
- **Legal actions:** Players may choose any listed option. If its preconditions are unmet (empty hand), the action has no effect.
- **Resolution:** In any multi-step effect, each step resolves in order; if one step's precondition is unmet, that step is skipped and later steps still resolve.
- **Game end:** If the game ends during any phase, all remaining phases are skipped.
- **Costs & pot:** Costs are paid to the pot before the effect resolves, unless otherwise specified. When an action lists a cost: if the player cannot afford the stated cost at the moment they select it, the action is not offered as an option. If the cost is probabilistic (e.g. a 50% chance of paying 1 chip), the player commits to the action first and the cost is resolved afterwards; if they cannot afford the resolved cost, the action is skipped and their choice slot is consumed (they do not get to pick again).
- **Showdown & misc.:** There is no hand size limit; hands grow and shrink freely via draws. Unless stated otherwise, all random selections are uniform. At Showdown, each player's hand is revealed as of the moment the Showdown phase begins, including any cards swapped, drawn, or stolen during earlier phases.
## Visibility
- **Hands**: Each player's cards are private to that player. Opponents cannot see hand contents unless explicitly revealed.
- **Actions**: All player actions (bets, checks, folds) and their effects are publicly announced unless stated otherwise. Actions marked as private are executed silently.
- **Inferability**: "Private" hides *which branch fired, what card was drawn/peeked, or who chose what* — not the observable side-effects. Deck sizes, pot totals, chip stacks, and position holdings are always public and reconcile at settlement. If one private branch would change a public number (e.g. shrink the Deck by 1) and another would not, the opponent can infer which branch ran from that number. Treat the privacy label as concealing the *identity/choice*, not the *occurrence*.
- **Position ownership**: Visibility of position assignments is stated per position (public or hidden).
- **Condition checks**: Checks marked "(public)" are announced to all players when evaluated.
- **Branch taken**: Private conditional blocks do not announce which branch was taken, even if players can infer it.
- **Private phases**: Phases marked *(private)* do not announce their start or end. Individual actions and conditions within them still follow their own visibility rules.
- **Card identity**: Peeked or observed cards are private to the observer unless stated otherwise.
- **Hidden information**: All hidden information is tracked and enforced; players are told only what the rules specify.

## Positions
A player may hold multiple positions simultaneously.
- **Sage** -- only one player can hold this position at a time. Assigning it to a new player removes it from the previous holder. The identity of the holder is hidden from other players.
- **Scout** -- any number of players may hold this position. Holding it is boolean (receiving it again has no additional effect). The identity of the holder is hidden from other players.
- **Sentinel** -- any number of players may hold this position. Holding it is boolean (receiving it again has no additional effect). The identity of the holder is hidden from other players.
Positions are held until explicitly reassigned by a later assignment phase. Assignment predicates, including predicates over mutable state such as chip counts or card counts, are evaluated only at the moment that assignment phase runs, not on every reference. Once assigned, a position persists even if the original predicate later becomes false.
- **Inferability of hidden positions:** effects gated by a hidden position (e.g. chip movements, card draws) may themselves be publicly observable. The assignment is not announced, but opponents may infer the holder from outcomes.

## Sequence of Play
1. **Setup**: shuffle deck, deal cards
2. **Determine Positions**
3. **Ante Phase**: each player antes into the pot
4. **Observations**
5. **Betting Phase**: players may bet, check, fold, or call
6. **Post-Betting Phase**
7. **Showdown**: reveal hands and settle the pot

## Phase Details

### Setup
- Shuffle the Deck.
- Deal 3 cards face-down to each player (Alice, Bob) (or as many as remain if the deck is short). Each player sees only their own cards.

### Determine Positions
- If Alice's Hand has exactly 3 Clubs: Alice receives the position Sentinel. (Only the recipient is notified.)
- If All cards in Bob's Hand are the same suit: Bob receives the position Sentinel. (Only the recipient is notified.)
- If Alice's Hand has exactly 0 Clubs: Alice receives the position Scout. (Only the recipient is notified.)
- If Bob's Hand has at least 2 Clubs: Bob receives the position Scout. (Only the recipient is notified.)

### Ante Phase
- Alice antes 3 chips (moved from Alice's chip stack into the pot).
- Bob antes 3 chips (moved from Bob's chip stack into the pot).

### Observations
- If Alice holds Sentinel: Alice peeks at 1 card in Bob's hand (selected uniformly at random). The card remains in Bob's hand. The peek is private: only Alice sees the card. If Bob's hand is empty, Alice is informed that it is empty.
- If Bob holds Sentinel: Bob peeks at 1 card in Alice's hand (selected uniformly at random). The card remains in Alice's hand. The peek is private: only Bob sees the card. If Alice's hand is empty, Bob is informed that it is empty.
- If Alice holds Scout: Alice peeks at 1 card in Bob's hand (selected uniformly at random). The card remains in Bob's hand. The peek is private: only Alice sees the card. If Bob's hand is empty, Alice is informed that it is empty.
- If Bob holds Scout: Bob peeks at 1 card in Alice's hand (selected uniformly at random). The card remains in Alice's hand. The peek is private: only Bob sees the card. If Alice's hand is empty, Bob is informed that it is empty.

### Betting Phase
- Betting proceeds for up to 2 rounds.
- Each round, Betting Round is played. This continues until a player has 3 or fewer chips, the round cap (2) has been reached, or the game has ended.

#### Sub-phase: Betting Round *(called from Betting Phase)*
- «bet round» starts at 0 and increases by 1 after each round.
- If «bet round» is greater than 0 and the game is still in progress (public): Perform **Pre-Betting Actions** steps (both players participate per this sub-phase's own rules), then continue.
- If «bet round» is greater than 0 and the game is still in progress (public): Perform **Inter-Round Draw (All Players, inter-round event 1)** steps, then continue.
- If «bet round» is greater than 0 and the game is still in progress: Perform **Assign Position: Sage** steps, then continue.
- If «bet round» is greater than 0 and the game is still in progress: Perform **Dynamic Role Effects** steps, then continue.
- If «bet round» is equal to 0 (public):
    - If Alice has chips remaining and Bob has chips remaining (public):
        Round 1: variable betting, wager any amount from 1 chip up to your remaining stack.
        Then, perform **Variable Betting Round (Round 1)** steps, then continue.
    - Otherwise:
        Round 1: variable betting unaffordable, falling back to fixed 1-chip betting for this round.
        Then, Alice picks one of: bet or check. (choice is announced to all players)
        - If Alice chooses 'check' (public): Perform **Bob: Bet or Check (Round 1, Fallback)** steps, then continue.
        - Otherwise, if Alice chooses 'bet':
            Alice bets 1 chip (moved from Alice's chip stack into the pot).
            Then, perform **Bob: Fold or Call (Round 1, Fallback)** steps, then continue.
  - Otherwise:
    Round 2: fixed betting, each bet or call is 1 chip.
    Then, Alice picks one of: bet or check. (choice is announced to all players)
    - If Alice chooses 'check' (public): Perform **Bob: Bet or Check (Round 2)** steps, then continue.
    - Otherwise, if Alice chooses 'bet':
        Alice bets 1 chip (moved from Alice's chip stack into the pot).
        Then, perform **Bob: Fold or Call (Round 2)** steps, then continue.
*(Play returns to **Betting Phase**.)*

#### Sub-phase: Pre-Betting Actions *(called from Betting Round)*
- Alice picks one of: Peek at opponent's hand (50% chance to cost 1 chip) or Pass. (the choice itself is not announced; any public side-effect a branch happens to produce — e.g. chip or card movements — can still reveal which branch ran)
  - If Alice chooses 'Pass': Alice does nothing.
  - Otherwise, if Alice chooses 'Peek at opponent's hand (50% chance to cost 1 chip)': Alice peeks at 1 card in Bob's hand (selected uniformly at random). The card remains in Bob's hand. The peek is private: only Alice sees the card. If Bob's hand is empty, Alice is informed that it is empty. Has a 50% chance of costing 1 chip (paid to the pot); otherwise free.
- Bob picks one of: Peek at opponent's hand (50% chance to cost 1 chip) or Pass. (the choice itself is not announced; any public side-effect a branch happens to produce — e.g. chip or card movements — can still reveal which branch ran)
  - If Bob chooses 'Pass': Bob does nothing.
  - Otherwise, if Bob chooses 'Peek at opponent's hand (50% chance to cost 1 chip)': Bob peeks at 1 card in Alice's hand (selected uniformly at random). The card remains in Alice's hand. The peek is private: only Bob sees the card. If Alice's hand is empty, Bob is informed that it is empty. Has a 50% chance of costing 1 chip (paid to the pot); otherwise free.
*(Play returns to **Betting Round**.)*

#### Sub-phase: Inter-Round Draw (All Players, inter-round event 1) *(called from Betting Round)*
- If the Deck has at least 2 cards (public): Inter-round draw: each player draws 1 card from the Deck (private).
  - Otherwise: Inter-round draw: the Deck has fewer than 2 cards remaining; the both-players draw is skipped.
*(Play returns to **Betting Round**.)*

#### Sub-phase: Assign Position: Sage *(private)* *(called from Betting Round)*
*Players are not told when this phase begins or ends.*
- If Alice has more chips than Bob: Alice receives the position Sage. (Only the recipient is notified.)
  - Otherwise, if Bob has more chips than Alice: Bob receives the position Sage. (Only the recipient is notified.)
  - Otherwise (tied): Coin flip -- heads: Alice receives the position Sage. (Only the recipient is notified.); tails: Bob receives the position Sage. (Only the recipient is notified.).
    *(The branch condition is evaluated privately; which branch fires is not announced directly.)*
*(Play returns to **Betting Round**.)*

#### Sub-phase: Dynamic Role Effects *(private)* *(called from Betting Round)*
*Players are not told when this phase begins or ends.*
- If Alice holds Sage: Alice peeks at 1 card in Bob's hand (selected uniformly at random). The card remains in Bob's hand. The peek is private: only Alice sees the card. If Bob's hand is empty, Alice is informed that it is empty.
- If Bob holds Sage: Bob peeks at 1 card in Alice's hand (selected uniformly at random). The card remains in Alice's hand. The peek is private: only Bob sees the card. If Alice's hand is empty, Bob is informed that it is empty.
*(Play returns to **Betting Round**.)*

#### Sub-phase: Bob: Fold or Call (Round 2) *(called from Betting Round)*
- Bob picks one of: fold or call. (choice is announced to all players)
  - If Bob chooses 'call' (public): Bob bets 1 chip (moved from Bob's chip stack into the pot).
  - Otherwise, if Bob chooses 'fold':
      The entire pot is given to Alice.
      Then, the game ends: Alice wins after Bob folds.

#### Sub-phase: Bob: Bet or Check (Round 2) *(called from Betting Round)*
- Bob picks one of: bet or check. (choice is announced to all players)
  - If Bob chooses 'check' (public): No action.
  - Otherwise, if Bob chooses 'bet':
      Bob bets 1 chip (moved from Bob's chip stack into the pot).
      Then, perform **Alice: Fold or Call (responding, Round 2)** steps, then continue.
*(Play returns to **Betting Round**.)*

#### Sub-phase: Alice: Fold or Call (responding, Round 2) *(called from Bob: Bet or Check (Round 2))*
- Alice picks one of: fold or call. (choice is announced to all players)
  - If Alice chooses 'call' (public): Alice bets 1 chip (moved from Alice's chip stack into the pot).
  - Otherwise, if Alice chooses 'fold':
      The entire pot is given to Bob.
      Then, the game ends: Bob wins after Alice folds.

#### Sub-phase: Bob: Fold or Call (Round 1, Fallback) *(called from Betting Round)*
- Bob picks one of: fold or call. (choice is announced to all players)
  - If Bob chooses 'call' (public): Bob bets 1 chip (moved from Bob's chip stack into the pot).
  - Otherwise, if Bob chooses 'fold':
      The entire pot is given to Alice.
      Then, the game ends: Alice wins after Bob folds.

#### Sub-phase: Bob: Bet or Check (Round 1, Fallback) *(called from Betting Round)*
- Bob picks one of: bet or check. (choice is announced to all players)
  - If Bob chooses 'check' (public): No action.
  - Otherwise, if Bob chooses 'bet':
      Bob bets 1 chip (moved from Bob's chip stack into the pot).
      Then, perform **Alice: Fold or Call (responding, Round 1, Fallback)** steps, then continue.
*(Play returns to **Betting Round**.)*

#### Sub-phase: Alice: Fold or Call (responding, Round 1, Fallback) *(called from Bob: Bet or Check (Round 1, Fallback))*
- Alice picks one of: fold or call. (choice is announced to all players)
  - If Alice chooses 'call' (public): Alice bets 1 chip (moved from Alice's chip stack into the pot).
  - Otherwise, if Alice chooses 'fold':
      The entire pot is given to Bob.
      Then, the game ends: Bob wins after Alice folds.

#### Sub-phase: Variable Betting Round (Round 1) *(called from Betting Round)*
- Alice and Bob may each bet any whole number of chips (minimum 1). When a player bets, their chosen amount is transferred from their chip stack into the pot immediately. When a player calls, their matched amount is transferred to the pot; if they cannot match the full bet, they go all-in with their remaining chips and the excess from the original bettor is returned to that bettor. If both players check in sequence, the round ends with no additional chips moving.
- Alice picks one of: bet or check. (choice is announced to all players)
  - If Alice chooses 'check' (public): Perform **Bob: Bet or Check (Round 1)** steps, then continue.
  - Otherwise, if Alice chooses 'bet':
      Alice chooses a wager amount (minimum 1 chip, up to Alice's current chip count). When Alice commits the chosen amount, it is transferred from Alice's chip stack into the pot via the normal public chip announcement. If the submitted amount exceeds the legal maximum, the bid is clamped to the maximum (all-in or declared cap). If the submitted amount is malformed or below the minimum, the engine substitutes a uniform-random legal amount. In either case the submitting player is notified privately and play continues with the resolved amount.
      Then, perform **Bob: Fold or Call (Round 1)** steps, then continue.
*(Play returns to **Betting Round**.)*

#### Sub-phase: Bob: Fold or Call (Round 1) *(called from Variable Betting Round (Round 1))*
- Bob picks one of: fold or call. (choice is announced to all players)
  - If Bob chooses 'call' (public): Bob matches the bet. If they cannot match the full amount, they go all-in and only the matched portion from each player remains in the pot.
  - Otherwise, if Bob chooses 'fold':
      The entire pot is given to Alice.
      Then, the game ends: Alice wins after Bob folds.

#### Sub-phase: Bob: Bet or Check (Round 1) *(called from Variable Betting Round (Round 1))*
- Bob picks one of: bet or check. (choice is announced to all players)
  - If Bob chooses 'check' (public): No action.
  - Otherwise, if Bob chooses 'bet':
      Bob chooses a wager amount (minimum 1 chip, up to Bob's current chip count). When Bob commits the chosen amount, it is transferred from Bob's chip stack into the pot via the normal public chip announcement. If the submitted amount exceeds the legal maximum, the bid is clamped to the maximum (all-in or declared cap). If the submitted amount is malformed or below the minimum, the engine substitutes a uniform-random legal amount. In either case the submitting player is notified privately and play continues with the resolved amount.
      Then, perform **Alice: Fold or Call (responding, Round 1)** steps, then continue.
*(Play returns to **Variable Betting Round (Round 1)**.)*

#### Sub-phase: Alice: Fold or Call (responding, Round 1) *(called from Bob: Bet or Check (Round 1))*
- Alice picks one of: fold or call. (choice is announced to all players)
  - If Alice chooses 'call' (public): Alice matches the bet. If they cannot match the full amount, they go all-in and only the matched portion from each player remains in the pot.
  - Otherwise, if Alice chooses 'fold':
      The entire pot is given to Bob.
      Then, the game ends: Bob wins after Alice folds.

### Post-Betting Phase
- Post-betting events may fire based on game state. Each of the following conditions is evaluated top-to-bottom at its own step; every condition whose check passes at that moment performs its associated action. The conditions are not mutually exclusive: more than one may fire in the same phase.
- If the current pot is more than 3 chips (public): Perform **Observation Round (post-betting)** steps, then continue.
- If at least one player has two cards of the same rank in their hand (public): Perform **Simultaneous Round (post-betting)** steps (both players participate per this sub-phase's own rules), then continue.
- If Alice's hand contains a King (the deck's top rank) (public): Perform **Post-Betting Actions** steps (both players participate per this sub-phase's own rules), then continue.

#### Sub-phase: Observation Round (post-betting) *(private)* *(called from Post-Betting Phase)*
*Players are not told when this phase begins or ends.*
- This observation round is private: non-observers are not notified that any observation occurs.
- Sample 1 card(s) from the current contents of the Deck with replacement (i.e., the Deck is unchanged). If the Deck is empty, 0 cards are observed. The holder(s) of the Sentinel position (if any) privately observe the sampled cards. Sampled cards are not kept. Only the observer sees which cards were sampled. If no player holds the designated position(s) at this point, the observation is skipped.
- Sample 1 card(s) from the current contents of the Deck with replacement (i.e., the Deck is unchanged). If the Deck is empty, 0 cards are observed. The holder(s) of the Sage position (if any) privately observe the sampled cards. Sampled cards are not kept. Only the observer sees which cards were sampled. If no player holds the designated position(s) at this point, the observation is skipped.
- If Alice holds Sentinel: Sample 1 card(s) from the current contents of Bob's Hand with replacement (i.e., Bob's Hand is unchanged). If Bob's Hand is empty, 0 cards are observed. Alice privately observes the sampled cards. Sampled cards are not kept. Only the observer sees which cards were sampled.
  - Otherwise: No effect.
    *(The branch condition is evaluated privately; which branch fires is not announced directly.)*
- If Bob holds Sentinel: Sample 1 card(s) from the current contents of Alice's Hand with replacement (i.e., Alice's Hand is unchanged). If Alice's Hand is empty, 0 cards are observed. Bob privately observes the sampled cards. Sampled cards are not kept. Only the observer sees which cards were sampled.
  - Otherwise: No effect.
    *(The branch condition is evaluated privately; which branch fires is not announced directly.)*
*(Play returns to **Post-Betting Phase**.)*

#### Sub-phase: Simultaneous Round (post-betting) *(called from Post-Betting Phase)*
- All players choose simultaneously and in secret. No player knows the others' choices until all are locked in. Once all choices are submitted, each player's choice is publicly revealed, then the combined outcome is resolved and announced publicly. If a player fails to choose, a uniformly random option is selected for them.
- Alice chooses one of the following options: Press, Hold, or Bluff.
- Bob chooses one of the following options: Press, Hold, or Bluff.
- Outcomes based on choices:
  - If Alice chooses 'Press' AND Bob chooses 'Hold':
      - If Bob holds Sage: Nothing occurs.
      - Otherwise: Bob pays 1 chip to Alice. If Bob has fewer than 1 chip, only the available amount is transferred.
          *(The branch condition is evaluated privately; which branch fires is not announced directly.)*.
          *(Note: different branches produce different observable chip effects; the branch taken may be partially inferable from chip changes.)*.
  - If Alice chooses 'Hold' AND Bob chooses 'Hold':
      - If Bob holds Sage: Nothing occurs.
      - Otherwise: Bob pays 1 chip to Alice. If Bob has fewer than 1 chip, only the available amount is transferred.
          *(The branch condition is evaluated privately; which branch fires is not announced directly.)*.
          *(Note: different branches produce different observable chip effects; the branch taken may be partially inferable from chip changes.)*.
  - If Alice chooses 'Hold' AND Bob chooses 'Bluff':
      - If Bob holds Sage: Nothing occurs.
      - Otherwise: Bob pays 1 chip to Alice. If Bob has fewer than 1 chip, only the available amount is transferred.
          *(The branch condition is evaluated privately; which branch fires is not announced directly.)*.
          *(Note: different branches produce different observable chip effects; the branch taken may be partially inferable from chip changes.)*.
  - If Alice chooses 'Press' AND Bob chooses 'Bluff':
      - If Alice's highest-ranked card has a higher rank than Bob's highest-ranked card:
          Alice draws 1 card from the Deck. Alice sees the card identity; the Deck-size decrement is public, signaling to both players that a draw occurred.
          Then, Bob pays 2 chips to Alice. If Bob has fewer than 2 chips, only the available amount is transferred.
      - Otherwise:
          Alice draws 1 card from the Deck. Alice sees the card identity; the Deck-size decrement is public, signaling to both players that a draw occurred.
          Then, Alice pays 2 chips to Bob. If Alice has fewer than 2 chips, only the available amount is transferred.
          *(The branch condition is evaluated privately; which branch fires is not announced directly.)*.
          *(Note: different branches produce different observable chip effects; the branch taken may be partially inferable from chip changes.)*.
  - If Alice chooses 'Hold' AND Bob chooses 'Press':
      Alice draws 1 card from the Deck. Alice sees the card identity; the Deck-size decrement is public, signaling to both players that a draw occurred.
      Then, Alice pays 1 chip to Bob. If Alice has fewer than 1 chip, only the available amount is transferred.
  - If Alice chooses 'Bluff' AND Bob chooses 'Hold':
      Alice draws 1 card from the Deck. Alice sees the card identity; the Deck-size decrement is public, signaling to both players that a draw occurred.
      Then, Alice pays 1 chip to Bob. If Alice has fewer than 1 chip, only the available amount is transferred.
  - For all other combinations (Press/Press, Bluff/Press, Bluff/Bluff):
      Alice pays 1 chip to Bob. If Alice has fewer than 1 chip, only the available amount is transferred.
      Then, Bob draws 1 card from the Deck. Bob sees the card identity; the Deck-size decrement is public, signaling to both players that a draw occurred
*(Play returns to **Post-Betting Phase**.)*

#### Sub-phase: Post-Betting Actions *(called from Post-Betting Phase)*
- Alice picks one of: Swap with opponent or Pass. (the choice itself is not announced, but some branches produce public side-effects — chip movements, card exchanges, or other observable state changes — that reveal which branch ran; other branches leave no public trace)
  - If Alice chooses 'Pass': Alice does nothing.
  - Otherwise, if Alice chooses 'Swap with opponent': Alice swaps one random card from their hand with one random card from Bob's hand. In a two-player game both participants learn each other's exchanged card, so after the swap the identities of the two exchanged cards are known to every player. If either hand is empty, nothing happens.
- Bob picks one of: Swap with opponent or Pass. (the choice itself is not announced, but some branches produce public side-effects — chip movements, card exchanges, or other observable state changes — that reveal which branch ran; other branches leave no public trace)
  - If Bob chooses 'Pass': Bob does nothing.
  - Otherwise, if Bob chooses 'Swap with opponent': Bob swaps one random card from their hand with one random card from Alice's hand. In a two-player game both participants learn each other's exchanged card, so after the swap the identities of the two exchanged cards are known to every player. If either hand is empty, nothing happens.
*(Play returns to **Post-Betting Phase**.)*

### Showdown
- Both players' hands are revealed to all players. Showdown: each player's evaluated hand is scored. The base score is the sum of the rank values of all cards in the evaluated hand. (An empty hand scores 0.) GOOD PAIRS: A pair of Js adds 3 points. POISON CARDS: Ranks K subtract 3 points each. Higher adjusted score wins. (Rank values: 10 is face value, J=11, Q=12, K=13.)
- The showdown winner takes the entire pot. On a tie, each player receives half the pot (rounded down); any remaining odd chip is awarded uniformly at random to one tied player.
- The game ends: Showdown complete.
\end{lstlisting}
}
\end{tcolorbox}

\paragraph{Per-turn observation prompt.} Before each turn, an agent receives a message from the engine. The game state is presented first and covers the current phase, the agent's own hand, public cards on the board, each player's chip stack, the current pot, role or position assignments, and who the opponents are. The engine also sends public events and private observations that, by the rules, should be visible to the agent. The final part of the message specifies the decision to be made. The agent must reply with a JSON-formatted response of the form \verb|{"action": <chosen_option>}|. If a response cannot be successfully parsed, the engine plays a uniformly random legal action.

\section{Complexity axes (formal)}
\label{app:axis_formals}

We compute five axis scores for each of the 50 benchmark games without any LLM play. Four of them are
\emph{value axes}, which are expectations over the decisions a player faces when simple policies play the
game. The fifth counts the information states a player can reach.

\paragraph{Objects.} Each axis is computed for one seat $i$ (Alice or Bob) at a time and then averaged over
both. A \emph{deal} is one play of the game from the shuffle to settlement, and the \emph{chip lead} of a seat
is its change in chips minus that of its opponent. In seed 4643 both seats ante 3 of their 10 chips. If Alice
bets at her first decision and Bob folds, she ends 3 chips up and Bob 3 chips down, and her chip lead is 6.
Three axes are divided by $\mathrm{STD}_g$, the standard deviation of the chip lead in a deal between two
members of the population defined below, averaged over random pairs of members. At a decision, seat $i$
observes its own cards, the public cards, the chip stacks, the pot, the phase, and the history of announced
actions. We call this its \emph{information state} $I$. A \emph{policy} maps each $I$ to a distribution over
the set $\mathcal A(I)$ of legal actions at $I$, and $A=|\mathcal A(I)|$. When a player chooses an amount from more
than six values, $\mathcal A(I)$ keeps at most six of them: the four smallest, the midpoint and the largest.

\paragraph{Population.} We drew 200 random behavior strategies. Strategy $j$ has one mixture over the legal
actions for each information state, drawn once from a Dirichlet(1) distribution. We ranked the 200 strategies
by mean chip lead against $L_0$, the uniform random policy, over 40 deals, and the better half (100
strategies) became \emph{tier~1}. Seat $i$ faces an opponent drawn from $P_\Theta$, with weight $1/3$ on $L_0$
and $2/3$ on tier~1. For its own later decisions we draw a policy from $P_K$, with $1/4$ on $L_0$ and $3/4$ on
tier~1.

\paragraph{Decision distribution.} We let both seats play policies drawn from the population and record every
decision of seat $i$ with two or more legal actions. $\mu_i$ is the distribution of the information states at
these decisions, with each visit counted once. A state that play reaches twice as often has twice the weight,
and forced moves have none.

\paragraph{Action values.} The temporal and opponent axes come from one decomposition of action values. How
the deal ends after an action $a$ at an information state $I$ of seat $i$ depends on a \emph{context} $c=(k,\theta,w)$. Here the continuation $k\sim P_K$ makes the later decisions of
seat $i$, and $\theta\sim P_\Theta$ is the opponent in the other seat. The world $w$ assigns every card seat $i$
has not seen (the opponent's hand and the undealt deck), uniformly over the assignments consistent with $I$.
We write $Q(I,a,c)$ for the expected chip lead of seat $i$ at the end of the deal when it takes $a$ at $I$ and
play continues under $c$. Its means are $Q(I,a)=\mathbb E_c\,Q(I,a,c)$, $\bar Q(I,c)=\frac1A\sum_a Q(I,a,c)$,
and the grand mean $\bar Q(I)$. Over a uniformly chosen action and a random context, the variance of
$Q(I,a,c)$ splits exactly into three terms,
\[
\begin{aligned}
V_{\text{total}}(I)\;=\;&\underbrace{\tfrac1A\textstyle\sum_a\big(Q(I,a)-\bar Q(I)\big)^2}_{V_a(I)}
\;+\;\underbrace{\mathbb E_c\big(\bar Q(I,c)-\bar Q(I)\big)^2}_{V_c(I)}\\
&\;+\;\underbrace{\mathbb E_c\,\tfrac1A\textstyle\sum_a\big(Q(I,a,c)-Q(I,a)-\bar Q(I,c)+\bar Q(I)\big)^2}_{V_{a\times c}(I)}.
\end{aligned}
\]
$V_c$ is mostly the luck of the cards. An action that is simply better on average raises $Q(I,a)$, which
enters $V_a$. Note that with two actions and gap $D(c)=Q(I,1,c)-Q(I,2,c)$, the interaction is one quarter of
the variance of the gap across futures. For the opponent, with $Q(I,a,\theta)=\mathbb E_{k,w}\,Q(I,a,c)$,
\[
V_{a\times\theta}(I)\;=\;\mathbb E_\theta\,\tfrac1A\textstyle\sum_a\big(Q(I,a,\theta)-Q(I,a)-\bar Q(I,\theta)+\bar Q(I)\big)^2 ,
\]
and $V_{a\times k}$ and $V_{a\times w}$ are defined in the same way, each after $Q$ is averaged over the other
two parts of the context.

\paragraph{State space.} $\log_{10}\big(\sum_p |\mathcal I_p|\big)$, where $\mathcal I_p$ is the set of
information states that player $p$ reaches in 3{,}000 deals in which both players move uniformly at random over all
legal actions, every legal amount included. A
state is identified by the decision, the player's own hand, its last three decisions, its chips (rounded), everything
it has been shown (the events, the messages, the board and the pot) and the positions. The count is a lower
bound on the number of reachable states.

\paragraph{Temporal contingency (``temporal'').}
\[
\mathrm{Temp}(g)\;=\;\tfrac12\textstyle\sum_{i}\;\mathbb E_{I\sim\mu_i}\!\left[\dfrac{V_{a\times c}(I)}{V_{\text{total}}(I)}\right]
\]
averages, over the decisions of each seat, the fraction of $V_{\text{total}}(I)$ that is driven by interaction between actions and context. The fraction is $0$ when every difference between action values is the same in every context, and $1$ when all of $V_{\text{total}}(I)$ lies in the interaction. The world $w$ in the context includes the opponent's current hand. As a result, a high score can come from the
unseen cards as well as from later play under $k$ and $\theta$. We call this temporal contingency, although it does not separate planning from these other sources. For example, a game of one move with actions $\pm1$, a fair coin $w=\pm1$ that the player cannot
see, and payoff $aw$ has $V_a=V_c=0$ and scores one.

\paragraph{Opponent dependence.}
\[
\mathrm{Opp}(g)\;=\;\tfrac12\textstyle\sum_{i}\;\mathbb E_{I\sim\mu_i}\!\Big[\sqrt{V_{a\times\theta}(I)}\,\Big]\Big/\mathrm{STD}_g .
\]
Here $\sqrt{V_{a\times\theta}(I)}$ is the typical size of the action$\times$opponent interaction, and it is zero
when the gap between two actions is the same against every opponent. We estimate $V_{a\times\theta}(I)$ as the mean
product of the interaction residuals from the two halves of the sampled worlds, clipped at zero. We use its mean in units of
$\mathrm{STD}_g$ as the measure of how much the comparison between actions depends on the opponent. Note that
it does not measure the value of adapting to a known opponent. We also computed that value, as the gain from
the action best against $\theta$ over the action best against the whole population, averaged over $\theta$, and
it lies within $\pm0.05\,\mathrm{STD}_g$ on every game.

A version of the axis defined on whole strategies gives the same result. For each game we played $L_0$
and 40 tier-1 members of each seat against each other on 60 common deals. Opponent dependence is then measured as the share of
outcome variance in the interaction between the strategies of the two seats, cross-fitted over two halves
of the deals. Its median over games is $0.06\%$, and it correlates with the axis at $\rho=0.74$. With this
measure in place of the axis, the capability profile of Section~\ref{sec:profiles} has eight
significant cells, five of them on temporal and none on opponent dependence.

\paragraph{Precision.}
\[
\mathrm{Prec}(g)\;=\;\tfrac12\textstyle\sum_{i}\;\mathbb E_{I\sim\mu_i}\Big[\,Q(I,a^*)-\tfrac{1}{|N|}\textstyle\sum_{a'\in N}Q(I,a')\Big]\Big/\mathrm{STD}_g ,
\]
where $a^*(I)=\arg\max_a Q(I,a)$ is the best action and $N(I)$ is the set of its neighbors. On ordered
menus, such as wagers and bids, the neighbors are the adjacent amounts. On categorical menus they are all
other actions. The value of a mixed strategy is linear in its probabilities, so the gap is the loss per
unit of probability moved from $a^*$ to its neighbors. On a menu of two actions it is the margin of the
runner-up.

\paragraph{Information.} $\mathrm{Lead}_i(\pi,\theta)$ is the expected chip lead of seat $i$ at the end of a
deal when it plays policy $\pi$ against opponent $\theta$. $\mathrm{BR}(\theta)$ is a limited-depth search policy.
At each decision it samples worlds consistent with its observations, values every legal action by replays in
which the opponent plays $\theta$, and takes the action with the best mean. Inside a replay, each of its later decisions is
chosen by a one-level search that values every legal action by two replays in which it then plays uniformly, while
the opponent keeps playing $\theta$. For $\mathrm{BR}_{\mathrm{pub}}(\theta)$ the
sampled worlds also redraw its own hand, and everything else it has observed stays as it was. The axis is
\[
\mathrm{Info}(g)\;=\;\tfrac12\textstyle\sum_{i}\;\mathbb E_{\theta}\Big[\mathrm{Lead}_i\big(\mathrm{BR}(\theta),\theta\big)
-\mathrm{Lead}_i\big(\mathrm{BR}_{\mathrm{pub}}(\theta),\theta\big)\Big]\Big/\mathrm{STD}_g ,
\]
with $\theta$ drawn with equal weight from $L_0$ and four tier-1 members. It is the loss per deal, in units of
$\mathrm{STD}_g$, of a search policy whose sampled worlds do not keep its own cards.

\paragraph{Estimation.} We estimate every expectation by simulating play. Up to 90 decisions per seat are
taken from population self-play. At each decision, all legal actions are replayed under the same 360 sampled
contexts with common random numbers (1{,}080 replays for three legal actions). If an axis needs a
best action, we choose it on one half of the worlds and value it on the other half, which mitigates the effect of selection bias (an action chosen because it did well on some draws should not be evaluated on those same draws). The information axis uses
40 deals against each opponent, and a three-hour cap on computation left fewer opponents for some seats and games. The 360 contexts are all combinations of 4 continuations, 9 opponents, and 10 worlds, and each is replayed once for every action. Table~\ref{tab:axis_diagnostics} reports, for the temporal, opponent and precision axes,
the sampling error, the seat agreement, and the correlation with a rerun at about two-thirds of the budget on an independent
population draw. The variance inflation factors of all five axes are in Appendix~\ref{app:vif}.

\begin{table}[H]
\centering\small
\caption{Diagnostics of the temporal, opponent and precision axes on the 50 benchmark games.}
\label{tab:axis_diagnostics}
\begin{tabular}{lccc}
\toprule
axis & SE / between-game SD & seat agreement ($\rho$) & reproduces across draws \\
\midrule
temporal & 0.21 & $+0.69$ & $\rho=0.92$ \\
opponent dependence & 0.27 & $+0.66$ & $0.88$ \\
precision & 0.31 & $+0.44$ & $0.86$ \\
\bottomrule
\end{tabular}
\end{table}

\paragraph{Selection basis.} Before computing the five axes, we selected the 50 games by farthest-point
sampling on six preliminary axes, which a cheaper screening run computed for all 2{,}000 games in the pool.
These axes are state space, temporal depth, information sensitivity, opponent modeling, risk, and
brittleness. They are different quantities from our five axes even where the names are
similar (the state space of the screening run did not count decisions in which a player chooses an amount, and it
did not tell apart the cards a player had seen), and
thus were only used for selecting games. The two sets of axes do agree on which games are
hard. Their composite complexity scores (Appendix~\ref{app:tertile_leaderboards}) correlate at $\rho=0.79$,
and seven of the eight hardest games are the same.

\section{Benchmark characteristics}
\label{app:benchmark}\label{app:vif}\label{app:axis_dist}\label{app:fps_scatter}

\begin{figure}[H]
\centering
\includegraphics[width=0.7\linewidth]{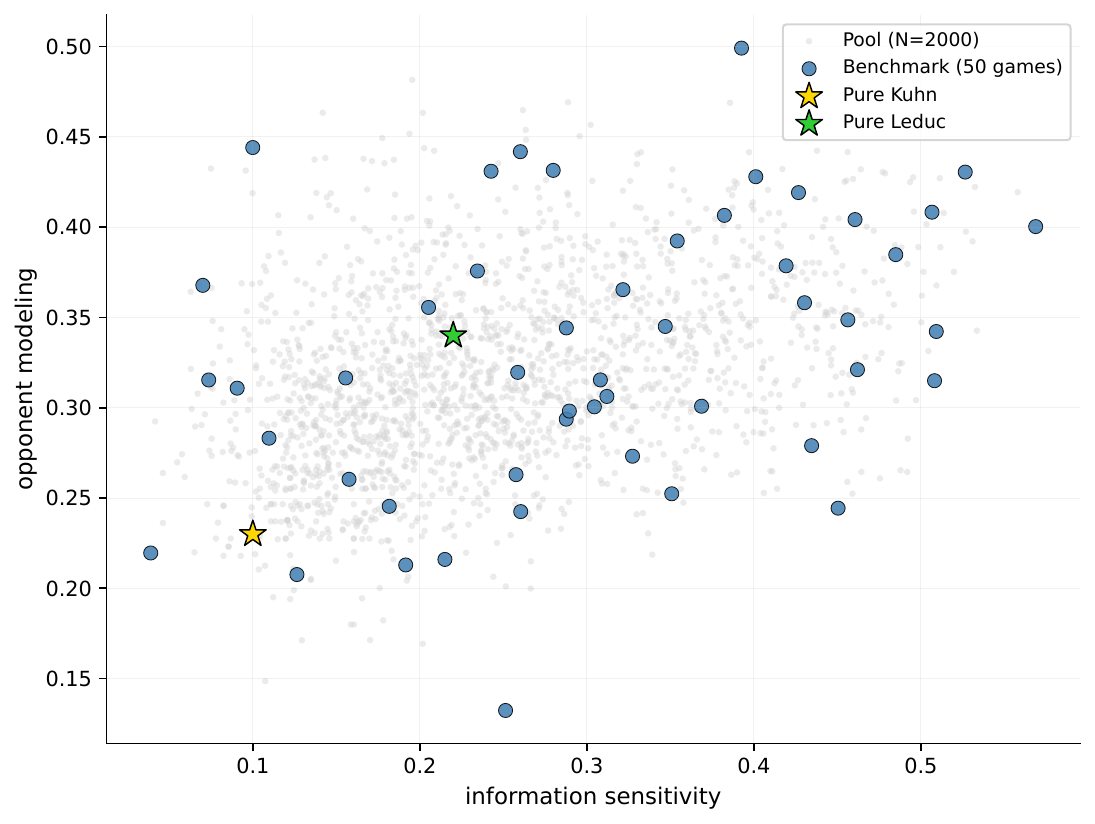}
\caption{The 2{,}000-game accepted pool vs.\ the 50 benchmark games selected by farthest-point
sampling (FPS), in two axes of the selection basis (information
sensitivity $\times$ opponent modeling, screening-run values on the
pool). Gold and green stars mark Kuhn and Leduc poker.}
\label{fig:fps_scatter}
\end{figure}

\begin{figure}[H]\centering\includegraphics[width=0.95\textwidth]{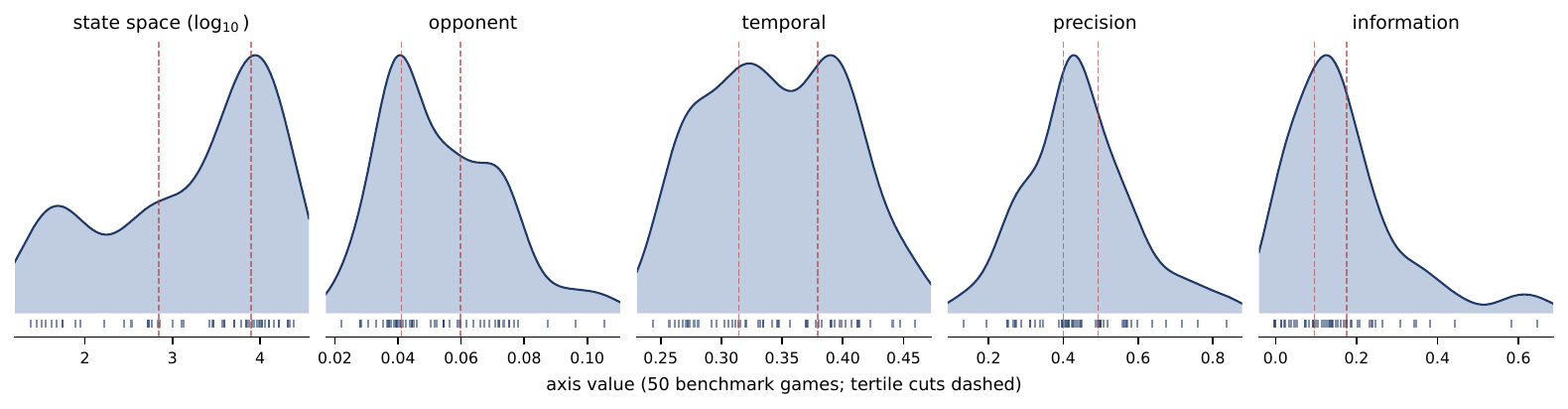}
\caption{Per-axis distributions of the 50 benchmark games: kernel-density estimates with tick-marked
observations and tertile cuts (dashed).}
\label{fig:axis_distributions}\end{figure}

\begin{figure}[H]
\centering
\includegraphics[width=0.55\linewidth]{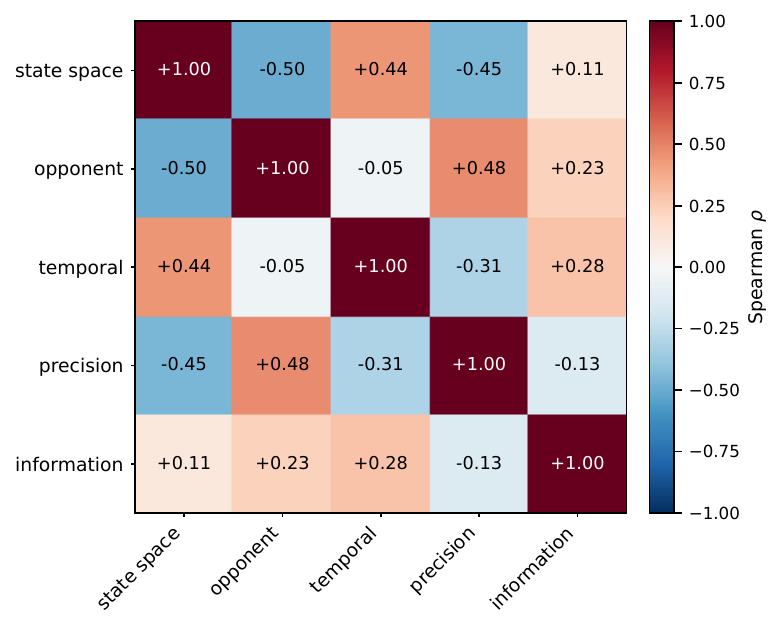}
\caption{\textbf{Pairwise Spearman correlation of the five complexity
axes across the 50 benchmark games.} The largest $|r|$ is $0.50$
(state space $\times$ opponent dependence), under the conventional
$|r| \approx 0.7$ threshold above which joint-regression standard
errors inflate substantially. The largest per-axis VIF, listed earlier in this appendix,
is $2.0$.}
\label{fig:axis_corr}
\end{figure}

Figure~\ref{fig:fps_scatter} shows the 2{,}000-game pool and the 50 selected games in two axes of the
selection basis. Figure~\ref{fig:axis_distributions} shows the distribution of the 50 games on each of
the five axes. The axes are moderately correlated (Figure~\ref{fig:axis_corr}): the largest pairwise
Spearman correlation in absolute value is state space $\times$ opponent dependence ($-0.50$), and state space also
correlates with temporal ($+0.44$) and precision ($-0.45$). Opponent dependence $\times$ precision is $0.48$. The two chip-valued per-decision
axes are smaller on long games, where a single decision carries a smaller share of the stakes. The
largest variance inflation factor, $1/(1-R^2)$ from regressing an axis on the other four, is $2.0$ (opponent dependence):

\begin{center}\small
\begin{tabular}{lc}
\toprule
\textbf{Axis} & \textbf{VIF (50-game benchmark)} \\
\midrule
State space ($\log_{10}$)   & $1.70$ \\
Opponent dependence         & $2.00$ \\
Temporal                    & $1.48$ \\
Precision                   & $1.66$ \\
Information                 & $1.30$ \\
\bottomrule
\end{tabular}
\end{center}

\section{Leaderboards}
\label{app:leaderboards}\label{app:tertile_leaderboards}\label{app:loo}\label{app:alpha_coverage}\label{app:bt}\label{app:no_llama}\label{app:h2h_table}

All leaderboards in the paper result from fitting a linear model on signed chip margins, $y_s = \alpha_i - \alpha_j + \varepsilon_s$ with $\sum_m\alpha_m=0$. We are thus able to estimate a strength parameter for each model, $\hat\alpha_m$, which we use to rank models. Confidence intervals are estimated using a cluster bootstrap in which the two seats of a run form a cluster.

\paragraph{Composite complexity.} We construct a composite complexity score for each game, which is a
linear combination of its five axis scores. Weights are given by the first principal component of the
matrix where element $i,j$ is the slope of model $i$ on axis $j$. Each slope comes from a regression of
$\hat\alpha_{m,g}$, the strength of model $m$ estimated from the deals of game $g$ alone, on the $z$-scored
axis. This principal component is the linear combination of axis scores along which the sensitivities of the
models differ the most, and it explains 93\% of the variance of the matrix after each column is centered over
models. We rescale it to unit absolute sum, with the sign chosen to make the temporal weight positive. As a
result, the composite score is a learned discrimination score. The weights are $+0.35$ for temporal, $+0.29$
for state space, $+0.09$ for information, $-0.19$ for precision, and $-0.07$ for opponent dependence.
Refitting on the eight games with the highest composite score widens the leaderboard spread from $3.3$
chips/game on all fifty to $5.6$ (Table~\ref{tab:hard8}).

Learning the weights on a random half of the games and scoring the other half gives $5.2 \pm 0.2$
chips/game on the eight hardest games (four from each half) over twenty splits. The temporal axis alone uses
no outcome information. By its tertiles, the leaderboard spans $1.83$, $3.00$, and $4.96$ chips/game from
easiest to hardest. An unweighted mean of the five $z$-scored axes gives $1.91$, $3.69$, and $4.36$, and its
eight hardest games widen the spread only to $3.85$, as precision and opponent dependence are often higher
on smaller games. Using the multivariate coefficients of Table~\ref{tab:per_axis_regression} in place of the
slopes gives a score that correlates with ours at $\rho=0.95$ and widens the spread $1.80$ times on its own
top eight. Table~\ref{tab:tertile_leaderboards} refits $\hat\alpha$ within tertiles of the composite score.

\paragraph{Payoff scale.}\label{app:payoff} Chip amounts are comparable within a game but not necessarily across games,
so the wider spread on the hard games could partly reflect larger stakes. In chips, the spread on the eight hardest
games is 1.69 times the spread on all fifty. Measured in each game's payoff standard deviation under random play, the
normalization used for the axes, it is 1.35 times $[1.23, 1.50]$. Measured in the standard deviation of the tournament's
own margins on each game, the most conservative unit because differences between models inflate it, it is still 1.12
times $[1.01, 1.25]$. Refit on normalized margins, the leaderboard keeps a Spearman correlation of 0.97 with the chip
leaderboard, and the tiers stay intact.

\begin{table}[H]
\centering\small
\caption{\textbf{Leaderboard on the eight hardest games} by composite complexity (seeds 9754, 1406, 11435, 2914, 203, 10760, 10427, 4643), sum-to-zero $\hat\alpha$ on 3{,}411 paired deals, paired-cluster
bootstrap 95\% CIs ($B{=}400$). $^\dagger$ September models.}
\label{tab:hard8}
\begin{tabular}{lrrr}
\toprule
\textbf{Model} & $\hat\alpha$ (hard 8) & \textbf{95\% CI} & \textbf{deals} \\
\midrule
\texttt{gpt-6-astra}$^\dagger$ & $+1.52$ & $[+0.95, +2.09]$ & 270 \\
\texttt{gemini-3.1-pro} & $+1.28$ & $[+0.96, +1.59]$ & 847 \\
\texttt{gpt-5.4} & $+1.22$ & $[+0.88, +1.57]$ & 457 \\
\texttt{claude-fable-5.1}$^\dagger$ & $+1.10$ & $[+0.58, +1.64]$ & 260 \\
\texttt{gemini-3.8-flash}$^\dagger$ & $+0.74$ & $[-0.23, +1.55]$ & 80 \\
\texttt{claude-sonnet-4.6} & $+0.72$ & $[+0.28, +1.20]$ & 346 \\
\texttt{gemini-2.5-pro} & $+0.11$ & $[-0.23, +0.45]$ & 720 \\
\texttt{deepseek-v3.1} & $-0.10$ & $[-0.41, +0.24]$ & 768 \\
\texttt{gemma-4-31b} & $-0.26$ & $[-0.61, +0.07]$ & 738 \\
\texttt{gemini-3.1-flash-lite} & $-1.07$ & $[-1.41, -0.74]$ & 788 \\
\texttt{qwen-3.5} & $-1.22$ & $[-1.62, -0.83]$ & 690 \\
\texttt{llama-3.3-70b} & $-4.05$ & $[-4.46, -3.61]$ & 858 \\
\bottomrule
\end{tabular}
\end{table}

\begin{table}[H]
\centering\small
\caption{\textbf{Leaderboard on the eight most temporal games}: sum-to-zero $\hat\alpha$ on $3{,}358$
paired deals, paired-cluster bootstrap 95\% CIs ($B{=}400$). $^\dagger$ September models.}
\label{tab:hard8_temporal}
\begin{tabular}{lrrr}
\toprule
\textbf{Model} & $\hat\alpha$ (temporal top 8) & \textbf{95\% CI} & \textbf{deals} \\
\midrule
\texttt{gpt-6-astra}$^\dagger$ & $+1.73$ & $[+1.19, +2.33]$ & 198 \\
\texttt{gpt-5.4} & $+1.39$ & $[+1.08, +1.74]$ & 406 \\
\texttt{claude-fable-5.1}$^\dagger$ & $+1.22$ & $[+0.64, +1.80]$ & 188 \\
\texttt{gemini-3.1-pro} & $+1.00$ & $[+0.73, +1.25]$ & 857 \\
\texttt{gemini-3.8-flash}$^\dagger$ & $+0.80$ & $[-0.19, +1.67]$ & 80 \\
\texttt{claude-sonnet-4.6} & $+0.54$ & $[+0.18, +0.93]$ & 398 \\
\texttt{gemini-2.5-pro} & $+0.14$ & $[-0.16, +0.47]$ & 720 \\
\texttt{gemma-4-31b} & $-0.26$ & $[-0.58, +0.06]$ & 750 \\
\texttt{deepseek-v3.1} & $-0.46$ & $[-0.76, -0.14]$ & 774 \\
\texttt{gemini-3.1-flash-lite} & $-1.15$ & $[-1.45, -0.87]$ & 799 \\
\texttt{qwen-3.5} & $-1.43$ & $[-1.74, -1.10]$ & 718 \\
\texttt{llama-3.3-70b} & $-3.52$ & $[-3.94, -3.11]$ & 828 \\
\bottomrule
\end{tabular}
\end{table}

\begin{table}[H]
\centering
\caption{\textbf{Per-tertile additive paired-comparison leaderboards.}}
\label{tab:tertile_leaderboards}
\begin{tcolorbox}[leaderboardstyle,
    title={Per-tertile leaderboards \,---\, $\hat\alpha$ by composite-complexity tertile}]
\setlength{\tabcolsep}{10pt}
\renewcommand{\arraystretch}{1.08}
\centering
\begin{tabular}{lrrr}
\toprule
\textbf{Model} & $\hat\alpha_{T1}$ & $\hat\alpha_{T2}$ & $\hat\alpha_{T3}$ \\
\midrule
\texttt{gpt-6-astra}$^\dagger$ & $-0.03$ & $+0.77$ & $+1.29$ \\
\texttt{gpt-5.4}               & $+0.21$ & $+0.63$ & $+1.05$ \\
\texttt{gemini-3.8-flash}$^\dagger$ & $+0.25$ & $+0.52$ & $+1.04$ \\
\texttt{gemini-3.1-pro}        & $+0.29$ & $+0.59$ & $+0.96$ \\
\texttt{claude-sonnet-4.6}     & $+0.23$ & $+0.51$ & $+0.60$ \\
\texttt{claude-fable-5.1}$^\dagger$ & $+0.28$ & $+0.28$ & $+0.86$ \\
\texttt{gemini-2.5-pro}        & $+0.21$ & $+0.14$ & $+0.13$ \\
\texttt{gemma-4-31b}           & $+0.16$ & $+0.09$ & $-0.12$ \\
\texttt{deepseek-v3.1}         & $-0.06$ & $-0.05$ & $-0.22$ \\
\texttt{gemini-3.1-flash-lite} & $-0.08$ & $-0.43$ & $-0.90$ \\
\texttt{qwen-3.5}              & $-0.15$ & $-0.51$ & $-1.01$ \\
\texttt{llama-3.3-70b}         & $-1.32$ & $-2.53$ & $-3.68$ \\
\bottomrule
\end{tabular}\\[6pt]
{\footnotesize
T1 = easiest 16 games, T2 = middle 17, T3 = hardest 17, by the
five-axis composite complexity. Paired-cluster bootstrap 95\% CI widths
($B{=}400$) reach at most $1.04$ chips/game across all 36 cells (widest on
\texttt{gemini-3.8-flash}, T3), with a median width of $0.29$.
Top-3 by tertile: T1 \texttt{gemini-3.1-pro}, \texttt{claude-fable-5.1}, \texttt{gemini-3.8-flash}; T2 \texttt{gpt-6-astra}, \texttt{gpt-5.4}, \texttt{gemini-3.1-pro};
T3 \texttt{gpt-6-astra}, \texttt{gpt-5.4}, \texttt{gemini-3.8-flash}. Bottom-3 by tertile: T1 \texttt{llama-3.3-70b}, \texttt{qwen-3.5}, \texttt{gemini-3.1-flash-lite};
T2 \texttt{llama-3.3-70b}, \texttt{qwen-3.5}, \texttt{gemini-3.1-flash-lite}; T3 \texttt{llama-3.3-70b}, \texttt{qwen-3.5}, \texttt{gemini-3.1-flash-lite}.}
\end{tcolorbox}
\end{table}

\subsection{Robustness}

Dropping each game in turn and refitting, we get a mean Kendall $\tau$ of $0.956$ with the full ranking
(minimum $0.879$). None of these refits breaks the three tiers, and where the order does change, models move only within their
tier, by at most three places (\texttt{gemini-3.1-pro} when seed 1406 is dropped). We also refit the leaderboard on random subsets of
games, drawing $1{,}000$ subsets at each of four sizes. The median Spearman correlation with the full ranking
is $0.89$ for 5 games, $0.92$ for 8, $0.94$ for 12 and $0.97$ for 25, with 5th percentiles of $0.68$, $0.76$, $0.78$
and $0.87$. Refitting the April round-robin alone gives the nine April models in exactly their
twelve-model order (Table~\ref{tab:april9}, Kendall $\tau=1.00$). With the bottom outlier
\texttt{llama-3.3-70b} ($-2.56$ in the full fit) dropped, the refit on the other eleven ($15{,}228$ paired
deals) keeps their order apart from \texttt{claude-sonnet-4.6} and \texttt{claude-fable-5.1}, tied at
$+0.45$ in the full fit, which swap (Table~\ref{tab:no_llama_leaderboard}). The leaderboard is fit on margins since the models were told to maximize expected
chips. A Bradley--Terry fit on the win indicator (ties dropped) orders the models at Spearman $0.77$ with the
margin fit, favoring frequent small-margin winners. Of the 47 model pairs that met, 41 have a mean
head-to-head margin in the direction of the leaderboard, and none of the six exceptions is significant
(Tables~\ref{tab:h2h_full} and~\ref{tab:h2h_n}).

\begin{table}[H]
\centering\small
\caption{\textbf{Leaderboard refit on the April round-robin alone} (nine models, paired-cluster
bootstrap 95\% CIs, $B{=}2{,}000$).}
\label{tab:april9}
\begin{tabular}{lrr}
\toprule
\textbf{Model} & $\hat\alpha$ (April nine) & \textbf{95\% CI} \\
\midrule
\texttt{gpt-5.4} & $+0.86$ & $[+0.76, +0.97]$ \\
\texttt{gemini-3.1-pro} & $+0.83$ & $[+0.75, +0.90]$ \\
\texttt{claude-sonnet-4.6} & $+0.64$ & $[+0.53, +0.76]$ \\
\texttt{gemini-2.5-pro} & $+0.36$ & $[+0.29, +0.43]$ \\
\texttt{gemma-4-31b} & $+0.25$ & $[+0.17, +0.32]$ \\
\texttt{deepseek-v3.1} & $+0.06$ & $[-0.01, +0.13]$ \\
\texttt{gemini-3.1-flash-lite} & $-0.27$ & $[-0.35, -0.20]$ \\
\texttt{qwen-3.5} & $-0.35$ & $[-0.44, -0.27]$ \\
\texttt{llama-3.3-70b} & $-2.37$ & $[-2.47, -2.27]$ \\
\bottomrule
\end{tabular}
\end{table}

\begin{table}[H]
\centering
\caption{\textbf{Llama-excluded leaderboard.}}
\label{tab:no_llama_leaderboard}
\begin{tcolorbox}[leaderboardstyle,
    title={Llama-excluded refit \,---\, $\hat\alpha$ (chips/game)}]
\setlength{\tabcolsep}{10pt}
\renewcommand{\arraystretch}{1.08}
\centering
\begin{tabular}{lrr}
\toprule
\textbf{Model} & $\hat\alpha$ (no llama) & \textbf{95\% CI} \\
\midrule
\texttt{gpt-6-astra}$^\dagger$ & $+0.50$ & $[+0.30, +0.70]$ \\
\texttt{gpt-5.4} & $+0.41$ & $[+0.32, +0.50]$ \\
\texttt{gemini-3.8-flash}$^\dagger$ & $+0.40$ & $[+0.20, +0.61]$ \\
\texttt{gemini-3.1-pro} & $+0.35$ & $[+0.28, +0.43]$ \\
\texttt{claude-fable-5.1}$^\dagger$ & $+0.20$ & $[+0.00, +0.39]$ \\
\texttt{claude-sonnet-4.6} & $+0.18$ & $[+0.07, +0.29]$ \\
\texttt{gemini-2.5-pro} & $-0.06$ & $[-0.14, +0.01]$ \\
\texttt{gemma-4-31b} & $-0.17$ & $[-0.25, -0.09]$ \\
\texttt{deepseek-v3.1} & $-0.35$ & $[-0.44, -0.27]$ \\
\texttt{gemini-3.1-flash-lite} & $-0.66$ & $[-0.74, -0.58]$ \\
\texttt{qwen-3.5} & $-0.79$ & $[-0.88, -0.70]$ \\
\bottomrule
\end{tabular}\\[6pt]
{\footnotesize
Paired-cluster bootstrap 95\% CI ($B{=}1{,}000$), sum-to-zero contrast.
Refit on the 11 non-llama models ($15{,}228$ paired deals).}
\end{tcolorbox}
\end{table}

\begin{table}[H]\centering\scriptsize
\caption{Pairwise mean chip margin (row $-$ column) over the benchmark games on which the pair played,
both seats balanced. Bold = 95\% paired-cluster bootstrap CI excludes 0 ($B{=}500$). Rows and
columns sorted by leaderboard rank. Each of the three September models played a subset of the
field, and ``--'' marks pairs that never met.}
\label{tab:h2h_full}
\setlength{\tabcolsep}{2pt}
\resizebox{\textwidth}{!}{\begin{tabular}{lrrrrrrrrrrrr}
\toprule
 & {\tiny gpt-6-astra$^\dagger$} & {\tiny gpt-5.4} & {\tiny gemini-3.8-flash$^\dagger$} & {\tiny gemini-3.1-pro} & {\tiny claude-sonnet-4.6} & {\tiny claude-fable-5.1$^\dagger$} & {\tiny gemini-2.5-pro} & {\tiny gemma-4-31b} & {\tiny deepseek-v3.1} & {\tiny gemini-3.1-flash-lite} & {\tiny qwen-3.5} & {\tiny llama-3.3-70b} \\
\midrule
\texttt{gpt-6-astra}$^\dagger$ & -- & $+0.02$ & -- & $-0.05$ & -- & $\mathbf{+0.84}$ & -- & -- & -- & -- & -- & $\mathbf{+3.20}$ \\
\texttt{gpt-5.4} & $-0.02$ & -- & $-0.04$ & $+0.20$ & $\mathbf{+0.45}$ & $-0.44$ & $\mathbf{+0.34}$ & $\mathbf{+0.60}$ & $\mathbf{+1.02}$ & $\mathbf{+0.91}$ & $\mathbf{+0.95}$ & $\mathbf{+3.22}$ \\
\texttt{gemini-3.8-flash}$^\dagger$ & -- & $+0.04$ & -- & $-0.04$ & -- & -- & -- & -- & -- & $\mathbf{+1.11}$ & -- & -- \\
\texttt{gemini-3.1-pro} & $+0.05$ & $-0.20$ & $+0.04$ & -- & $+0.17$ & $-0.06$ & $\mathbf{+0.42}$ & $\mathbf{+0.58}$ & $\mathbf{+0.71}$ & $\mathbf{+0.83}$ & $\mathbf{+1.29}$ & $\mathbf{+3.54}$ \\
\texttt{claude-sonnet-4.6} & -- & $\mathbf{-0.45}$ & -- & $-0.17$ & -- & $+0.08$ & $-0.01$ & $+0.39$ & $\mathbf{+0.65}$ & $\mathbf{+1.25}$ & $\mathbf{+0.81}$ & $\mathbf{+3.26}$ \\
\texttt{claude-fable-5.1}$^\dagger$ & $\mathbf{-0.84}$ & $+0.44$ & -- & $+0.06$ & $-0.08$ & -- & -- & $\mathbf{+0.25}$ & -- & -- & -- & -- \\
\texttt{gemini-2.5-pro} & -- & $\mathbf{-0.34}$ & -- & $\mathbf{-0.42}$ & $+0.01$ & -- & -- & $+0.07$ & $+0.17$ & $\mathbf{+0.76}$ & $\mathbf{+0.60}$ & $\mathbf{+2.71}$ \\
\texttt{gemma-4-31b} & -- & $\mathbf{-0.60}$ & -- & $\mathbf{-0.58}$ & $-0.39$ & $\mathbf{-0.25}$ & $-0.07$ & -- & $+0.13$ & $\mathbf{+0.45}$ & $\mathbf{+0.74}$ & $\mathbf{+2.54}$ \\
\texttt{deepseek-v3.1} & -- & $\mathbf{-1.02}$ & -- & $\mathbf{-0.71}$ & $\mathbf{-0.65}$ & -- & $-0.17$ & $-0.13$ & -- & $\mathbf{+0.48}$ & $\mathbf{+0.22}$ & $\mathbf{+2.31}$ \\
\texttt{gemini-3.1-flash-lite} & -- & $\mathbf{-0.91}$ & $\mathbf{-1.11}$ & $\mathbf{-0.83}$ & $\mathbf{-1.25}$ & -- & $\mathbf{-0.76}$ & $\mathbf{-0.45}$ & $\mathbf{-0.48}$ & -- & $\mathbf{+0.33}$ & $\mathbf{+1.79}$ \\
\texttt{qwen-3.5} & -- & $\mathbf{-0.95}$ & -- & $\mathbf{-1.29}$ & $\mathbf{-0.81}$ & -- & $\mathbf{-0.60}$ & $\mathbf{-0.74}$ & $\mathbf{-0.22}$ & $\mathbf{-0.33}$ & -- & $\mathbf{+2.08}$ \\
\texttt{llama-3.3-70b} & $\mathbf{-3.20}$ & $\mathbf{-3.22}$ & -- & $\mathbf{-3.54}$ & $\mathbf{-3.26}$ & -- & $\mathbf{-2.71}$ & $\mathbf{-2.54}$ & $\mathbf{-2.31}$ & $\mathbf{-1.79}$ & $\mathbf{-2.08}$ & -- \\
\bottomrule\end{tabular}}
\end{table}

\begin{table}[H]\centering\scriptsize
\caption{Paired deals per model pair in the unified corpus ($20{,}206$ in all). Rows and columns in leaderboard order,
and ``--'' marks pairs that never met.}
\label{tab:h2h_n}
\setlength{\tabcolsep}{2pt}
\resizebox{\textwidth}{!}{\begin{tabular}{lrrrrrrrrrrrr}
\toprule
 & {\tiny gpt-6-astra$^\dagger$} & {\tiny gpt-5.4} & {\tiny gemini-3.8-flash$^\dagger$} & {\tiny gemini-3.1-pro} & {\tiny claude-sonnet-4.6} & {\tiny claude-fable-5.1$^\dagger$} & {\tiny gemini-2.5-pro} & {\tiny gemma-4-31b} & {\tiny deepseek-v3.1} & {\tiny gemini-3.1-flash-lite} & {\tiny qwen-3.5} & {\tiny llama-3.3-70b} \\
\midrule
\texttt{gpt-6-astra}$^\dagger$ & -- & 260 & -- & 170 & -- & 100 & -- & -- & -- & -- & -- & 160 \\
\texttt{gpt-5.4} & 260 & -- & 160 & 220 & 240 & 80 & 200 & 260 & 216 & 220 & 210 & 260 \\
\texttt{gemini-3.8-flash}$^\dagger$ & -- & 160 & -- & 170 & -- & -- & -- & -- & -- & 170 & -- & -- \\
\texttt{gemini-3.1-pro} & 170 & 220 & 170 & -- & 240 & 170 & 700 & 720 & 716 & 720 & 699 & 720 \\
\texttt{claude-sonnet-4.6} & -- & 240 & -- & 240 & -- & 170 & 220 & 199 & 217 & 239 & 234 & 280 \\
\texttt{claude-fable-5.1}$^\dagger$ & 100 & 80 & -- & 170 & 170 & -- & -- & 160 & -- & -- & -- & -- \\
\texttt{gemini-2.5-pro} & -- & 200 & -- & 700 & 220 & -- & -- & 720 & 700 & 720 & 720 & 720 \\
\texttt{gemma-4-31b} & -- & 260 & -- & 720 & 199 & 160 & 720 & -- & 719 & 720 & 719 & 700 \\
\texttt{deepseek-v3.1} & -- & 216 & -- & 716 & 217 & -- & 700 & 719 & -- & 698 & 714 & 719 \\
\texttt{gemini-3.1-flash-lite} & -- & 220 & 170 & 720 & 239 & -- & 720 & 720 & 698 & -- & 718 & 720 \\
\texttt{qwen-3.5} & -- & 210 & -- & 699 & 234 & -- & 720 & 719 & 714 & 718 & -- & 699 \\
\texttt{llama-3.3-70b} & 160 & 260 & -- & 720 & 280 & -- & 720 & 700 & 719 & 720 & 699 & -- \\
\bottomrule\end{tabular}}
\end{table}

\section{Per-game results}
\label{app:per_game}\label{app:per_cell_alpha}\label{app:variance}

Table~\ref{tab:per_game} gives each model's raw mean margin on each game and Table~\ref{tab:per_cell_alpha}
the per-game strengths $\hat\alpha_{m,g}$, the leaderboard fit restricted to one game's deals, which
corrects for the opponents a model happened to face there. Rows of both are sorted by the mean of the five
$z$-scored axes.

\paragraph{Variance decomposition.} The noise-corrected model$\times$game interaction $\sigma^2_{MG}$ of the
$12\times50$ grid of $\hat\alpha_{m,g}$ is about a third of the model main effect $\sigma^2_M$ on the full data
(ratio $0.34$, 95\% CI $[0.20, 0.54]$) and comparable to it without \texttt{llama-3.3-70b} ($0.65$,
$[0.14, 1.31]$). Here the main effect is the variance of the per-model means and the interaction is the variance of
the residuals after removing model and game means. Each cell is estimated from a finite number of deals, and
estimation noise accounts for roughly half of the raw interaction. We subtract from each component the noise
variance estimated for it by the paired-cluster bootstrap.

\begin{center}\small
\begin{tabular}{lrrrr}
\toprule
\textbf{Quantity} & \textbf{observed} & \textbf{noise} & \textbf{corrected} & \textbf{95\% CI} \\
\midrule
\multicolumn{5}{l}{\emph{Full 12 models}} \\
$\sigma^2_M$ & $0.77$ & $0.00$ & $0.76$ & $[0.70, 0.82]$ \\
$\sigma^2_{MG}$ & $0.48$ & $0.22$ & $0.26$ & $[0.15, 0.41]$ \\
$\sigma^2_{MG}/\sigma^2_M$ & $0.62$ & --- & $0.34$ & $[0.20, 0.54]$ \\
\midrule
\multicolumn{5}{l}{\emph{Llama-excluded}} \\
$\sigma^2_M$ & $0.20$ & $0.00$ & $0.19$ & $[0.16, 0.24]$ \\
$\sigma^2_{MG}$ & $0.31$ & $0.19$ & $0.13$ & $[0.03, 0.27]$ \\
$\sigma^2_{MG}/\sigma^2_M$ & $1.58$ & --- & $0.65$ & $[0.14, 1.31]$ \\
\bottomrule
\end{tabular}
\end{center}

\begin{table}[H]\centering\tiny
\caption{Per-(model, game) pair-mean chip margin on the 50-game benchmark, computed
as an exact mean over the rows played and sorted ascending by \texttt{cmplx}, the mean of the five $z$-scored axes. No cell is missing (a
missing cell would print as ``--''). Entries of $+0.00$ are mean margins of
exactly zero. In 59 of these 68 cells the two deals of every run cancelled, and in the other 9 the runs
net to zero.}
\label{tab:per_game}
\setlength{\tabcolsep}{2pt}
\resizebox{\textwidth}{!}{\begin{tabular}{lrrrrrrrrrrrrr}
\toprule
seed & cmplx & {\tiny gpt-6-astra$^\dagger$} & {\tiny gpt-5.4} & {\tiny gemini-3.8-flash$^\dagger$} & {\tiny gemini-3.1-pro} & {\tiny claude-sonnet-4.6} & {\tiny claude-fable-5.1$^\dagger$} & {\tiny gemini-2.5-pro} & {\tiny gemma-4-31b} & {\tiny deepseek-v3.1} & {\tiny gemini-3.1-flash-lite} & {\tiny qwen-3.5} & {\tiny llama-3.3-70b} \\\midrule
2385 & -0.82 & +0.30 & +0.16 & +0.30 & +0.33 & +0.24 & +0.00 & +0.25 & -0.13 & +0.04 & +0.23 & -0.11 & -0.81 \\
6903 & -0.80 & +0.10 & +0.17 & -0.20 & +0.85 & -0.35 & -0.50 & +0.30 & +0.36 & -0.07 & -0.14 & -0.37 & -1.20 \\
10173 & -0.80 & +1.50 & +0.18 & +0.00 & +0.73 & -0.30 & +0.30 & +0.47 & -0.03 & +0.07 & -0.12 & +0.56 & -1.94 \\
7052 & -0.70 & +1.00 & -0.25 & +0.20 & +1.09 & +0.70 & -0.40 & +0.61 & +0.89 & -0.19 & -0.75 & -0.17 & -2.11 \\
2495 & -0.60 & -0.40 & +0.28 & +0.10 & +0.08 & -0.07 & +0.20 & +0.38 & -0.08 & +0.06 & -0.17 & -0.42 & +0.06 \\
11784 & -0.55 & +0.00 & +0.55 & +0.20 & +0.72 & +0.68 & +0.00 & +0.30 & +0.29 & +0.40 & -0.20 & -0.19 & -1.93 \\
5149 & -0.54 & +0.10 & +0.00 & +0.00 & +0.12 & +0.25 & -0.10 & +0.11 & +0.46 & +0.31 & +0.01 & -0.29 & -0.95 \\
9548 & -0.50 & +0.00 & +0.46 & +0.80 & +1.24 & +0.40 & +0.00 & +0.27 & +0.25 & +0.74 & -0.33 & -0.63 & -2.17 \\
9398 & -0.44 & -0.20 & +1.01 & -0.20 & +1.59 & +2.86 & -0.90 & +0.81 & +0.09 & +0.55 & -0.66 & -1.23 & -2.81 \\
963 & -0.38 & +0.00 & +0.00 & +0.30 & +0.45 & +0.00 & +0.60 & +0.49 & +0.01 & +0.21 & +0.05 & +0.03 & -1.64 \\
8489 & -0.37 & +0.50 & +0.96 & +0.10 & +0.76 & -0.15 & -0.50 & +0.78 & +0.38 & -0.58 & +0.57 & +0.34 & -3.00 \\
4968 & -0.36 & +0.90 & +2.30 & +1.30 & +0.94 & +0.40 & +0.10 & +0.88 & +0.66 & -0.42 & -0.16 & -0.31 & -1.86 \\
2914 & -0.34 & +0.50 & +1.67 & +1.00 & +2.31 & -0.37 & +0.70 & -0.34 & -0.67 & +0.44 & -1.41 & -1.25 & +0.19 \\
2345 & -0.29 & +0.00 & +0.53 & +0.00 & +0.48 & +0.45 & +0.00 & +0.21 & +0.35 & -0.14 & +0.12 & -0.30 & -1.19 \\
4426 & -0.27 & +0.20 & +0.15 & +0.00 & +0.11 & +0.04 & +0.00 & +0.14 & +0.04 & +0.10 & +0.00 & -0.05 & -0.32 \\
5470 & -0.24 & +0.50 & +0.28 & -0.20 & +0.22 & +0.10 & -0.20 & +0.23 & -0.06 & +0.03 & -0.11 & -0.13 & -0.50 \\
10738 & -0.21 & +1.00 & -0.18 & -0.10 & +0.83 & -0.10 & -0.20 & +0.52 & +0.16 & -0.35 & +0.20 & -0.51 & -1.04 \\
862 & -0.20 & +0.00 & +0.00 & +0.00 & +0.54 & +0.00 & +0.00 & +0.47 & +0.00 & +0.28 & +0.43 & -0.01 & -2.05 \\
2601 & -0.19 & +2.70 & +0.08 & +0.30 & +0.69 & -0.15 & -0.30 & +0.62 & +0.28 & -0.19 & +0.59 & +0.39 & -2.39 \\
10137 & -0.15 & -0.20 & +0.65 & +0.60 & +1.61 & +0.53 & +0.00 & +0.22 & +1.19 & +0.66 & +0.09 & -0.50 & -3.83 \\
6586 & -0.14 & +2.50 & +0.30 & +0.00 & +0.21 & +0.80 & +0.40 & +0.57 & -0.07 & +0.18 & -0.06 & +0.20 & -1.50 \\
1247 & -0.13 & +4.80 & +0.10 & +0.00 & +1.38 & +0.01 & +0.00 & +0.35 & +0.65 & +0.03 & -0.54 & +0.44 & -2.70 \\
10786 & -0.08 & -0.10 & +0.62 & +0.00 & +0.88 & +0.20 & -0.50 & +0.75 & +0.10 & +0.33 & +0.12 & -0.23 & -2.71 \\
11517 & -0.06 & +0.00 & +0.00 & +0.00 & +0.50 & +0.88 & +0.00 & +0.62 & +0.00 & +0.51 & +0.00 & +0.67 & -2.66 \\
10790 & -0.04 & -1.00 & +0.91 & +0.00 & +0.48 & +0.00 & +0.30 & +0.84 & +0.55 & -0.31 & -0.11 & -0.01 & -2.83 \\
3483 & +0.05 & +0.00 & +0.65 & -0.30 & +1.16 & +1.00 & +0.10 & +0.55 & +1.36 & +0.34 & -0.31 & -1.10 & -2.61 \\
7142 & +0.06 & +0.00 & +0.50 & +0.40 & +0.37 & -0.05 & +0.00 & +0.15 & +0.54 & -0.24 & -0.14 & +0.05 & -1.00 \\
8297 & +0.06 & +0.00 & +0.00 & +1.50 & +0.64 & +0.62 & +0.30 & -0.19 & +1.33 & +0.02 & -0.35 & -0.06 & -2.31 \\
10427 & +0.08 & +0.74 & +0.29 & +0.50 & +1.24 & +3.79 & -0.22 & +0.27 & +0.95 & -0.32 & -0.16 & +0.56 & -4.04 \\
9754 & +0.12 & -0.43 & -0.27 & -0.50 & +2.52 & +2.15 & +0.63 & +0.21 & +1.27 & -0.56 & -1.04 & -0.12 & -3.95 \\
11435 & +0.16 & +1.54 & -0.01 & +1.10 & +2.06 & +6.05 & -0.09 & +1.05 & +0.31 & +1.90 & -0.52 & +0.60 & -6.14 \\
10760 & +0.16 & +2.15 & +0.97 & -1.90 & +1.78 & +0.25 & -0.52 & +2.47 & -0.33 & +0.66 & -1.58 & +0.25 & -5.13 \\
6520 & +0.16 & +0.30 & +0.24 & -0.60 & +0.73 & +1.36 & +0.70 & +0.30 & +0.87 & +0.02 & -0.53 & -0.03 & -2.50 \\
6635 & +0.16 & +0.40 & +1.89 & -1.00 & +1.23 & +1.77 & -0.40 & +0.96 & +0.15 & +0.77 & +0.01 & -0.78 & -3.06 \\
7830 & +0.18 & +4.00 & +0.30 & +0.00 & +0.90 & -0.50 & -0.90 & +0.83 & +0.79 & +0.05 & -0.11 & -0.89 & -2.54 \\
7379 & +0.21 & -0.20 & +1.23 & +0.50 & +1.42 & +0.05 & +1.00 & +0.96 & +0.44 & +0.87 & -0.64 & -0.12 & -3.48 \\
5368 & +0.26 & +0.50 & +0.40 & +0.50 & +0.45 & +0.10 & +0.00 & +0.38 & +0.45 & +1.08 & +0.00 & +0.04 & -3.15 \\
4643 & +0.27 & +0.30 & +2.15 & +0.30 & +2.14 & +1.43 & -1.40 & +1.20 & -0.10 & -1.07 & +1.38 & -1.29 & -3.22 \\
6644 & +0.30 & +0.00 & +1.50 & +4.30 & +2.02 & +1.20 & +1.20 & +1.64 & +0.20 & +0.33 & -0.98 & -0.97 & -3.58 \\
933 & +0.32 & +0.00 & +2.37 & +2.20 & +1.11 & +0.70 & -0.70 & +0.74 & -0.03 & +0.55 & -0.41 & -0.61 & -2.48 \\
1406 & +0.39 & +2.35 & +1.07 & +3.00 & +1.66 & +0.64 & -0.04 & +1.83 & +0.82 & +0.52 & -0.68 & -0.47 & -5.31 \\
3651 & +0.42 & +3.30 & +0.65 & +1.00 & +1.71 & +0.70 & -0.20 & +0.89 & -0.06 & +0.60 & +0.66 & -0.66 & -3.79 \\
11520 & +0.46 & +0.00 & +0.04 & +0.00 & +0.14 & +0.00 & +0.00 & +0.00 & +0.15 & +0.20 & +0.20 & -0.02 & -0.78 \\
12140 & +0.52 & +0.00 & +2.74 & +0.00 & +2.02 & +2.62 & +0.60 & +0.99 & -0.27 & +0.55 & -0.38 & -0.61 & -4.28 \\
4059 & +0.56 & +4.50 & +0.25 & -0.70 & +0.68 & +1.10 & -0.20 & +0.69 & +0.26 & -0.01 & +0.25 & -0.17 & -2.06 \\
203 & +0.58 & +0.00 & +2.60 & -0.70 & +2.13 & +0.45 & -0.10 & +1.09 & +1.93 & +1.24 & -0.80 & -1.38 & -4.66 \\
1608 & +0.65 & +0.00 & +1.00 & +0.00 & +0.14 & -0.15 & +0.00 & +0.48 & +0.48 & +0.12 & +0.17 & +0.24 & -1.99 \\
1949 & +0.92 & +0.00 & +1.28 & +0.00 & +1.01 & +0.00 & +0.00 & +0.00 & +0.56 & +0.47 & +0.82 & -0.02 & -3.39 \\
2526 & +1.01 & +0.00 & +2.98 & +4.60 & +2.84 & +1.36 & +1.00 & +1.59 & +2.79 & +0.06 & -1.39 & -2.16 & -4.76 \\
2522 & +1.13 & +1.90 & +0.00 & +0.00 & +0.00 & +3.00 & +0.00 & +0.00 & +0.59 & +0.53 & +0.61 & +0.56 & -2.54 \\
\bottomrule\end{tabular}}
\end{table}

\begin{table}[H]\centering\tiny
\caption{Per-game strength estimates $\hat\alpha_{m,g}$
(chips/game), per model. Rows are the 50 benchmark games sorted
ascending by the mean of the five $z$-scored axes, and columns are the models in
overall leaderboard order. In game 12140, \texttt{gpt-6-astra} and \texttt{gemini-3.8-flash} each played only
\texttt{gpt-5.4} and broke exactly even with it, so all three receive the same value. September models met one to
three opponents per game, so their cells rest on few matchups.}
\label{tab:per_cell_alpha}
\setlength{\tabcolsep}{2pt}
\resizebox{\textwidth}{!}{\begin{tabular}{rrrrrrrrrrrrrr}
\toprule
seed & cmplx & gpt-6-astra$^\dagger$ & gpt-5.4 & gemini-3.8-flash$^\dagger$ & gemini-3.1-pro & claude-sonnet-4.6 & claude-fable-5.1$^\dagger$ & gemini-2.5-pro & gemma-4-31b & deepseek-v3.1 & gemini-3.1-flash-lite & qwen-3.5 & llama-3.3-70b \\
\midrule
2385 & $-0.82$ & $+0.50$ & $-0.01$ & $+0.29$ & $+0.20$ & $+0.01$ & $+0.01$ & $+0.06$ & $-0.11$ & $-0.12$ & $+0.08$ & $-0.14$ & $-0.77$ \\
6903 & $-0.80$ & $+0.25$ & $+0.15$ & $+0.43$ & $+0.63$ & $-0.20$ & $+0.13$ & $+0.23$ & $+0.22$ & $-0.22$ & $-0.26$ & $-0.24$ & $-1.11$ \\
10173 & $-0.80$ & $-0.33$ & $+0.41$ & $+0.41$ & $+0.43$ & $+0.13$ & $+0.38$ & $+0.30$ & $+0.08$ & $-0.16$ & $+0.07$ & $+0.10$ & $-1.83$ \\
7052 & $-0.70$ & $+1.95$ & $+0.34$ & $+1.15$ & $+0.95$ & $+0.19$ & $-0.21$ & $+0.23$ & $+0.11$ & $-0.25$ & $-1.52$ & $-0.66$ & $-2.29$ \\
2495 & $-0.60$ & $-0.50$ & $+0.33$ & $+0.02$ & $+0.12$ & $+0.02$ & $+0.22$ & $+0.32$ & $-0.02$ & $+0.02$ & $-0.08$ & $-0.35$ & $-0.10$ \\
11784 & $-0.55$ & $+0.30$ & $+0.52$ & $-0.15$ & $+0.30$ & $+0.60$ & $+0.20$ & $+0.26$ & $+0.20$ & $+0.04$ & $-0.35$ & $-0.03$ & $-1.88$ \\
5149 & $-0.54$ & $+0.29$ & $+0.11$ & $+0.19$ & $+0.19$ & $+0.31$ & $+0.12$ & $+0.11$ & $+0.22$ & $-0.04$ & $-0.08$ & $-0.40$ & $-1.02$ \\
9548 & $-0.50$ & $+0.36$ & $+0.36$ & $+1.41$ & $+0.61$ & $+0.61$ & $-0.10$ & $+0.21$ & $-0.10$ & $+0.07$ & $-0.65$ & $-0.60$ & $-2.19$ \\
9398 & $-0.44$ & $+0.69$ & $+0.89$ & $+0.69$ & $+1.05$ & $+1.53$ & $+0.63$ & $+0.45$ & $-0.38$ & $-0.24$ & $-0.56$ & $-1.67$ & $-3.07$ \\
963 & $-0.38$ & $+0.25$ & $+0.18$ & $+0.42$ & $+0.25$ & $-0.06$ & $+0.66$ & $+0.18$ & $+0.06$ & $-0.06$ & $+0.12$ & $-0.36$ & $-1.63$ \\
8489 & $-0.37$ & $+0.81$ & $+1.30$ & $+1.40$ & $+0.31$ & $+0.25$ & $-0.93$ & $+0.59$ & $-0.43$ & $-0.28$ & $-0.30$ & $+0.20$ & $-2.93$ \\
4968 & $-0.36$ & $-0.95$ & $+0.45$ & $+1.17$ & $+0.58$ & $-0.21$ & $+0.68$ & $+0.48$ & $+0.64$ & $-0.64$ & $-0.13$ & $-0.22$ & $-1.85$ \\
2914 & $-0.34$ & $+3.13$ & $+2.63$ & $-0.66$ & $+1.30$ & $-0.12$ & $-0.38$ & $-0.77$ & $-1.08$ & $-0.19$ & $-1.66$ & $-1.37$ & $-0.81$ \\
2345 & $-0.29$ & $+0.48$ & $+0.33$ & $+0.33$ & $+0.48$ & $+0.30$ & $+0.48$ & $+0.15$ & $+0.03$ & $-0.47$ & $-0.20$ & $-0.46$ & $-1.46$ \\
4426 & $-0.27$ & $-0.11$ & $-0.16$ & $+0.09$ & $+0.09$ & $+0.06$ & $+0.06$ & $+0.09$ & $+0.09$ & $+0.07$ & $+0.01$ & $+0.01$ & $-0.31$ \\
5470 & $-0.24$ & $-0.04$ & $+0.20$ & $-0.00$ & $+0.24$ & $+0.17$ & $+0.04$ & $+0.15$ & $-0.08$ & $+0.10$ & $-0.13$ & $-0.11$ & $-0.54$ \\
10738 & $-0.21$ & $+0.84$ & $-0.16$ & $-0.26$ & $+0.60$ & $+0.23$ & $+0.40$ & $+0.33$ & $-0.03$ & $-0.17$ & $-0.19$ & $-0.58$ & $-1.01$ \\
862 & $-0.20$ & $+0.22$ & $+0.13$ & $+0.24$ & $+0.22$ & $+0.20$ & $+0.16$ & $+0.24$ & $+0.16$ & $-0.07$ & $+0.24$ & $+0.13$ & $-1.89$ \\
2601 & $-0.19$ & $+0.54$ & $+0.16$ & $+0.51$ & $+0.38$ & $-0.13$ & $+0.08$ & $+0.22$ & $+0.02$ & $+0.02$ & $+0.21$ & $+0.15$ & $-2.16$ \\
10137 & $-0.15$ & $+0.37$ & $+1.31$ & $+0.67$ & $+0.57$ & $+0.55$ & $+0.66$ & $-0.24$ & $+0.66$ & $+0.29$ & $+0.07$ & $-1.12$ & $-3.80$ \\
6586 & $-0.14$ & $+1.06$ & $+0.39$ & $+0.09$ & $+0.09$ & $+0.12$ & $+0.49$ & $+0.14$ & $-0.22$ & $-0.23$ & $-0.17$ & $-0.31$ & $-1.44$ \\
1247 & $-0.13$ & $+2.26$ & $+0.31$ & $+0.31$ & $+0.66$ & $+0.16$ & $+0.16$ & $-0.19$ & $+0.25$ & $-0.48$ & $-0.54$ & $-0.39$ & $-2.54$ \\
10786 & $-0.08$ & $+0.48$ & $+0.66$ & $+0.58$ & $+0.58$ & $+0.63$ & $+0.13$ & $+0.03$ & $+0.21$ & $-0.19$ & $-0.28$ & $-0.14$ & $-2.68$ \\
11517 & $-0.06$ & $+0.29$ & $+0.29$ & $+0.09$ & $+0.09$ & $+0.29$ & $+0.23$ & $+0.15$ & $+0.23$ & $+0.20$ & $+0.21$ & $+0.37$ & $-2.44$ \\
10790 & $-0.04$ & $+0.10$ & $+1.10$ & $-0.36$ & $+0.70$ & $+0.51$ & $+1.00$ & $+0.37$ & $+0.02$ & $-0.15$ & $-0.36$ & $-0.08$ & $-2.84$ \\
3483 & $+0.05$ & $+0.37$ & $+0.74$ & $+0.07$ & $+0.37$ & $+1.28$ & $+1.38$ & $+0.29$ & $+0.19$ & $+0.09$ & $-0.29$ & $-1.72$ & $-2.77$ \\
7142 & $+0.06$ & $+0.26$ & $+0.10$ & $+0.50$ & $+0.26$ & $-0.32$ & $+0.26$ & $+0.10$ & $+0.36$ & $-0.35$ & $-0.27$ & $+0.07$ & $-0.98$ \\
8297 & $+0.06$ & $+0.55$ & $+0.55$ & $+1.01$ & $+0.55$ & $+0.28$ & $+1.04$ & $-0.65$ & $+0.74$ & $-0.20$ & $-0.49$ & $-0.79$ & $-2.61$ \\
10427 & $+0.08$ & $+0.22$ & $+0.54$ & $+1.04$ & $+1.04$ & $+1.95$ & $+0.29$ & $-0.29$ & $-0.06$ & $-0.21$ & $-0.93$ & $+0.26$ & $-3.85$ \\
9754 & $+0.12$ & $+1.39$ & $+1.51$ & $+1.20$ & $+1.70$ & $+1.82$ & $+2.13$ & $-1.02$ & $-0.07$ & $-0.49$ & $-2.02$ & $-1.60$ & $-4.55$ \\
11435 & $+0.16$ & $+0.79$ & $+0.65$ & $+1.12$ & $+1.26$ & $+0.36$ & $+0.59$ & $+0.07$ & $+0.49$ & $+0.98$ & $+0.02$ & $-0.61$ & $-5.69$ \\
10760 & $+0.16$ & $+2.31$ & $+1.75$ & $-0.66$ & $+1.24$ & $+1.94$ & $+1.36$ & $+1.69$ & $-0.89$ & $-0.85$ & $-1.38$ & $-1.73$ & $-4.80$ \\
6520 & $+0.16$ & $+1.35$ & $+1.05$ & $-0.28$ & $+0.32$ & $+1.72$ & $+0.93$ & $-0.04$ & $+0.23$ & $-0.36$ & $-1.30$ & $-0.60$ & $-3.01$ \\
6635 & $+0.16$ & $+1.17$ & $+0.77$ & $-0.15$ & $+0.85$ & $+0.78$ & $+0.45$ & $+0.50$ & $-0.48$ & $+0.34$ & $-0.20$ & $-1.21$ & $-2.84$ \\
7830 & $+0.18$ & $+1.44$ & $+0.58$ & $+0.58$ & $+0.76$ & $+0.08$ & $-0.14$ & $+0.50$ & $+0.35$ & $+0.08$ & $-0.46$ & $-1.19$ & $-2.56$ \\
7379 & $+0.21$ & $+0.73$ & $+0.93$ & $+1.43$ & $+0.83$ & $+0.13$ & $+0.80$ & $+0.20$ & $-0.20$ & $+0.22$ & $-0.77$ & $-0.99$ & $-3.32$ \\
5368 & $+0.26$ & $+1.54$ & $+1.04$ & $-0.17$ & $+0.42$ & $+0.85$ & $+0.85$ & $+0.21$ & $-0.02$ & $-0.05$ & $-0.67$ & $-0.54$ & $-3.47$ \\
4643 & $+0.27$ & $+1.74$ & $+1.44$ & $+1.45$ & $+1.15$ & $+0.96$ & $-0.44$ & $+0.06$ & $-0.49$ & $-0.69$ & $-0.17$ & $-1.82$ & $-3.19$ \\
6644 & $+0.30$ & $+0.73$ & $+2.21$ & $+2.86$ & $+0.73$ & $-0.45$ & $+0.75$ & $+0.71$ & $-0.22$ & $-0.70$ & $-1.44$ & $-1.39$ & $-3.81$ \\
933 & $+0.32$ & $+0.92$ & $+0.92$ & $+2.73$ & $+0.53$ & $+0.71$ & $+0.01$ & $-0.05$ & $-0.51$ & $-0.48$ & $-1.19$ & $-0.95$ & $-2.63$ \\
1406 & $+0.39$ & $+2.37$ & $+1.46$ & $+2.26$ & $-0.07$ & $+0.47$ & $+1.60$ & $+0.54$ & $-0.51$ & $-0.69$ & $-0.74$ & $-1.24$ & $-5.44$ \\
3651 & $+0.42$ & $-0.29$ & $+1.00$ & $+2.29$ & $+1.29$ & $+0.35$ & $+0.15$ & $-0.21$ & $-0.03$ & $-0.18$ & $-0.00$ & $-0.77$ & $-3.59$ \\
11520 & $+0.46$ & $+0.03$ & $+0.12$ & $+0.12$ & $+0.03$ & $+0.05$ & $+0.05$ & $+0.06$ & $+0.09$ & $+0.07$ & $+0.06$ & $+0.05$ & $-0.71$ \\
12140 & $+0.52$ & $+1.75$ & $+1.75$ & $+1.75$ & $+1.05$ & $+1.17$ & $-0.01$ & $-0.59$ & $-0.61$ & $-0.54$ & $-0.92$ & $-0.83$ & $-3.96$ \\
4059 & $+0.56$ & $+2.57$ & $+0.20$ & $-0.45$ & $+0.25$ & $+0.15$ & $+0.05$ & $+0.30$ & $-0.25$ & $+0.03$ & $-0.41$ & $-0.49$ & $-1.93$ \\
203 & $+0.58$ & $+0.47$ & $+0.47$ & $-0.23$ & $+1.38$ & $+1.69$ & $+1.59$ & $+0.31$ & $+0.92$ & $+0.72$ & $-0.84$ & $-2.11$ & $-4.38$ \\
1608 & $+0.65$ & $+0.40$ & $+0.40$ & $-0.11$ & $+0.25$ & $+0.16$ & $+0.25$ & $+0.23$ & $+0.22$ & $+0.20$ & $-0.11$ & $-0.05$ & $-1.85$ \\
1949 & $+0.92$ & $+0.57$ & $+0.20$ & $+0.20$ & $+0.57$ & $+0.16$ & $+0.30$ & $+0.21$ & $+0.30$ & $+0.03$ & $+0.31$ & $+0.25$ & $-3.11$ \\
2526 & $+1.01$ & $+1.08$ & $+1.08$ & $+3.21$ & $+0.82$ & $+0.65$ & $+1.65$ & $+0.12$ & $+1.00$ & $-0.47$ & $-1.39$ & $-2.91$ & $-4.82$ \\
2522 & $+1.13$ & $-0.39$ & $+0.22$ & $+0.26$ & $+0.23$ & $+0.71$ & $+0.23$ & $+0.21$ & $+0.19$ & $+0.13$ & $+0.26$ & $+0.25$ & $-2.29$ \\
\bottomrule
\end{tabular}}
\end{table}

\section{Capability profiles}
\label{app:profiles}\label{app:regression_full}\label{app:radar_predicted_alpha}

\begin{figure}[H]
\centering
\includegraphics[width=\textwidth]{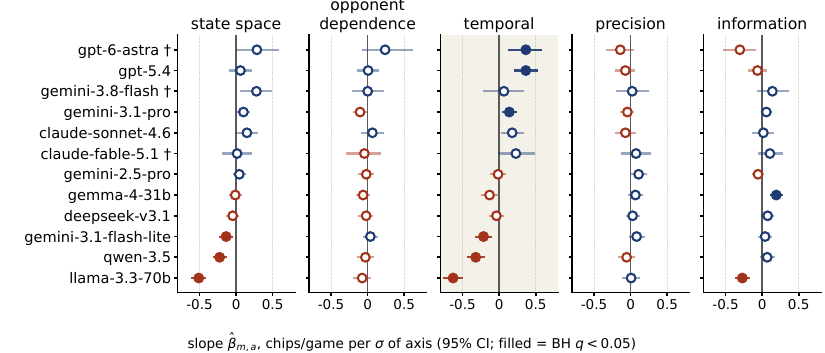}
\caption{\textbf{Capability profile with intervals.} For each model, we regress its game strength $\hat\alpha_{m,g}$
on all five $z$-scored axes jointly, with an intercept, over the 50 games. Each panel shows the coefficients for one
axis, with models in leaderboard order. Coefficients (chips/game per SD of the axis) are positive for models whose
advantage over the average increases as a game increases in that axis value. Bars are 95\% clustered bootstrap
CIs ($B{=}2{,}000$). Filled markers (11 of 60 cells, 6 on the shaded temporal panel) are significant after applying
Benjamini--Hochberg correction at $q<0.05$. Blue markers are positive coefficients and red markers negative.
$^\dagger$ September models.}
\label{fig:profile_intervals}
\end{figure}

\begin{figure}[H]
\centering
\includegraphics[width=0.95\textwidth]{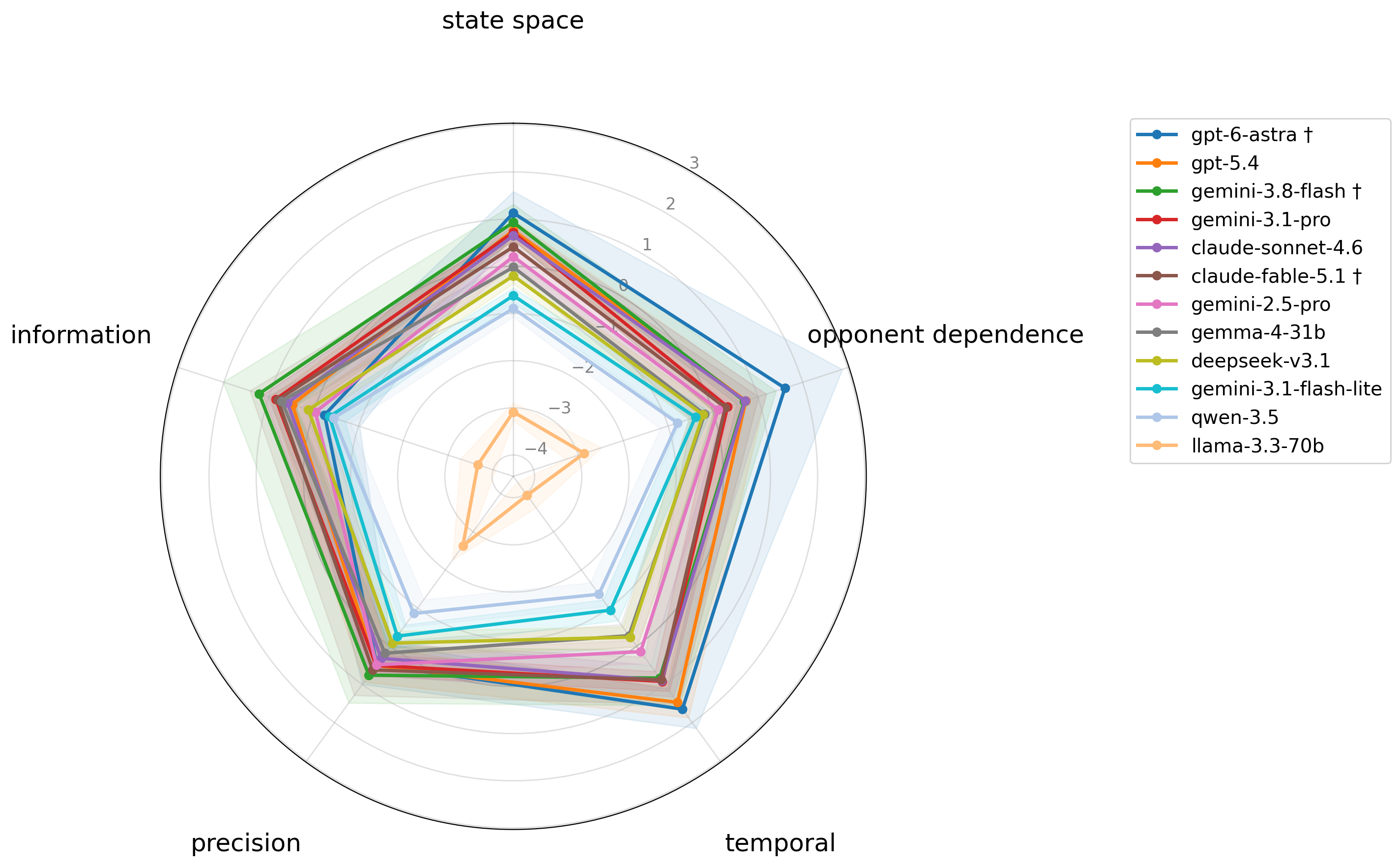}
\caption{\textbf{Capability profile in absolute units.} The profile regression of Figure~\ref{fig:radar} is used
to predict each model's per-game strength $\hat\alpha_{m,g_a^*}$ on a synthetic game. On that game axis $a$
takes its 50-game maximum $z$-score and the other four axes take their 50-game medians. Units are chips/game
relative to the across-model mean. The shaded bands are 95\% paired-cluster bootstrap CIs ($B{=}2{,}000$)
propagated through the linear predictor.}
\label{fig:radar_predicted_alpha}
\end{figure}

For each model $m$ we fit, over the 50 games and with all five $z$-scored axes entered jointly, the profile
regression $\hat\alpha_{m,g} = \beta_{m,0} + \sum_{a=1}^{5}\beta_{m,a} z_a(g) + \varepsilon_{m,g}$.
Table~\ref{tab:per_axis_regression} has the point estimates and Figure~\ref{fig:profile_intervals} plots them with
their paired-cluster bootstrap intervals ($B{=}2{,}000$). Benjamini--Hochberg correction at $q<0.05$ across the 60
coefficients leaves eleven, which are in bold. Figure~\ref{fig:radar_predicted_alpha} shows the same profiles in chips per game, on a game at the extreme of one axis.

\begin{table}[H]\centering\scriptsize
\caption{Per-model OLS slopes of $\hat\alpha_{m,g}$ on the five
$z$-scored axes, fitted jointly. Bold =
BH-adjusted $q < 0.05$ across the 60 coefficients.}
\label{tab:per_axis_regression}
\setlength{\tabcolsep}{4pt}
\begin{tabular}{lrrrrr}
\toprule
 & {\scriptsize state-sp.} & {\scriptsize opponent} & {\scriptsize temporal} & {\scriptsize precision} & {\scriptsize information} \\\midrule
\texttt{gpt-6-astra}$^\dagger$ & $+0.28$ & $+0.24$ & $\mathbf{+0.37}$ & $-0.14$ & $-0.30$ \\
\texttt{gpt-5.4} & $+0.06$ & $+0.01$ & $\mathbf{+0.37}$ & $-0.07$ & $-0.06$ \\
\texttt{gemini-3.8-flash}$^\dagger$ & $+0.28$ & $+0.00$ & $+0.07$ & $+0.02$ & $+0.14$ \\
\texttt{gemini-3.1-pro} & $+0.10$ & $-0.10$ & $\mathbf{+0.14}$ & $-0.04$ & $+0.06$ \\
\texttt{claude-sonnet-4.6} & $+0.15$ & $+0.07$ & $+0.18$ & $-0.07$ & $+0.02$ \\
\texttt{claude-fable-5.1}$^\dagger$ & $+0.01$ & $-0.04$ & $+0.23$ & $+0.07$ & $+0.11$ \\
\texttt{gemini-2.5-pro} & $+0.04$ & $-0.02$ & $-0.01$ & $+0.11$ & $-0.06$ \\
\texttt{gemma-4-31b} & $-0.01$ & $-0.06$ & $-0.13$ & $+0.07$ & $\mathbf{+0.19}$ \\
\texttt{deepseek-v3.1} & $-0.05$ & $-0.02$ & $-0.04$ & $+0.03$ & $+0.08$ \\
\texttt{gemini-3.1-flash-lite} & $\mathbf{-0.14}$ & $+0.04$ & $\mathbf{-0.21}$ & $+0.08$ & $+0.04$ \\
\texttt{qwen-3.5} & $\mathbf{-0.22}$ & $-0.03$ & $\mathbf{-0.32}$ & $-0.06$ & $+0.07$ \\
\texttt{llama-3.3-70b} & $\mathbf{-0.51}$ & $-0.07$ & $\mathbf{-0.63}$ & $+0.01$ & $\mathbf{-0.27}$ \\
\bottomrule\end{tabular}
\end{table}

In Figures~\ref{fig:radar} and~\ref{fig:profile_intervals} the per-(model, axis) slope $\hat\beta_{m,a}$ is in
chips/game per $\sigma$ of axis, the local sensitivity of model $m$'s edge to axis $a$. The companion plot below, with each spoke in chips/game on a benchmark-extreme game
for that axis instead of a per-$\sigma$ slope, comes from the same fitted regression. We evaluate each
model's predicted per-game strength on a synthetic game where axis $a$ is at its observed maximum z-score
and the other four axes are at their observed medians,
\[
\hat\alpha_{m,g_a^*} \;=\; \hat\beta_{m,0}
   + \hat\beta_{m,a}\,z_a^{\max}
   + \!\!\sum_{a'\neq a}\!\! \hat\beta_{m,a'}\,z_{a'}^{\text{med}}.
\]

\paragraph{Robustness to rulebook-based confounds.} Verbosity is the most direct way a rulebook could in
principle confound the slopes. Rulebooks run from roughly 8{,}600 characters to just under 39{,}000.
Adding the logarithm (base 10) of the character count as a regressor to the regression of
Table~\ref{tab:per_axis_regression}, we find that the slopes are robust to rulebook verbosity. All eleven
cells significant after BH correction in Figure~\ref{fig:profile_intervals} keep their sign, and the temporal slopes
change by at most $0.09$ chips/game. The added term mainly moves the state space slopes, with which log length
correlates at $\rho=0.82$ (no other slope moves by more than $0.09$). After BH correction across the twelve models it
is significant for two of them. With a coefficient of $-0.41$ per standard deviation of log length for \texttt{llama-3.3-70b}, it moves the state space deficit of
\texttt{llama-3.3-70b} from $-0.51$ to $-0.20$, so part of what this axis captures for that model is rulebook length.
With $+0.46$ for \texttt{qwen-3.5}, it deepens the state space deficit of \texttt{qwen-3.5} from $-0.23$
to $-0.57$.

\paragraph{Where the ordering departs from the leaderboard.} On seed 4968, in the easiest tertile
(Appendix~\ref{app:tertile_leaderboards}), astra has a strength of $-0.95$ chips/game
(Table~\ref{tab:per_cell_alpha}), although it ranks first overall. Games whose ranking departs from the
leaderboard in this way are mostly small, and most of the departure on them is compression of the differences between models. On the five games where
the departure exceeds sampling noise (862, 1949, 2522, 11517 and 11520), opponent dependence is highest (mean $z$ of
$+1.76$), and precision and information are also above average. Seed 4968 is not one of them, but it shows the
mechanism clearly. In this game of three cards a
position publicly reveals the holder of $Q$, so a bet into $Q$ comes from $J$ or $K$. The traces of
\texttt{gpt-5.4} and astra both derive the sequential equilibrium, in which $K$ is bet for value, $J$
is bluffed with probability $9/13$, and $Q$ is exactly indifferent. In the tournament the bettor held $K$
almost always, and \texttt{gpt-5.4} folds $Q$, which happens to be right against a field that never
bluffs. Astra bluffs against llama, which calls with $Q$ every time, and loses six chips per bluff.
\texttt{deepseek-v3.1} also calls every time, and its trace puts probability $1/2$ on each of $J$
and $K$ and computes a positive expected value. In a single deal (Section~\ref{sec:setup}) what separates
these models is the prior in each trace about the opponent, and whether a prior is a failure depends on the
field.

\subsection{How many patterns the profiles share}
\label{app:factor_rank}

We choose the number of common factors in the profiles of Section~\ref{sec:profiles} by parallel analysis
\citep{horn1965rationale,buja1992remarks} of $\hat A$, the $12\times50$ matrix of the strength
$\hat\alpha_{m,g}$ of each model on each game. Table~\ref{tab:factor_rank} reports the analysis for all twelve
models ($20{,}206$ paired deals). It also covers the nine April models, estimated from April deals only
($18{,}436$), and the eleven models other than \texttt{llama-3.3-70b} ($15{,}228$). With all twelve models,
the first factor accounts for 38\% of the standardized variance and the 95th percentile of the null with no
common factor is 33\% ($p{=}0.003$). The second factor accounts for 16\%, compared with 18\% under the null of
exactly one factor ($p{=}0.19$). The April subset gives $p{<}0.004$ (none of 300 null draws) and $p{=}0.10$. Without
\texttt{llama-3.3-70b} the two $p$-values are $0.013$ and $0.37$. The game scores of the first factor are
highly correlated across the three sets of models ($|\rho|\ge0.93$), and their correlation with the temporal
axis lies between $0.65$ and $0.73$ (with state space, $+0.65$ in the full set). After the first factor is removed, no pairwise residual correlation
between models is significant under Benjamini--Hochberg correction \citep{benjamini1995controlling}. The test has
limited power, and weaker patterns may go undetected at fifty games.

Horn's procedure compares the observed eigenvalues with those of random normal matrices. We use bootstrap
replicates of the estimated strengths for the reference distribution. Let $\tilde A$ be $\hat A$ with each row
centered and scaled to unit variance across games, with singular value decomposition
$\tilde A=\sum_k s_k u_k v_k^\top$. Factor $k$ accounts for a share $s_k^2/\sum_j s_j^2$ of the standardized
variance. The entries of $v_k$ are its game scores, and $u_k$ holds its model loadings. We draw $B{=}300$
bootstrap replicates of $\hat A$, in which the two seats of a run form one cluster, and subtract the
bootstrap mean from each. The resulting 300 matrices of estimation noise are added to the row means of
$\hat A$ under the null of no common factor, and to the rank one approximation of $\hat A$ in chips under
the null of exactly one factor. Each null matrix is then standardized and decomposed in the same way as
$\tilde A$. The $p$-value for factor $k$ is the fraction of null matrices in which the $k$th factor explains at least as large a share of the variance as the $k$th factor of $\tilde A$. Factors are retained in order until the first one that is not significant at the
5\% level.

\begin{table}[H]\centering\small
\caption{\textbf{Parallel analysis of the standardized model$\times$game strength matrix.} Columns give the share of
variance explained by each of the first three factors, the $p$-values $p_1$ against the null of no
common factor and $p_2$, $p_3$ against the null of exactly one factor, and the Spearman correlation between
the first factor's game scores and temporal contingency. The nulls use $B{=}300$ bootstrap replicates, clustered by run.}
\label{tab:factor_rank}
\setlength{\tabcolsep}{4pt}
\begin{tabular}{lrccccc}
\toprule
\textbf{Subset} & \textbf{deals} & \textbf{variance share} & $p_1$ & $p_2$ & $p_3$ & $\rho_{\text{temporal}}$ \\
 & & \textbf{factors 1 / 2 / 3} & & & & \\
\midrule
full twelve & 20,206 & $0.38$ / $0.16$ / $0.11$ & $0.003$ & $0.19$ & $0.35$ & $+0.73$ \\
April nine, April deals & 18,436 & $0.43$ / $0.18$ / $0.12$ & $<0.004$ & $0.10$ & $0.37$ & $+0.72$ \\
llama excluded & 15,228 & $0.39$ / $0.17$ / $0.14$ & $0.013$ & $0.37$ & $0.11$ & $+0.65$ \\
\bottomrule
\end{tabular}
\end{table}

\section{Local jaggedness}
\label{app:jaggedness_formal}

For model $m$ on game $g$, set $y_m(g) = (\hat\alpha_{m,g}-\bar\alpha_{m\cdot})/\sigma_U(g)$, subtracting
the model's mean strength over games and dividing by the standard deviation of the final payoff when both
seats move uniformly at random, as in the state-space count of Appendix~\ref{app:axis_formals}. Denote the bootstrap standard error on this scale by
$\mathrm{se}_m(g)$. A Gaussian kernel ridge fit over the five $z$-scored axes gives the smooth surface
$f_m$. Both its length scale and ridge penalty are chosen by leave-one-game-out cross-validation. We use
exact residuals $r_m(g) = y_m(g) - f_{m,-g}(x_g)$, with game $g$ omitted from its comparison surface, to calculate
\[
J_m \;=\; \sqrt{\max\Big(0,\ \mathrm{var}_g\,r_m(g) - \mathrm{mean}_g\,\mathrm{se}_m(g)^2\Big)},
\]
and refit on $B=500$ bootstrap samples of games for the 95\% intervals. Table~\ref{tab:jaggedness_local}
pairs $J_m$ with the surface's leave-one-game-out $R^2$. Both can be small when a model varies little
across games, as for \texttt{deepseek-v3.1} ($0.05$, $-0.02$).

\paragraph{Validation.} Independent jumps of size $0.20$ give $J_m=0.20$ in synthetic data. A trend mixed
with jumps of size $0.14$ gives $0.15$. We used the 50 benchmark games and 200 draws per case with
\texttt{gpt-5.4}'s sampling noise. The flat truth has zero variation, with an estimated $J_m$ of $0.03$.
For the other cases we set the true SD across games at about $0.20$. Linear dependence on one axis or two
gives $0.05$ in either case. With curvature in the temporal axis, the estimate is $0.11$, since the surface
removes only part of this trend. Across draws, the SD of $J_m$
is about $0.04$, or $0.06$ with a September model's sampling noise.

\paragraph{Resolution.} The September models had only one opponent on most games. The widest intervals
belong to \texttt{gpt-6-astra} and \texttt{gemini-3.8-flash}, both $[0.00, 0.18]$. Among the nine April models, the
orderings from random halves of each game's paired deals have a Spearman correlation of $0.57$. We cannot resolve
differences below about $0.1$. Among the April models, both halves rank \texttt{llama-3.3-70b} first
($0.34$ and $0.36$) and \texttt{qwen-3.5} second ($0.18$ and $0.16$).

\begin{table}[H]\centering\small
\caption{Local jaggedness $J_m$ with 95\% game-bootstrap intervals. We also report $R^2$ of a
model's smooth surface over the five axes (LOO CV). $^\dagger$ denotes the September models.}
\label{tab:jaggedness_local}
\begin{tabular}{lcc}
\toprule
\textbf{Model} & \textbf{local jaggedness} $J_m$ [95\% CI] & \textbf{smooth-surface} $R^2$ \\
\midrule
\texttt{llama-3.3-70b} & $0.35$ $[0.27, 0.42]$ & $+0.44$ \\
\texttt{qwen-3.5} & $0.17$ $[0.13, 0.21]$ & $+0.24$ \\
\texttt{gpt-5.4} & $0.12$ $[0.05, 0.17]$ & $+0.35$ \\
\texttt{gemini-3.1-flash-lite} & $0.12$ $[0.09, 0.14]$ & $+0.33$ \\
\texttt{gpt-6-astra}$^\dagger$ & $0.11$ $[0.00, 0.18]$ & $+0.32$ \\
\texttt{claude-sonnet-4.6} & $0.10$ $[0.03, 0.15]$ & $+0.14$ \\
\texttt{gemini-3.8-flash}$^\dagger$ & $0.09$ $[0.00, 0.18]$ & $+0.26$ \\
\texttt{gemma-4-31b} & $0.08$ $[0.03, 0.11]$ & $+0.11$ \\
\texttt{gemini-3.1-pro} & $0.08$ $[0.03, 0.10]$ & $+0.37$ \\
\texttt{gemini-2.5-pro} & $0.05$ $[0.00, 0.09]$ & $+0.10$ \\
\texttt{deepseek-v3.1} & $0.05$ $[0.00, 0.08]$ & $-0.02$ \\
\texttt{claude-fable-5.1}$^\dagger$ & $0.00$ $[0.00, 0.08]$ & $+0.01$ \\
\bottomrule
\end{tabular}
\end{table}

\section{Ablations and trace analysis}
\label{app:ablations}\label{app:reasoning}\label{app:rubric}\label{app:programs}

\subsection{Strategy programs}

\paragraph{Programs.} From the readings alone, independent readers predicted the move in all 755 sampled
situations of the 100 programs written by \texttt{claude-sonnet-4.6}, and in all 45 of a re-test of six revised
readings. We generate each reading as a procedural description of the program in natural language.
In disclosed play, LLMs receive both this reading and the program in its strategy language.
The strategy language allows deterministic and randomized branches, using only the observable
state of the game. Four of the 40 programs of \texttt{gemma-4-31b} are the default program, which always takes the
first option, used after two failed attempts to parse its output (Bob in 10427, both seats in 10760, Alice in 10790).

\paragraph{Design.} A pair uses one play seed in both seats. We take the pair's average chip margin as
a single observation. In disclosed play, the program accompanies the rulebook in the LLM's prompt.
In hidden play, the program and its reading are omitted. The efficiency
reported in Table~\ref{tab:programs} uses the worst response
as $0$ and the best as $1$, so $0.5$ is their midpoint and is not neutral performance.

\begin{table}[H]
\centering\small
\caption{\textbf{Programmed opponent ablation.} 520 complete pairs. Slope coefficients relate players' chip margins
against the programs to the tournament ratings of Table~\ref{tab:leaderboard}.
We resample deals, $B{=}300$ times for slopes
and $B{=}2{,}000$ for disclosure effects and efficiencies.}
\label{tab:programs}
\begin{tabular}{lcc}
\toprule
\textbf{Quantity} & \textbf{Estimate} & \textbf{95\% CI} \\
\midrule
$\lambda_{\text{disclosed}}$ (five players) & $+1.10$ & $[+0.76, +1.41]$ \\
$\lambda_{\text{hidden}}$ (three players) & $+0.92$ & $[+0.24, +1.58]$ \\
$\lambda_{\text{disclosed}} - \lambda_{\text{hidden}}$ & $+0.18$ & $[-0.40, +0.75]$ \\
Disclosure effect, chips/game (160 deals) & $+0.72$ & $[+0.19, +1.30]$ \\
\quad \texttt{gpt-5.4} & $+0.87$ & $[+0.27, +1.53]$ \\
\quad \texttt{claude-sonnet-4.6} & $+0.90$ & $[-0.23, +2.08]$ \\
\quad \texttt{llama-3.3-70b} & $+0.45$ & $[-0.55, +1.53]$ \\
\midrule
\multicolumn{3}{l}{\emph{Play efficiency on the five exactly solved games ($0$ = worst, $1$ = best response)}} \\
 & \textbf{disclosed} & \textbf{hidden} \\
\texttt{gpt-5.4} & $1.02$ $[0.82, 1.28]$ & $0.81$ $[0.70, 0.90]$ \\
\texttt{claude-sonnet-4.6} & $1.01$ $[0.80, 1.25]$ & $0.71$ $[0.53, 0.84]$ \\
\texttt{gemini-3.1-pro} & $0.98$ $[0.81, 1.15]$ & not played \\
\texttt{gemma-4-31b} & $0.97$ $[0.84, 1.09]$ & not played \\
\texttt{llama-3.3-70b} & $0.53$ $[0.37, 0.72]$ & $0.53$ $[0.24, 0.79]$ \\
\bottomrule
\end{tabular}
\end{table}

\subsection{Thinking budget}
\label{sec:ablation}

We estimate the expected gain in chip margin within a pair from low to high effort,
$\Delta = \mathbb{E}[\text{chip margin}_\text{high}-\text{chip margin}_\text{low}]$.
Our estimate $\hat\Delta$ uses the $260$ sibling pairs from seven (family, anchor) cells that were played on the
benchmark games. For 9 of its seeds the block served the default-parameter variant of the seed from a cache instead of the
benchmark game, and we leave those matches out. Confidence intervals
(95\%) are estimated using a cluster bootstrap on game seed. The two anchors are
\texttt{gemini-3.1-pro} (a frontier model near the top of the leaderboard) and
\texttt{gemma-4-31b} (an open model from the middle of the pack).

\begin{table}[H]
\centering\small
\caption{\textbf{Thinking ablation.} Paired gains in chip margin by family. Significant estimates are starred.}
\label{tab:ablation}
\begin{tabular}{lrcr}
\toprule
\textbf{Model family} & $\hat\Delta$ & \textbf{95\% CI} & \textbf{pairs} \\
\midrule
\texttt{claude-sonnet-4.6} & $+0.37$ & $[-0.31, +1.21]$ & 104 \\
\texttt{gemini-2.5-pro} & $+0.06$ & $[-1.83, +2.00]$ & 32 \\
\texttt{gemini-3.1-pro}$^*$ & $+0.53$ & $[+0.40, +0.60]$ & 30 \\
\texttt{gpt-5.4} & $+0.19$ & $[+0.00, +0.36]$ & 94 \\
\bottomrule
\end{tabular}
\\[2pt]\footnotesize $^*$ 95\% CI excludes 0.
\end{table}

All four estimates in Table~\ref{tab:ablation} are positive, from $+0.06$ for \texttt{gemini-2.5-pro} to $+0.53$
for \texttt{gemini-3.1-pro}. Only the estimate for \texttt{gemini-3.1-pro} is significant, and it rests on three
games. The interval for \texttt{gpt-5.4} touches zero.

\subsection{Reasoning traces}
\label{sec:rubric}

Reasoning about the hand is worth $+1.15$ chips per rubric point more on games in the top tertile of the information
axis than on other games (SE $0.38$, $p=0.002$), and planning depth $+0.93$ more in the top tertile of the temporal
axis (SE $0.34$, $p=0.007$). Here the focal chip edge of a move, the final margin of the model that made it in that
match, is regressed on the rubric scores defined below, each interacted with an indicator for the top tertile of its
matched axis. Like every regression in this section, it has model fixed effects and standard errors clustered by game.
A control for the same interaction with state space gives $+1.06$ ($p=0.001$) and $+0.71$ ($p=0.046$). For the
earlier six axes neither interaction is significant ($+0.69$, $p=0.10$, and $+0.40$, $p=0.30$). Comparing top and
bottom tertiles gives the same answer, $+0.59$ (SE $0.20$) and $+0.69$ (SE $0.18$), against $+0.08$ (SE $0.21$) for
opponent modeling. Across games, only the opponent axis predicts its matched rubric ($+0.04$ points per $\sigma$,
$q=0.03$). The rubric scores are lower on larger games, by $0.07$ points per $\sigma$ of state space for opponent
modeling and $0.10$ for planning depth (both $q<0.001$). Note that all of
these estimates are associations. For example, a trace that reasons about the hand often comes from a move where
the hand is decisive.

\paragraph{Rubric.} For each sampled move, the judge (\texttt{gemini-3-flash-preview}) is given the game's rulebook, the agent's prompt, the agent's full response, and the parsed action. The prompt is the game context, rendered
deterministically, and we treat it as ground truth. The judge scores the reasoning visible in the response on
three ordinal $0$--$3$ dimensions, each matched to one value axis. Move quality is not scored. The judge prompt also contained a fourth dimension, risk reasoning, which matched the risk axis of the
earlier selection basis, and we do not use its scores. Anchors:

\begin{center}\small
\begin{tabular}{lp{0.63\linewidth}}
\toprule
\textbf{Dimension} & \textbf{Anchors (0 / 1 / 2 / 3)} \tabularnewline
\midrule
Opponent-modeling engagement & ignores the opponent / opponent mentioned but treated as fixed / forms an
explicit model of the opponent's likely action / recursive: anticipates the opponent's anticipation \tabularnewline
Private-information reasoning & does not use own hand as evidence / hand named, possibly with a brief strength
label / analysis along one line: combines two private sources, or reasons about win rate or expected value / both
lines, multi-source and quantitative \tabularnewline
Forward-planning depth & myopic / mentions future turns without conditions / one conditional fork about a future
state / two or more chained or parallel conditionals \tabularnewline
\bottomrule
\end{tabular}\\[2pt]
{\footnotesize The sample covers thirteen model variants, among them the variants with low reasoning
effort from the thinking ablation. The three dimensions are matched to opponent dependence, information,
and temporal contingency, in the order listed.}
\end{center}

The anchors follow how traces are coded in the poker and game reasoning literature, including the
theory of mind coding of \citet{lin2026readableminds}. We drew 10{,}740 moves from the tournament and the thinking ablation, stratified by model variant, capability tier
and quadrant of the earlier opponent and temporal axes. Of these, $10{,}721$ were parsed. We analyse the $9{,}778$
whose game is the benchmark game (101 to 372 per game). The 943 moves from ablation matches played on a
default-parameter variant are left out (Appendix~\ref{sec:ablation}). A subset was also scored by a second judge,
\texttt{gemini-3.1-pro}, which gives the same signs for the private-information and planning interactions on the five axes.

\subsection{Judge robustness}
\label{app:judges}

\begin{figure}[h]
\centering
\includegraphics[width=0.62\linewidth]{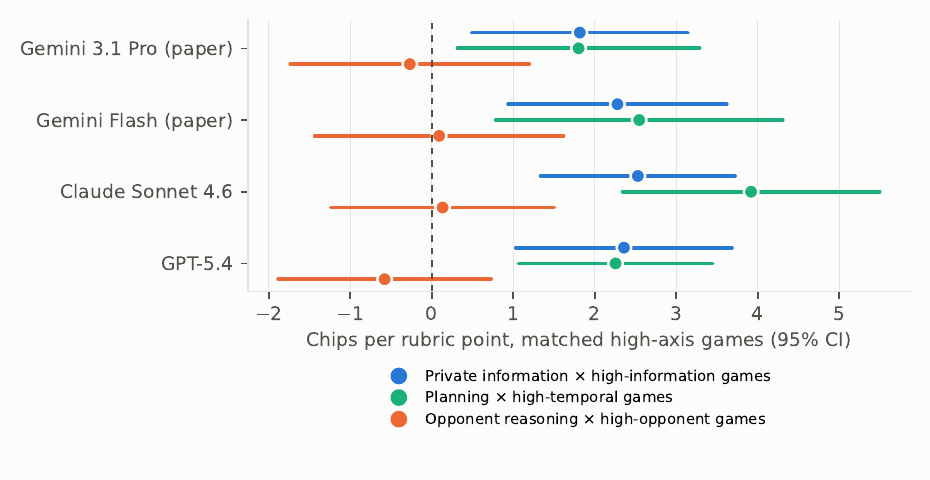}
\caption{\textbf{The trace findings under four judges.} Chips per rubric point on matched high-axis games (95\% CI),
for each judge and rubric.}
\label{fig:judges}
\end{figure}

The analysis above uses judges from one family. We re-scored the same 869 traces with four judges from three families
(Gemini, Claude, GPT) and re-estimated the regression of chip margin on rubric score for each (Figure~\ref{fig:judges}).
Under all four judges, reasoning about private information is associated with higher margins on high-information games
and reasoning about future play on high-temporal games, while reasoning about the opponent stays null. Judges from different
families agree with each other as closely as two judges from the same family do, and all four rank the 13 model
variants in almost the same order.

\section{Data notes}
\label{app:dataquality}\label{app:fallback}

\paragraph{Parser fallbacks.} We did not need a fallback for seven of the twelve models. The highest
rate among the others was $0.33\%$ of moves. If neither the strict nor the lenient parser recovers a
legal action, the engine plays a uniformly random legal action and records the fallback.

\begin{center}\small
\begin{tabular}{lrrrrr}
\toprule
\textbf{Model} & \textbf{matches} & \textbf{moves} & \textbf{fallbacks}
& \textbf{move rate} & \textbf{match rate} \\
\midrule
\texttt{gemini-3.1-flash-lite} & 9,850  & 23,734 & 79 & 0.33\% & 0.8\% \\
\texttt{gpt-5.4}          & 4,652  & 11,021 & 13 & 0.12\% & 0.3\% \\
\texttt{gemini-3.1-pro}        & 10,490 & 24,520 & 9  & 0.04\% & 0.1\% \\
\texttt{gemini-2.5-pro}        & 9,400  & 21,975 & 9  & 0.04\% & 0.1\% \\
\texttt{deepseek-v3.1}         & 9,398  & 22,682 & 8  & 0.04\% & 0.1\% \\
\texttt{gpt-6-astra}      & 1,380  & 3,809  & 0  & 0.00\% & 0.0\% \\
\texttt{gemini-3.8-flash}      & 1,000  & 2,194  & 0  & 0.00\% & 0.0\% \\
\texttt{claude-fable-5.1} & 1,360  & 3,628  & 0  & 0.00\% & 0.0\% \\
\texttt{claude-sonnet-4.6}     & 4,078  & 9,272  & 0  & 0.00\% & 0.0\% \\
\texttt{gemma-4-31b}           & 9,834  & 23,497 & 0  & 0.00\% & 0.0\% \\
\texttt{qwen-3.5}              & 9,426  & 21,559 & 0  & 0.00\% & 0.0\% \\
\texttt{llama-3.3-70b}         & 9,956  & 25,945 & 0  & 0.00\% & 0.0\% \\
\bottomrule
\end{tabular}
\end{center}

\paragraph{Rulebook fidelity audit.} Eight independent LLM auditors and the authors checked all 50 served rulebooks
against the engine and found no engine defect. The later checks of Appendix~\ref{app:validation} found one, now
fixed. In 32 rulebooks the text leaves out a detail that the engine resolves
itself: that an auction prize swaps or peeks at a random card (14 and 12 games), that a bonus prize is capped at the
losing bidder's stack (9), or that a forced discard or a forced play takes the card received first (9 and 3). In 12
rulebooks it also says that chip costs in hidden menus are settled at the end of the game, whereas the engine deducts
them at once and the opponent sees the new stack. In all, 34 of the 50 rulebooks are affected. Every defect is
symmetric across seats, since both players read the same rulebook and each model plays both seats of every deal. We
refit the leaderboard without the games of each class in turn. The three tiers of Table~\ref{tab:leaderboard} are
unchanged, models change places only within their tier, and no rating moves by more than $0.28$ chips/game. Because
each defect goes with one mechanic, this test cannot separate an effect of the text from an effect of the mechanic.
Appendix~\ref{app:validation} tests the text directly.

\paragraph{Rerun and retained matches.} During the project we found and fixed two bugs: a rulebook
rendering defect in one April tournament session, and a parser bug that could turn a legal answer into
a random fallback. We ran 45 affected matches again on the same play seeds and replaced them.
The rendering defect also altered play on 21 matches, which are excluded from every analysis.
Another 58 could not be run again and were retained as played ($0.14\%$ of the corpus). They occur
in 54 paired deals. If we drop all 54, the largest change to a rating in Table~\ref{tab:leaderboard}
is about $0.010$ chips/game. Both repair logs are distributed with the data, along with the lists
of excluded and retained matches.

\section{Models and identifiers}
\label{app:models}

Table~\ref{tab:model_ids} lists the provider identifier of each model. Every request setting is stored with the data.
Runs made after the tournament (Appendices~\ref{app:solved} and~\ref{app:oos}) reach the three
open-weight models through a different provider.

\begin{table}[h]
\centering\small
\caption{\textbf{Models.} $^\dagger$ September block.}
\label{tab:model_ids}
\begin{tabular}{lll}
\toprule
Short name & Provider identifier & Block \\
\midrule
\texttt{gpt-6-astra}$^\dagger$ & \path{gpt-6-astra-high} & September 2026 \\
\texttt{gpt-5.4} & \path{gpt-5-4-high} & April--May 2026 \\
\texttt{gemini-3.8-flash}$^\dagger$ & \path{gemini-3-8-flash} & September 2026 \\
\texttt{gemini-3.1-pro} & \path{gemini-3.1-pro-preview} & April--May 2026 \\
\texttt{claude-sonnet-4.6} & \path{claude-sonnet-4-6-max} & April--May 2026 \\
\texttt{claude-fable-5.1}$^\dagger$ & \path{claude-fable-5-1-high} & September 2026 \\
\texttt{gemini-2.5-pro} & \path{gemini-2.5-pro} & April--May 2026 \\
\texttt{gemma-4-31b} & \path{gemma-4-31b-it} & April--May 2026 \\
\texttt{deepseek-v3.1} & \path{deepseek-v3.1-together} & April--May 2026 \\
\texttt{gemini-3.1-flash-lite} & \path{gemini-3.1-flash-lite-preview} & April--May 2026 \\
\texttt{qwen-3.5} & \path{qwen-3.5-together} & April--May 2026 \\
\texttt{llama-3.3-70b} & \path{llama-3.3-70b-together} & April--May 2026 \\
\bottomrule
\end{tabular}
\end{table}

\section{Play against the exact equilibrium}
\label{app:solved}

\paragraph{Solving the games.} We solved 27 of the 50 benchmark games, 26 of them exactly and one to within a negligible
error. Most of the other 23 are too large to solve on one machine, since sampling just one of the hardest already
visits millions of distinct states. On the games where the benchmark separates models most, play between models is
therefore the only feasible evaluation.

\begin{figure}[h]
\centering
\includegraphics[width=\linewidth]{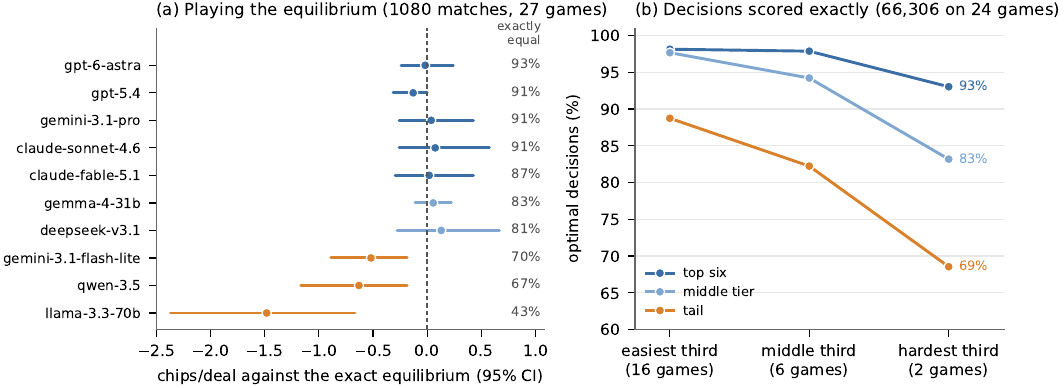}
\caption{\textbf{Play against the exact equilibrium.} (a) What each model wins or loses per deal against the
equilibrium on the 27 solved games, playing both seats of the same deals, with 95\% CIs. (b) How often each tier picks
an action with zero regret, by the composite hardness of the game, over 66,306 decisions on 24 games.}
\label{fig:solver}
\end{figure}

\paragraph{Matches against the equilibrium.} Ten models played both seats of the same deals against the equilibrium on
all 27 solved games (Figure~\ref{fig:solver}a). The frontier models break even with it, so almost every move they make is a best response to it. The three
weakest lose almost a chip per deal, and no model beats it.

\begin{figure}[h]
\centering
\includegraphics[width=0.84\linewidth]{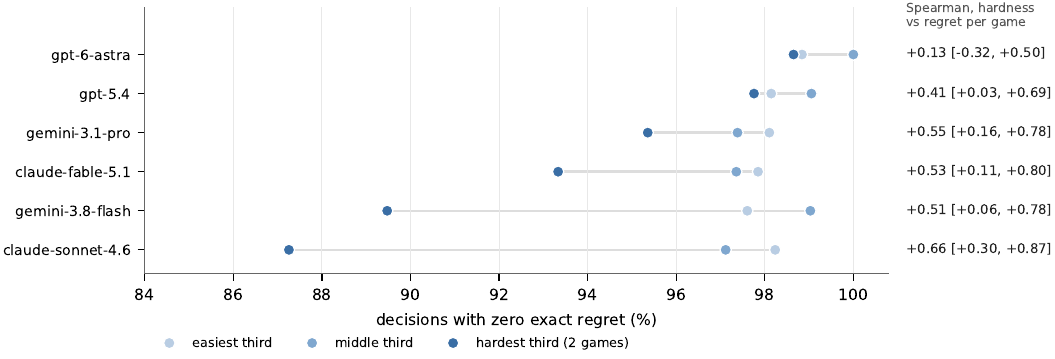}
\caption{\textbf{Easy games saturate while hard games separate models effectively.} Left, how often each frontier model picks a
zero-regret action in the easiest, middle and hardest third of the 24 scored games. Right, the Spearman correlation
between a game's hardness and the model's mean regret per decision on it, with a 95\% CI from resampling games.}
\label{fig:frontier_split}
\end{figure}

\paragraph{Every decision scored exactly.} We also scored each tournament decision on 24 of the solved games by its
regret, the chips it gives up against the best action when the rest of the deal is played at equilibrium. Every tier
picks fewer zero-regret actions as the games get harder (Figure~\ref{fig:solver}b). On the easiest games the frontier
models are hard to tell apart, and they only separate on the harder ones (Figure~\ref{fig:frontier_split}).

\begin{figure}[h]
\centering
\includegraphics[width=0.62\linewidth]{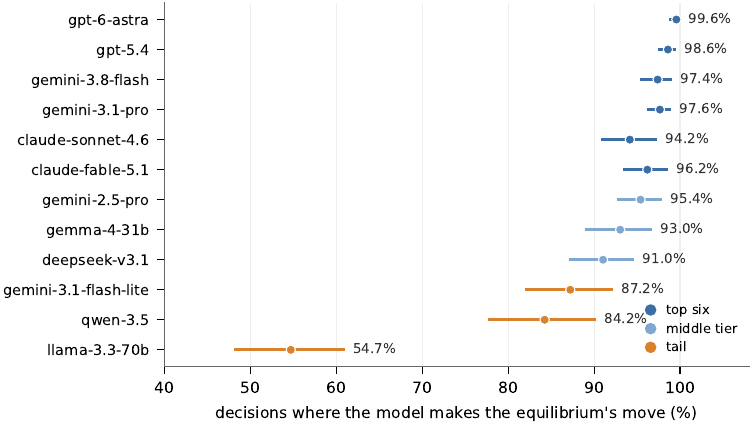}
\caption{\textbf{How often each model makes the equilibrium's move.} We count only decisions on the 24 scored
games where no move outside the equilibrium's support is also optimal, about 37,000 in all. Lines are 95\% CIs from
resampling games, and models are in leaderboard order.}
\label{fig:eq_move}
\end{figure}

\paragraph{One model plays an equilibrium action almost everywhere.} \texttt{gpt-6-astra} picks a zero-regret action
99\% of the time and never falls below 98.6\% in any hardness third, while most other frontier models drop to between
87\% and 95\% on the hardest games. As far as we can measure, it plays the solved games at the quality of the
equilibrium, though it does not randomize as the equilibrium does. Figure~\ref{fig:eq_move} shows, for every model, how
often it makes the equilibrium's move at the decisions where no move outside the equilibrium's support is also optimal.

\paragraph{Mixed strategies.} At the information sets where the equilibrium mixes, we gave each model the identical
prompt ten times. No model mixes the way the equilibrium does. The strongest models nearly always give the more common
action, and \texttt{gpt-6-astra} changes its answer at only 9\% of the sets. In these games that costs little. With every other move held at equilibrium, a best-responding opponent could
exploit the way a model plays at these sets by about 0.02 chips per deal against \texttt{gpt-6-astra}, 0.06 against
the top tier on average and 0.13 against the tail. Weaker models also more often settle on the rarer action.
Table~\ref{tab:mixing} gives each model's distance from the equilibrium mix, net of the distance that sampling alone
would produce.

\begin{table}[h]
\centering\small
\caption{\textbf{How far each model's answers lie from the equilibrium mix}, at information sets where every optimal
strategy mixes. Each entry is the total variation distance between the model's answers and the equilibrium mix, minus
the distance a perfect equilibrium mixer would show with as many answers, averaged over sets, with a 95\% CI from
resampling sets. Zero means the model cannot be told apart from the equilibrium mix. Tournament decisions reach each
set only a few times, so that column is much less precise. $^\dagger$ The Google calls used a fixed seed, so their ten
answers repeat.}
\label{tab:mixing}
\begin{tabular}{@{}l cc@{}}
\toprule
Model & Identical prompt, 10 answers & Tournament decisions \\
\midrule
\texttt{gpt-6-astra} & 0.19 [0.11, 0.30] & 0.02 [-0.03, 0.07] \\
\texttt{gpt-5.4} & 0.07 [-0.00, 0.15] & 0.07 [0.03, 0.12] \\
\texttt{gemini-3.8-flash} & 0.28 [0.21, 0.35]$^\dagger$ & -0.01 [-0.06, 0.06] \\
\texttt{gemini-3.1-pro} & 0.19 [0.12, 0.26] & 0.17 [0.12, 0.22] \\
\texttt{claude-sonnet-4.6} & 0.16 [0.01, 0.31] & 0.11 [0.05, 0.17] \\
\texttt{claude-fable-5.1} & 0.22 [0.08, 0.35] & 0.05 [-0.02, 0.13] \\
\texttt{gemini-2.5-pro} & --- & 0.13 [0.08, 0.18] \\
\texttt{gemma-4-31b} & 0.34 [0.26, 0.41]$^\dagger$ & 0.20 [0.14, 0.27] \\
\texttt{deepseek-v3.1} & 0.18 [0.11, 0.24] & 0.13 [0.07, 0.19] \\
\texttt{gemini-3.1-flash-lite} & 0.43 [0.35, 0.52]$^\dagger$ & 0.25 [0.19, 0.32] \\
\texttt{qwen-3.5} & 0.26 [0.18, 0.33] & 0.16 [0.10, 0.23] \\
\texttt{llama-3.3-70b} & 0.39 [0.29, 0.48] & 0.37 [0.29, 0.45] \\
\bottomrule
\end{tabular}
\end{table}

\section{Games outside the benchmark}
\label{app:oos}

We tested the leaderboard on three sets of games that no model had played before, each further from the benchmark
than the last: ten fresh games from the same generator, ten generated card games with no betting, and six new games
built on the \textsc{GENSTRAT} engine outside the betting-game generator. Table~\ref{tab:heldout} collects all of
them, with a general-capability leaderboard for comparison, and each set orders the models like the 50-game
leaderboard.

\begin{table}[h]
\centering\scriptsize
\caption{\textbf{Strength on every held-out set of games, and on a general-capability leaderboard.} 50 games: the
paper's leaderboard (chips/game). Fresh: ten new games from the generator, observed strength (chips/game). No bet:
ten generated games with no betting (chips/game). HA, MS, DR, Board, Market, Barg.: Hidden Agenda, Market Salad,
Declare Race, the board game, the market and bargaining (Appendix~\ref{app:families}), in units of each game's
random-play payoff SD. LMArena: text leaderboard score, style control on,
11 November 2026. In grey, each model's rank among the models the column covers. \texttt{gemini-2.5-pro} played no
held-out set and is not listed. Bottom rows: Spearman correlation of each column with the 50-game leaderboard, over the
models that column covers.}
\label{tab:heldout}
\setlength{\tabcolsep}{2.6pt}
\resizebox{\linewidth}{!}{\begin{tabular}{@{}l r r r r r r r r r r@{}}
\toprule
 & & \multicolumn{2}{c}{Generator} & \multicolumn{6}{c}{New games ($\hat\alpha_{m,g}/\sigma_U$)} & \\
\cmidrule(lr){3-4}\cmidrule(lr){5-10}
Model & 50 games & Fresh & No bet & HA & MS & DR & Board & Market & Barg. & LMArena \\
\midrule
\texttt{gpt-6-astra} & $+0.72$\,{\tiny\color{black!60}(1)} & $+0.82$\,{\tiny\color{black!60}(1)} & $+1.69$\,{\tiny\color{black!60}(1)} & $+1.46$\,{\tiny\color{black!60}(1)} & $+0.84$\,{\tiny\color{black!60}(2)} & $+0.97$\,{\tiny\color{black!60}(1)} & $+1.21$\,{\tiny\color{black!60}(2)} & $+2.16$\,{\tiny\color{black!60}(1)} & $+0.35$\,{\tiny\color{black!60}(1)} & 1474\,{\tiny\color{black!60}(5)} \\
\texttt{gpt-5.4} & $+0.64$\,{\tiny\color{black!60}(2)} & $+0.73$\,{\tiny\color{black!60}(2)} & $+0.44$\,{\tiny\color{black!60}(4)} & $+0.23$\,{\tiny\color{black!60}(4)} & $+1.28$\,{\tiny\color{black!60}(1)} & $+0.13$\,{\tiny\color{black!60}(5)} & $+0.70$\,{\tiny\color{black!60}(5)} & $+0.47$\,{\tiny\color{black!60}(5)} & $+0.28$\,{\tiny\color{black!60}(2)} & 1475\,{\tiny\color{black!60}(4)} \\
\texttt{gemini-3.8-flash} & $+0.63$\,{\tiny\color{black!60}(3)} & --- & --- & $+1.13$\,{\tiny\color{black!60}(2)} & $-0.06$\,{\tiny\color{black!60}(7)} & $+0.51$\,{\tiny\color{black!60}(3)} & $+0.84$\,{\tiny\color{black!60}(4)} & $+1.01$\,{\tiny\color{black!60}(4)} & $-0.00$\,{\tiny\color{black!60}(6)} & 1497\,{\tiny\color{black!60}(2)} \\
\texttt{gemini-3.1-pro} & $+0.63$\,{\tiny\color{black!60}(4)} & $+0.65$\,{\tiny\color{black!60}(4)} & $+0.48$\,{\tiny\color{black!60}(3)} & $+0.32$\,{\tiny\color{black!60}(3)} & $+0.11$\,{\tiny\color{black!60}(4)} & $+0.63$\,{\tiny\color{black!60}(2)} & $+1.00$\,{\tiny\color{black!60}(3)} & $+1.08$\,{\tiny\color{black!60}(2)} & $+0.02$\,{\tiny\color{black!60}(5)} & 1487\,{\tiny\color{black!60}(3)} \\
\texttt{claude-sonnet-4.6} & $+0.45$\,{\tiny\color{black!60}(5)} & $+0.52$\,{\tiny\color{black!60}(5)} & $-0.14$\,{\tiny\color{black!60}(7)} & --- & --- & --- & --- & --- & --- & 1472\,{\tiny\color{black!60}(6)} \\
\texttt{claude-fable-5.1} & $+0.45$\,{\tiny\color{black!60}(6)} & $+0.69$\,{\tiny\color{black!60}(3)} & $+0.65$\,{\tiny\color{black!60}(2)} & $-0.41$\,{\tiny\color{black!60}(8)} & $+0.05$\,{\tiny\color{black!60}(6)} & $-0.15$\,{\tiny\color{black!60}(6)} & $+1.28$\,{\tiny\color{black!60}(1)} & $+1.07$\,{\tiny\color{black!60}(3)} & $+0.04$\,{\tiny\color{black!60}(3)} & 1502\,{\tiny\color{black!60}(1)} \\
\texttt{gemma-4-31b} & $+0.05$\,{\tiny\color{black!60}(7)} & $-0.19$\,{\tiny\color{black!60}(6)} & $-0.05$\,{\tiny\color{black!60}(6)} & $+0.11$\,{\tiny\color{black!60}(5)} & $-0.60$\,{\tiny\color{black!60}(8)} & $-0.21$\,{\tiny\color{black!60}(7)} & $-0.81$\,{\tiny\color{black!60}(8)} & $-0.92$\,{\tiny\color{black!60}(8)} & $-0.08$\,{\tiny\color{black!60}(8)} & 1452\,{\tiny\color{black!60}(7)} \\
\texttt{deepseek-v3.1} & $-0.14$\,{\tiny\color{black!60}(8)} & $-0.33$\,{\tiny\color{black!60}(7)} & $+0.06$\,{\tiny\color{black!60}(5)} & $-0.07$\,{\tiny\color{black!60}(6)} & $+0.34$\,{\tiny\color{black!60}(3)} & $+0.28$\,{\tiny\color{black!60}(4)} & $-0.55$\,{\tiny\color{black!60}(6)} & $-0.98$\,{\tiny\color{black!60}(9)} & $+0.04$\,{\tiny\color{black!60}(4)} & 1417\,{\tiny\color{black!60}(10)} \\
\texttt{gemini-3.1-flash-lite} & $-0.47$\,{\tiny\color{black!60}(9)} & $-0.52$\,{\tiny\color{black!60}(8)} & $-0.72$\,{\tiny\color{black!60}(8)} & $-0.26$\,{\tiny\color{black!60}(7)} & $-0.62$\,{\tiny\color{black!60}(9)} & $-0.47$\,{\tiny\color{black!60}(9)} & $-0.78$\,{\tiny\color{black!60}(7)} & $-0.86$\,{\tiny\color{black!60}(7)} & $-0.03$\,{\tiny\color{black!60}(7)} & 1433\,{\tiny\color{black!60}(9)} \\
\texttt{qwen-3.5} & $-0.55$\,{\tiny\color{black!60}(10)} & $-0.61$\,{\tiny\color{black!60}(9)} & $-1.23$\,{\tiny\color{black!60}(10)} & $-1.28$\,{\tiny\color{black!60}(10)} & $+0.08$\,{\tiny\color{black!60}(5)} & $-0.26$\,{\tiny\color{black!60}(8)} & $-0.90$\,{\tiny\color{black!60}(9)} & $-0.47$\,{\tiny\color{black!60}(6)} & $-0.34$\,{\tiny\color{black!60}(10)} & 1441\,{\tiny\color{black!60}(8)} \\
\texttt{llama-3.3-70b} & $-2.56$\,{\tiny\color{black!60}(11)} & $-1.76$\,{\tiny\color{black!60}(10)} & $-1.18$\,{\tiny\color{black!60}(9)} & $-1.24$\,{\tiny\color{black!60}(9)} & $-1.42$\,{\tiny\color{black!60}(10)} & $-1.45$\,{\tiny\color{black!60}(10)} & $-2.00$\,{\tiny\color{black!60}(10)} & $-2.58$\,{\tiny\color{black!60}(10)} & $-0.28$\,{\tiny\color{black!60}(9)} & 1318\,{\tiny\color{black!60}(11)} \\
\midrule
\textbf{Spearman with the 50 games} & --- & $+0.96$ & $+0.83$ & $+0.88$ & $+0.66$ & $+0.84$ & $+0.79$ & $+0.78$ & $+0.82$ & $+0.75$ \\
Models & 11 & 10 & 10 & 10 & 10 & 10 & 10 & 10 & 10 & 11 \\
\bottomrule
\end{tabular}}
\end{table}

\subsection{Fresh games and games with no betting}

\begin{figure}[h]
\centering
\includegraphics[width=0.9\linewidth]{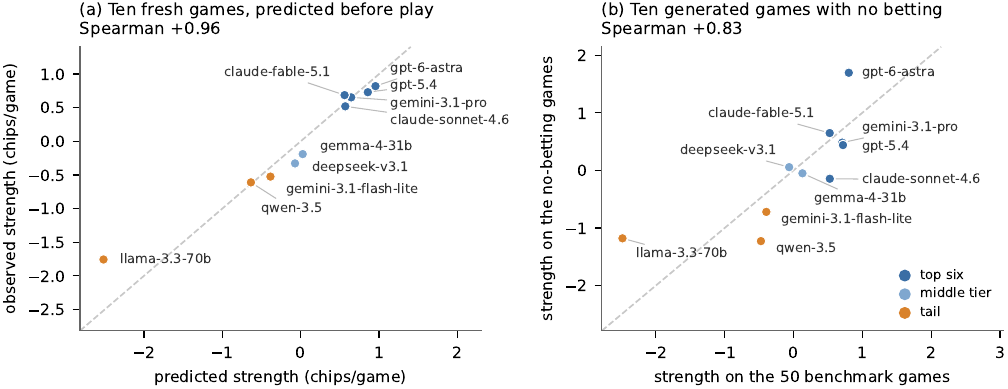}
\caption{\textbf{The ranking carries over to new games.} (a) Ten fresh games from the generator: observed strength
against the strength predicted from the 50 games before any match. (b) Ten generated games with no betting: strength
there against strength on the 50 benchmark games.}
\label{fig:add_games}
\end{figure}

\paragraph{Fresh games, predicted before play.} We drew ten new games from the generator and predicted each
model's strength on them from its results on the 50 games, freezing the predictions
before any match. Over 440 matches, observed strength follows the prediction at Spearman $+0.96$ $[+0.84, +0.99]$,
and the per-game predictions are calibrated (slope $+0.81$ $[+0.38, +1.21]$, Figure~\ref{fig:add_games}(a)).

\paragraph{Games with no betting.} Ten games from \textsc{GENSTRAT}'s general card-game generator have no betting at all:
drafting, playing cards for their effects, simultaneous moves, and scoring by sums, high cards or sets. Strength on
them agrees with strength on the 50 benchmark games at Spearman $+0.83$ $[+0.65, +0.95]$ over 440 matches
(Figure~\ref{fig:add_games}(b)).

\subsection{Six new games}
\label{app:families}

To test whether the ranking reflects strategic skill rather than familiarity with betting games, we built six games
outside the GBG generator (Table~\ref{tab:families}). All six run on the \textsc{GENSTRAT} engine, which shows that it is
not tied to betting games. The three card games are composed directly from its game-operation and phase modules. The
other three come from its families for board games, markets and contracts: a hex board game with fog of war, a repeated
market, and a negotiation, the last two general-sum. The board game is drawn from a seed by a builder that composes
registries of pieces, terrain, goals and fighting rules, and every sentence of its rulebook is the text of the module
that executes it. Before any match, an outside model audited each served rulebook against its engine, and every
random quantity is drawn for each match from the distribution its rulebook states. Ten models played seat-swapped
pairs, each pair playing the same deal twice with the seats swapped. The analysis was fixed before play: per game,
the paper's chip-margin estimator gives $\hat\alpha_{m,g}$ in that game's payoff units, tested by a one-sided exact
Spearman permutation test against the paper's leaderboard. Pooling uses $\hat\alpha_{m,g}/\sigma_U(g)$, where
$\sigma_U(g)$ is the standard deviation of a player's payoff when both play uniformly at random. For the two general-sum
games, the margin is the seat-swapped difference of the two players' own payoffs.

\begin{table}[h]
\centering\small
\caption{\textbf{The six new games and their agreement with the paper's leaderboard.} $\rho$: Spearman correlation of
$\hat\alpha_{m,g}$ with the paper's leaderboard over the ten models.}
\label{tab:families}
\setlength{\tabcolsep}{4pt}
\begin{tabular}{@{}l p{0.52\linewidth} c r@{}}
\toprule
Game & What it tests & Pairs & $\rho$ \\
\midrule
Hidden Agenda & card game: secret objectives, steal, peek and guard, then an accusation round & 36 & $+0.88$ \\
Market Salad & card game: market drafting with a poison rank and a scoring branch on discarded sets & 36 & $+0.66$ \\
Declare Race & card game of timing and bluffing: a challengeable declare ends the game & 36 & $+0.84$ \\
Board & hex board game: fog of war, secret goals, face-down tiles (Figure~\ref{fig:board_moves}) & 36 & $+0.79$ \\
Market & repeated product competition with persistent capacity, redesign costs and a random horizon & 30 & $+0.78$ \\
Bargaining & four-issue contract negotiation with private values, warranties, escrow and cheap talk & \EthreePairs & \EthreeRho \\
\midrule
Pooled, mean $\hat\alpha_{m,g}/\sigma_U(g)$ & all six games & & \PooledSixRho \\
\bottomrule
\end{tabular}
\end{table}

\paragraph{Results.} Every game orders the ten models like the paper's leaderboard (Table~\ref{tab:families}), and
pooled over the six games the agreement is almost perfect (Figure~\ref{fig:families_scatter}a), much closer than with
a general-capability leaderboard (Figure~\ref{fig:families_scatter}b).
Single games agree at $+0.66$ to $+0.88$, and pooling the games raises the agreement to \PooledSixRho, which is the
paper's argument for evaluating on a distribution of games rather than on any single game. In the two games with a
clear tier split, the board game and the market, all five frontier models have positive strength and all five others negative. Individual models keep
recognizable profiles across games: \texttt{claude-fable-5.1}, fifth on the benchmark, ranks first on the board game
and third on the market but sixth to eighth on the three card games.

\paragraph{Bargaining.} In the negotiation, a model's own earnings also follow the paper's order, with a rank correlation of \EthreeOwnRho{}, so
stronger models claim more of each deal. Joint value created over the no-deal outcome follows the paper's order less closely,
at \EthreeValueRho{}. The only deals accepted below the acceptor's walk-away
value were made, with one exception, by the five weaker models.

\begin{figure}[h]
\centering
\includegraphics[width=\linewidth]{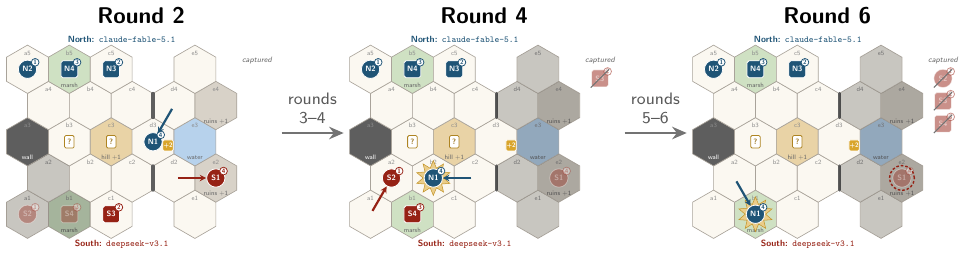}
\caption{\textbf{The board game.} A graphical representation of the procedurally generated board game, over three rounds of
a tournament match between \texttt{claude-fable-5.1} (North, blue) and \texttt{deepseek-v3.1} (South, red).}
\label{fig:board_moves}
\end{figure}

\subsection{General capability and distance from poker}

\paragraph{General capability.} A ranking that holds across games could simply track general capability. We
tested this against LMArena's text leaderboard (style control on, snapshot of 11 November 2026), with the analysis
fixed before any score was looked up. LMArena agrees with the paper's leaderboard at Spearman $+0.75$ over eleven
models (Table~\ref{tab:heldout}, last column), lower than every held-out set of games except Market Salad. For the
test we use the ten models of Appendix~\ref{app:families} and their mean $\hat\alpha_{m,g}/\sigma_U(g)$ over the three
new games all ten had completed when the test was fixed (Hidden Agenda, Market Salad and Board). LMArena agrees with
that strength at $+0.68$, and the paper's leaderboard at $+0.99$. Controlling for LMArena, the paper's leaderboard
still predicts it almost perfectly: the partial rank correlation is $+0.98$, with a 95\% CI of $[+0.76, +1.00]$ from resampling models
(one-sided exact permutation $p\approx6\times10^{-6}$). With all six new games pooled, a check made after the test, it is $+0.90$ $[+0.34, +1.00]$ ($p\approx4\times10^{-4}$), and LMArena alone agrees with that strength at $+0.72$. The models that do better in games than their LMArena rank suggests (\texttt{gpt-6-astra},
\texttt{gpt-5.4}) do so on both the benchmark and the new games, and so do the models that do worse
(\texttt{claude-fable-5.1}, \texttt{qwen-3.5}).

\paragraph{Distance from known poker games.} If models won by recognizing poker, the leaderboard should reproduce
best on games close to Kuhn or Leduc poker. We measured each game's distance from Kuhn and Leduc poker in two ways,
by structure and by the five axes. Controlling for the five axes, distance from poker predicts nothing about how
strongly a game reproduces the leaderboard: every estimate lies within $\pm0.1$ chips/game per SD of distance.
Without that control the frontier's lead \emph{grows} with distance from poker, by $+0.26$ to $+0.35$ chips/game per
SD, the opposite of what prior knowledge of poker would produce.

\section{Where the frontier's lead comes from}
\label{app:where}

\begin{figure}[h]
\centering
\includegraphics[width=0.7\linewidth]{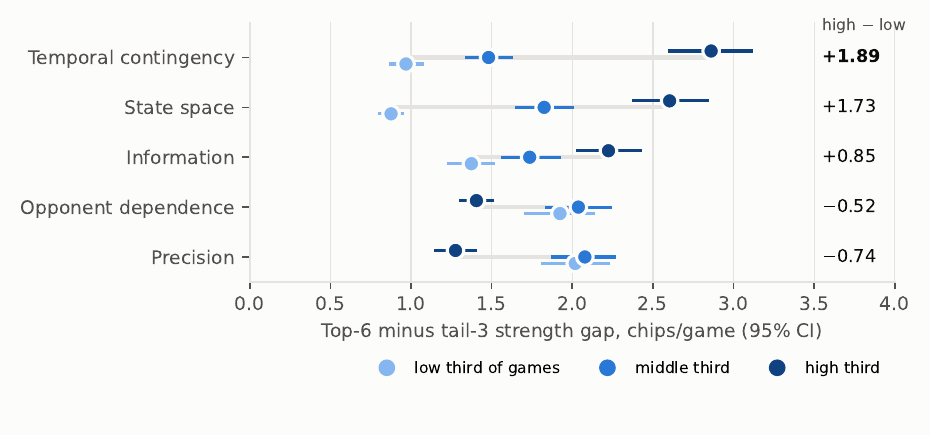}
\caption{\textbf{Strength gap between the top six models and the three weakest} in the low, middle and high third of
the 50 games on each axis.}
\label{fig:gap_by_axis}
\end{figure}

\paragraph{The lead grows with temporal contingency.} Figure~\ref{fig:gap_by_axis} splits the 50 games into thirds on each axis and
refits the strength gap between the top six models and the three weakest. From the low to the high third, the gap grows
by about 1.9 chips/game with temporal contingency and 1.7 with state space, grows less with information, and shrinks
with opponent dependence and precision. The models thus separate on games where the best action depends on how the deal unfolds, and the
gap narrows on games
that reward modeling the opponent. This fits the trace analysis and the program ablation (Section~\ref{sec:ablations}), where reasoning
about the opponent does not pay off and frontier models fully exploit a strategy only when they are told it.

\paragraph{Two levels of skill.} Weaker models lose on basic hand strength: 56\% of the tail's deficit against the
top six comes from deals with weak hands, and the costliest single error is calling a bet with one. Frontier models
make few such errors (Appendix~\ref{app:solved}) and separate from one another on hard games high in temporal contingency.

\begin{figure}[H]
\centering
\begin{minipage}[t]{0.5\linewidth}\vspace{0pt}\centering\includegraphics[width=\linewidth]{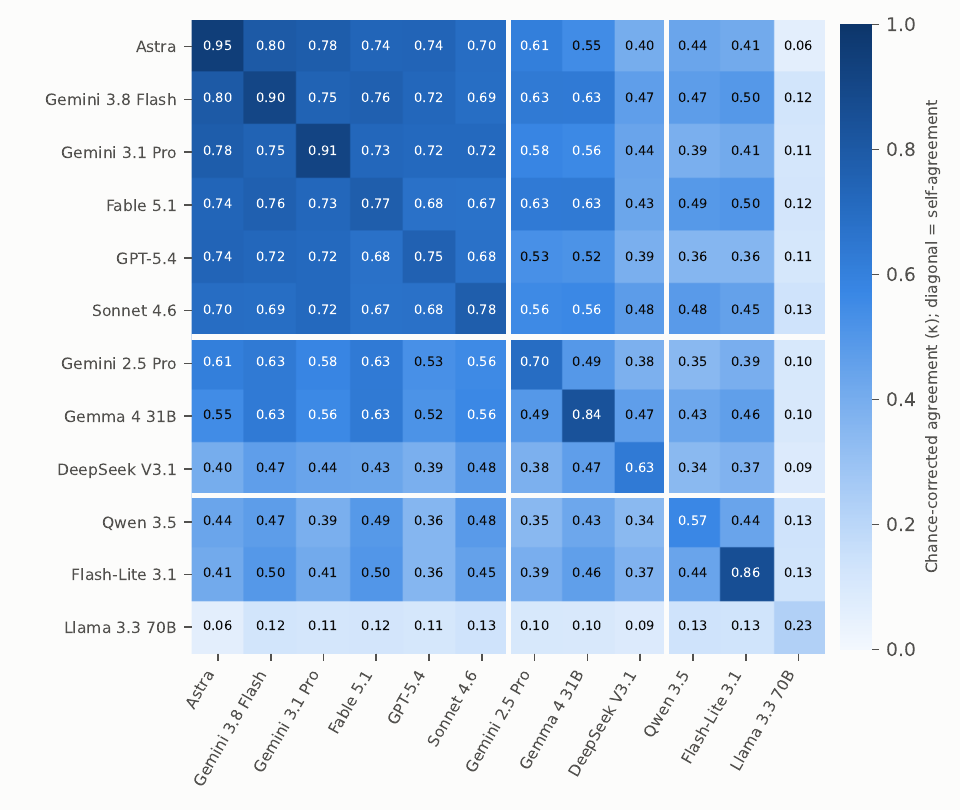}\end{minipage}\hfill
\begin{minipage}[t]{0.47\linewidth}\vspace{0pt}\centering\includegraphics[width=\linewidth]{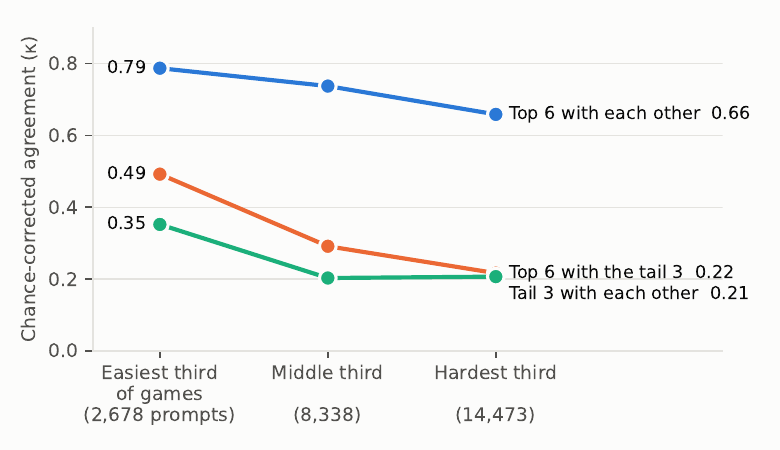}\end{minipage}
\caption{\textbf{Decision-level agreement between models.} Left: chance-corrected agreement on identical prompts, ordered by
average-linkage clustering. Right: agreement within the top six and between the top six and the tail, by the game's
hardness third.}
\label{fig:similarity}
\end{figure}

\paragraph{Models cluster by capability tier.} On $25{,}489$ corpus prompts that two or more models answered with
identical input, models in the same capability tier agree with one another more often than other pairs do, even once lab is
controlled for, while models from the same lab agree no more often than others (Figure~\ref{fig:similarity}). Harder games pull
the tiers apart, and agreement drops both within the top six and between the top six and the tail. Disagreement
between strong models is mostly harmless. When \texttt{gpt-6-astra} and \texttt{gemini-3.1-pro} disagree on a solved
game, both actions are optimal 83\% of the time.

\section{Validating generated games}
\label{app:validation}

Validation works at two levels. Each module and its rule text are checked once, and for the board-game and economic
families behind the new games of Appendix~\ref{app:families}, LLM reviewers audited the modules one by one against
their code. Every composed game is then checked automatically: by the generation gates (Section~\ref{sec:gbg}), by
replaying the engine against the recorded play, by a checker of 19 kinds of consistency between the served rulebook
and the engine, and by having a model from another family rebuild each game from its rulebook alone. People only
adjudicate what these checks flag, so the effort grows with the number of modules rather than the number of games. On
the 50 benchmark games, the checks found one engine bug, which affected about 1\% of matches and is now fixed, and a
number of wording issues, mostly details the text left out. One of these is the hidden-cost sentence, a single shared
sentence in 12 games that moved no chips, because the rulebooks also say that both stacks are public. To see whether
the wording changed how models play, we asked models the same prompt with the original sentence and with only that
sentence clarified. Clarifying changed no model's decisions in any of the 12 games we tested, apart from an auction
sentence we had already found, where it changed one model's bids. We then reran in full the two games with the engine
bug and the two games with an auction round where our tests showed the wording was most likely to matter. No model
moved more than one rank, and the largest change in a model's strength was 0.13 chips per game. We also reran 542
tournament matches across all 34 games whose rulebooks were revised, on the same deals as the originals but with the
corrected text and engine. Decisions matched the originals 84 to 94\% of the time, a few points below how often the same models
repeat their own answer to an identical prompt (88 to 97\%), no model changed tier, and one of seven
model pairings changed significantly (\texttt{gpt-5.4} against \texttt{gpt-6-astra}), which may also reflect changes in
the models since the tournament.

\paragraph{Testing the checks.} We tested the rebuild by planting rule changes in the engine of ten benchmark games,
under both the served and the corrected rulebooks. Four changes alter play: dropping a legal action, charging a cost
one step late, delaying the opponent's visible chip count, and discarding the newest card instead of the oldest. Two
do not: rewording the text and splitting a simultaneous step in two. The rebuild flagged 66 of the 68 harmful cases,
all 20 for each of the first three changes and 6 of the 8 game versions where the discard change applies, and none of
the harmless ones. Run against the engine itself, it raised no flag in 680 episodes, and it reproduced all 80 recorded
matches it replayed. On the 50 benchmark games the rebuild raised 82 flags, each adjudicated by replay. Most were
errors in the rebuild itself, some reproduced issues we already knew, and two were new wording issues, a discard pile
described as face-up although only the discarder learns the card, and an undefined ``top card''. Both are fixed in the
text. On the corrected rulebooks, the rulebook--engine checker verifies 2,927 claims across the 50 games in over
220,000 checks with no failure, and seeded replays reproduced 1,540 recorded matches exactly.

\end{document}